\documentclass[10pt,twocolumn,letterpaper]{article}
\usepackage{cvpr}
\usepackage{times}
\usepackage{amsmath}
\usepackage{amssymb}
\usepackage{graphicx}
\usepackage{sidecap}
\usepackage{subfigure}
\usepackage{tabularx}
\usepackage[english]{babel}
\usepackage{multirow}

\usepackage[pagebackref=true,breaklinks=true,letterpaper=true,colorlinks,bookmarks=false]{hyperref}

\newcommand{\specialcell}[2][c]{\begin{tabular}[#1]{@{}c@{}}#2\end{tabular}}

 \cvprfinalcopy % *** Uncomment this line for the final submission
\def\cvprPaperID{471} % *** Enter the CVPR Paper ID here

\ifcvprfinal\fi
\begin{document}
\title{Tracking Revisited using RGBD Camera: Baseline and Benchmark}

\author{Shuran Song\\ \texttt{srsong@ust.hk} \\Hong Kong University of Science and Technology\and Jianxiong Xiao\\ \texttt{jxiao@csail.mit.edu} \\Massachusetts Institute of Technology}
\maketitle

\begin{abstract}
Although there has been significant progress in the past decade,
tracking is still a very challenging computer vision task, due to problems such as occlusion and model drift.
Recently, the increased popularity of depth sensors (\eg Microsoft Kinect) has made it easy to obtain depth data at low cost.
This may be a game changer for tracking, since depth information can be used to prevent model drift and handle occlusion.
In this paper, we construct a benchmark dataset of 100 RGBD videos with high diversity, including deformable objects, various occlusion conditions and moving cameras.
We propose a very simple but strong baseline model for RGBD tracking, and present a quantitative comparison of several state-of-the-art tracking algorithms.
Experimental results show that including depth information and reasoning about occlusion significantly improves tracking performance.
The datasets, evaluation details, source code for the baseline algorithm, and instructions for submitting new models will be made available online after acceptance.

\end{abstract}

\section{Introduction}
In the last decade, tracking algorithms have evolved significantly 
in both their sophistication and quality of results. 
However, tracking is still considered a very challenging task in computer vision, particularly because a slight mistake in one frame may be reinforced after an online learning step, resulting in the so-called model drift problem.
Furthermore, occlusion of target objects occurs quite often in real world scenarios, and it is not clear how to model occlusions robustly.
The state-of-the-art methods (\eg \cite{tld,mil,semiB,ct}) usually employ very powerful learning and energy minimization methods in the hopes of better handling these issues.

Fortunately, we are moving into a 3D era for digital devices.
Accurate and affordable depth sensors,
such as Microsoft Kinect, Asus Xtion and PrimeSense, makes depth acquisition easy and cheap.
With an accurate depth map,
many traditional computer vision tasks become significantly easier (\eg human pose estimation \cite{Jamie}).
For tracking, the depth map
can provide valuable additional information to significantly improve results with 
much more robust occlusion and model drift handling.

\begin{figure}[t]
\centering	
\includegraphics[width=0.159\linewidth]{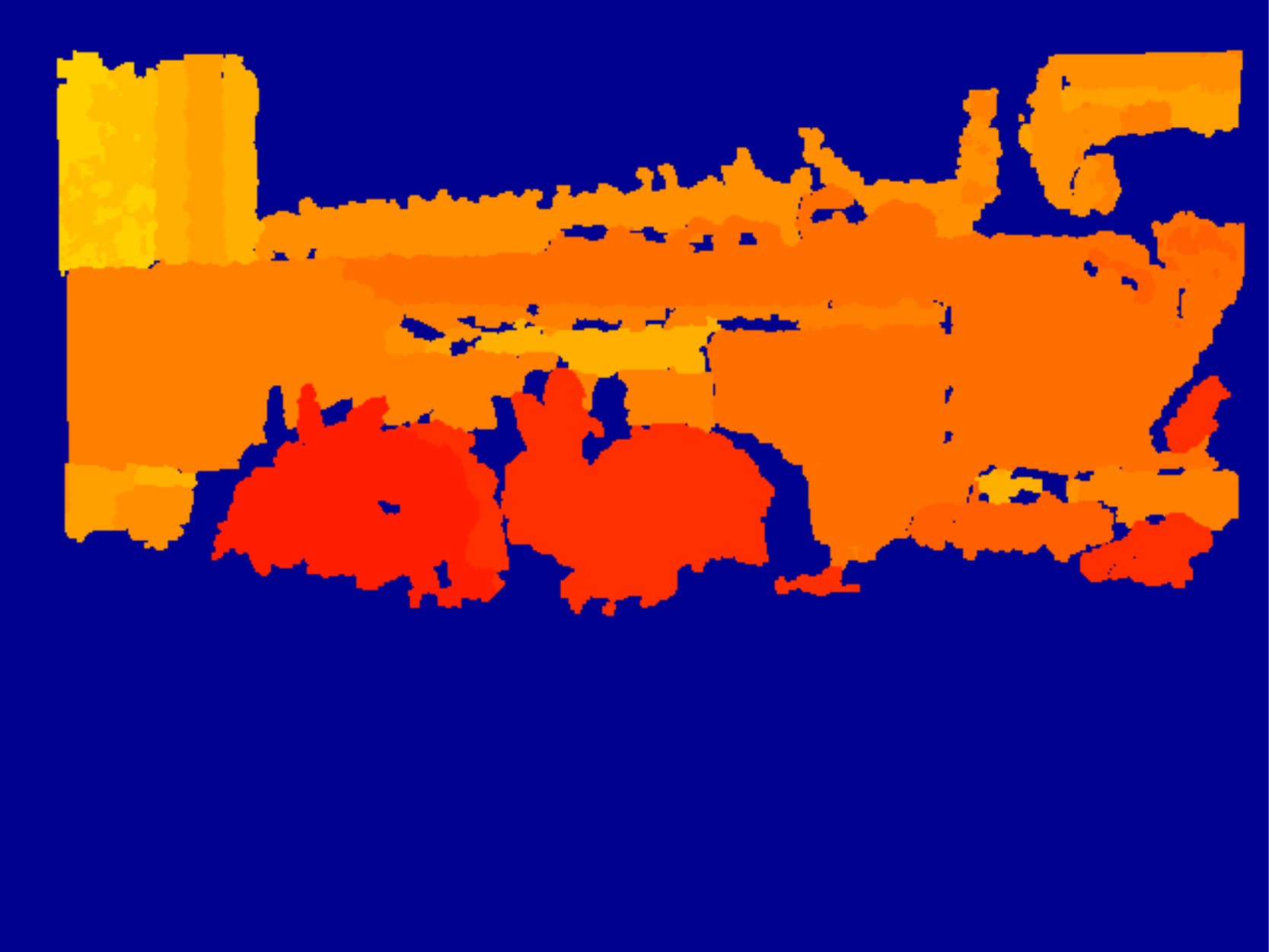}~%
\includegraphics[width=0.159\linewidth]{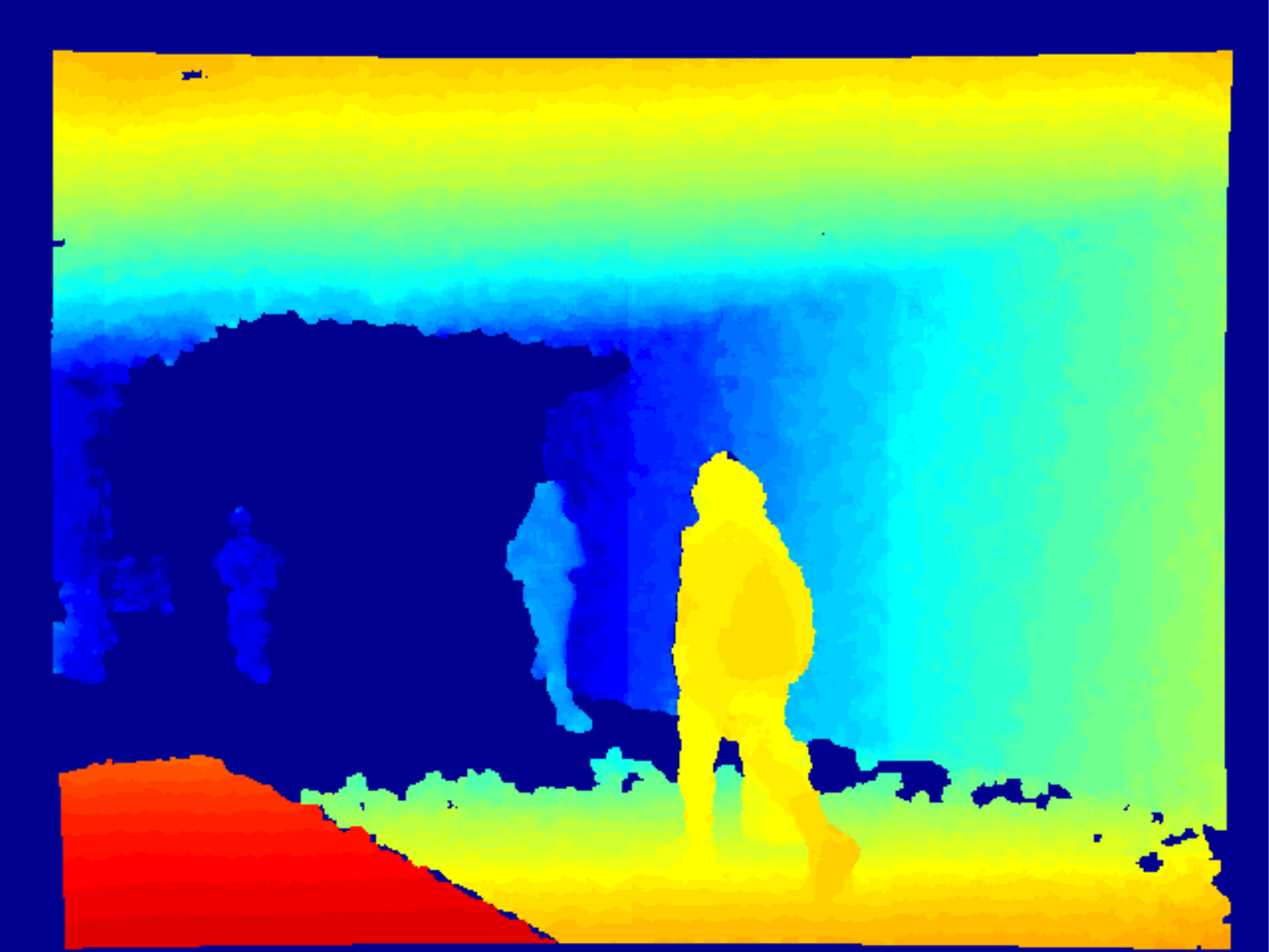}~%
\includegraphics[width=0.159\linewidth]{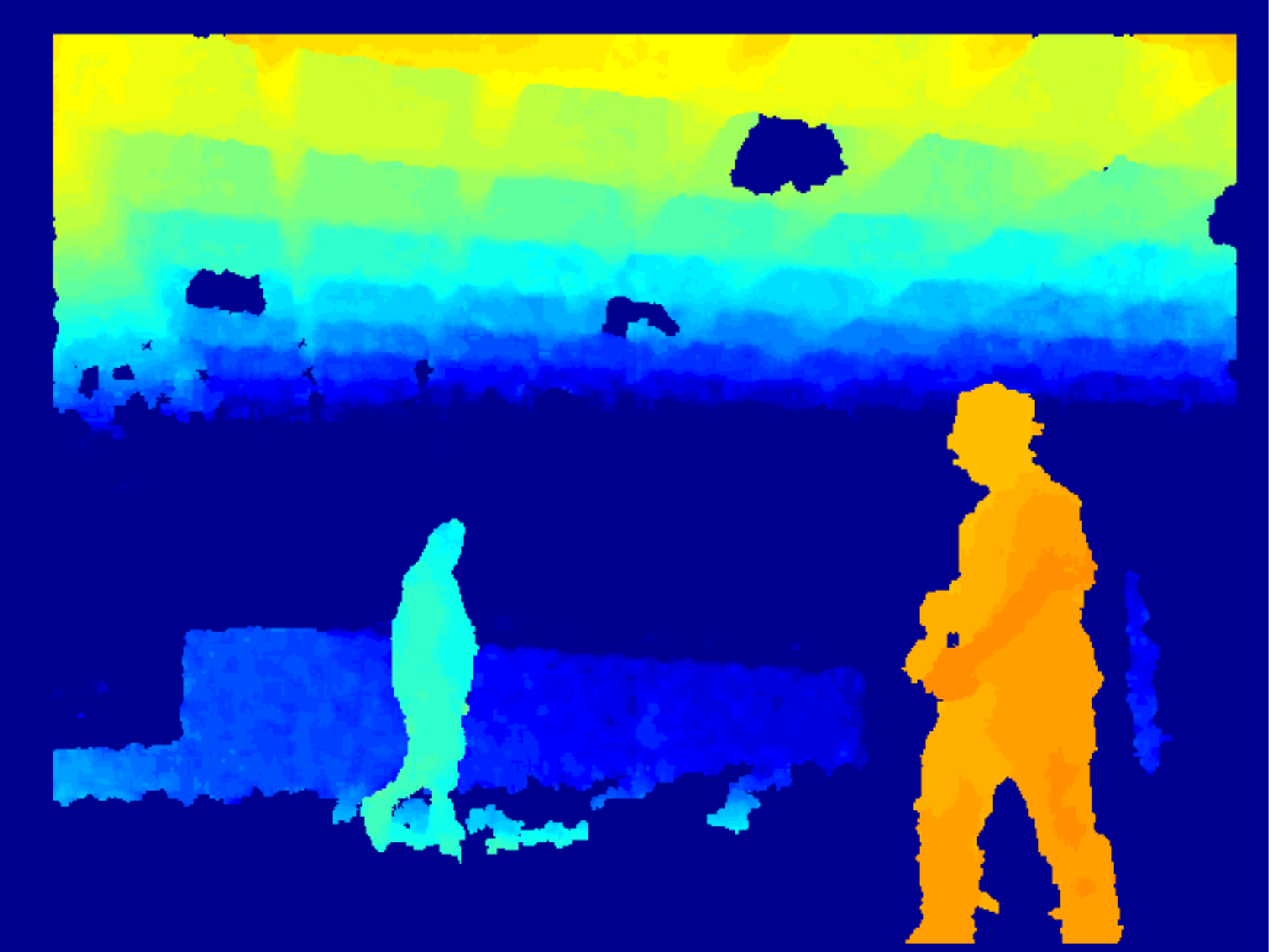}~%
\includegraphics[width=0.159\linewidth]{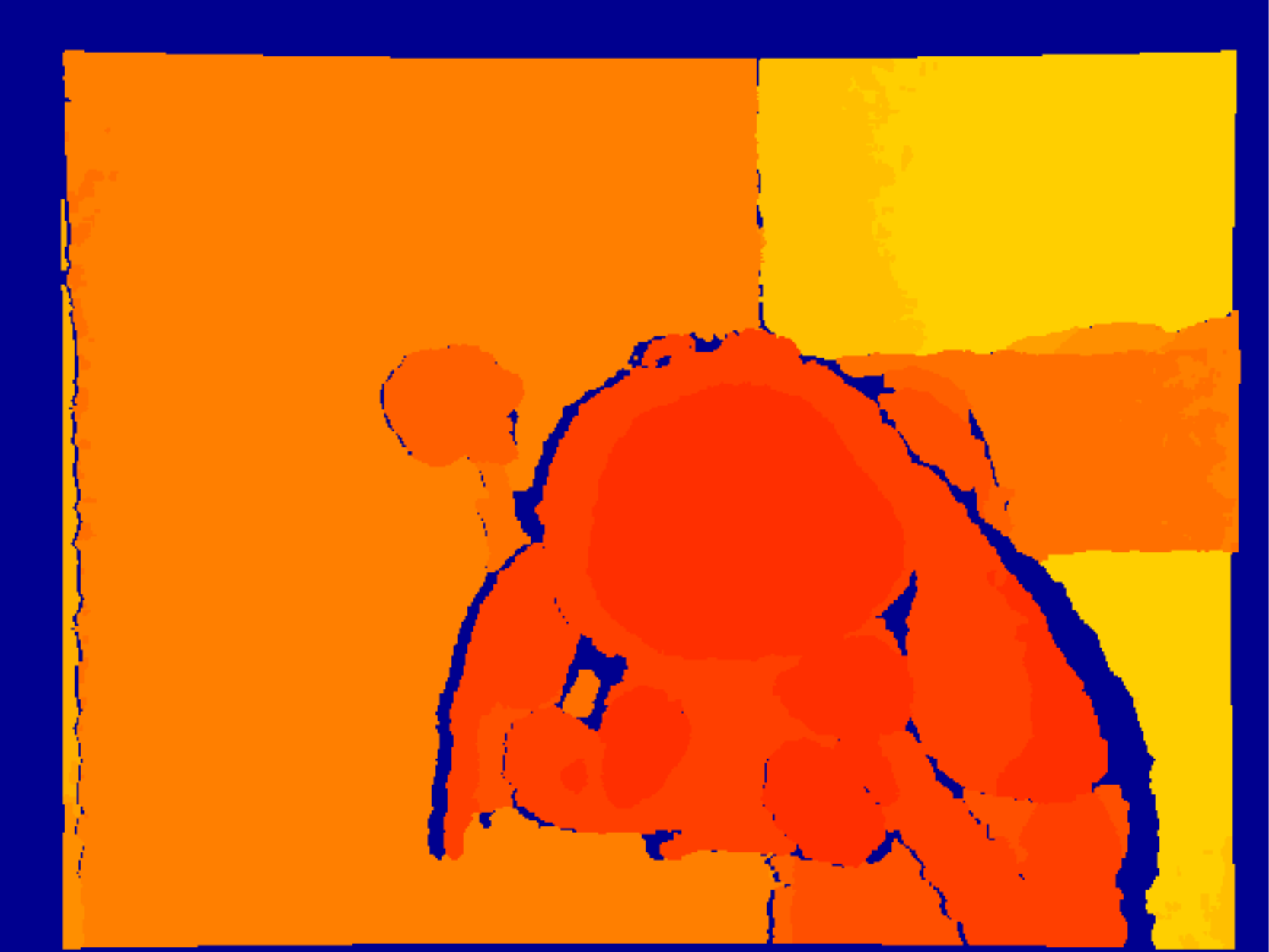}~%
\includegraphics[width=0.159\linewidth]{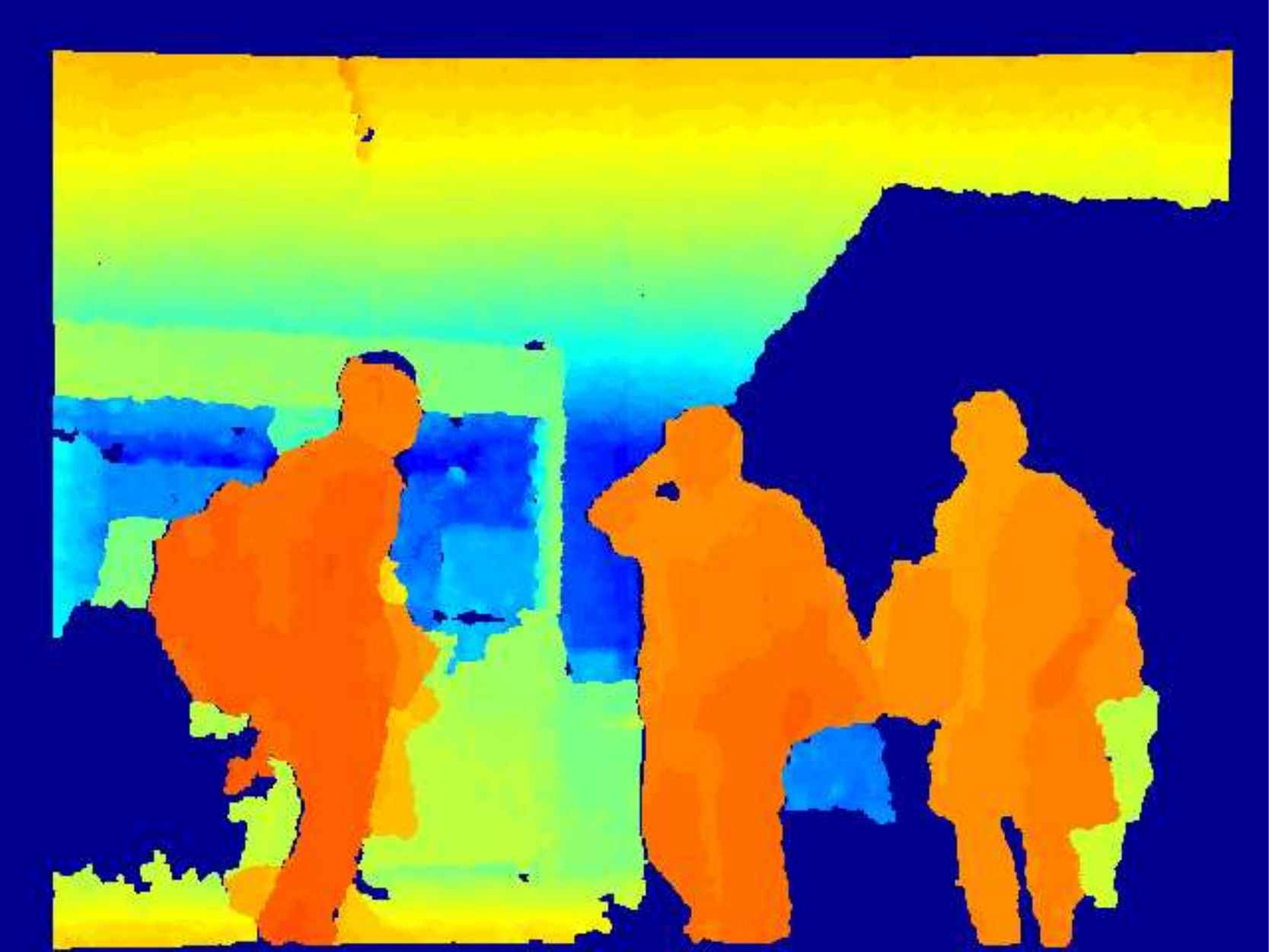}~%
\includegraphics[width=0.159\linewidth]{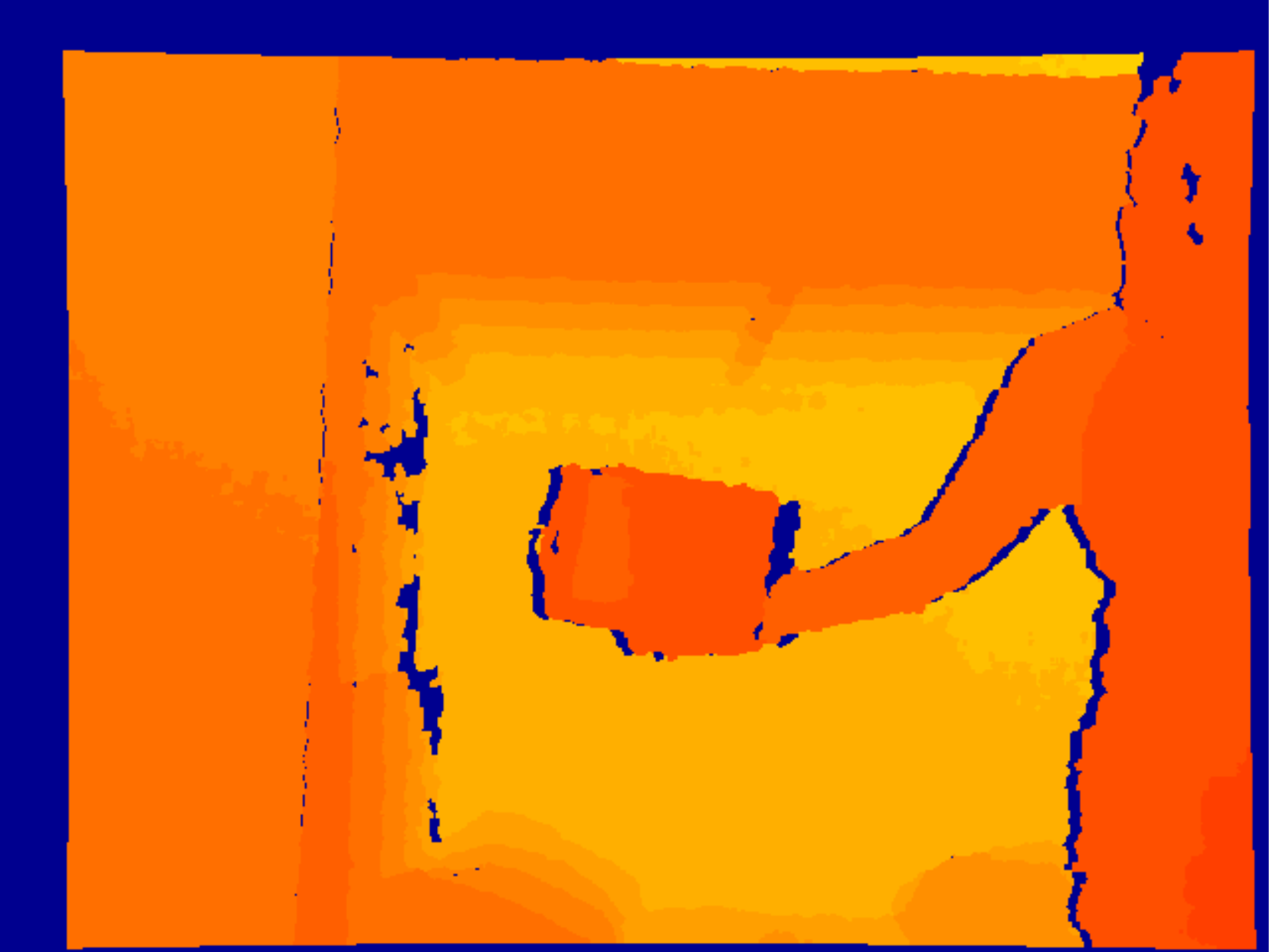}

\vspace{1mm}

\includegraphics[width=0.159\linewidth]{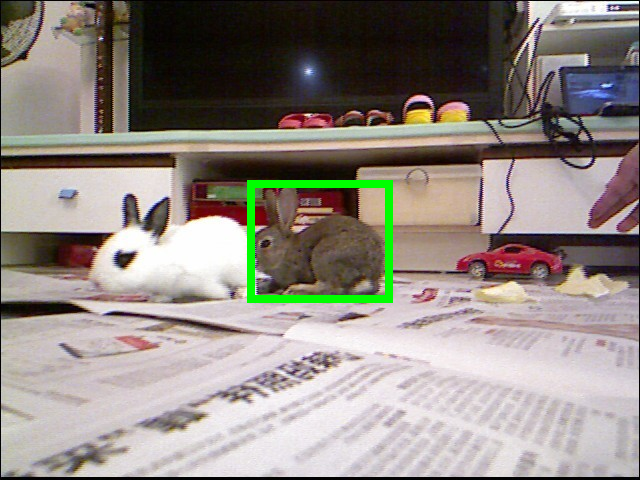}~%
\includegraphics[width=0.159\linewidth]{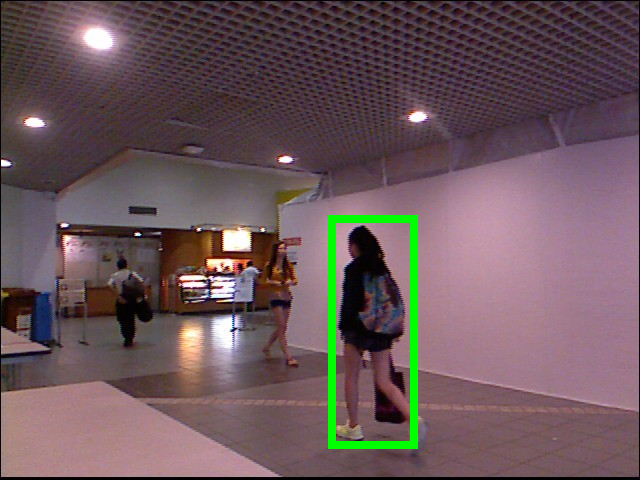}~%
\includegraphics[width=0.159\linewidth]{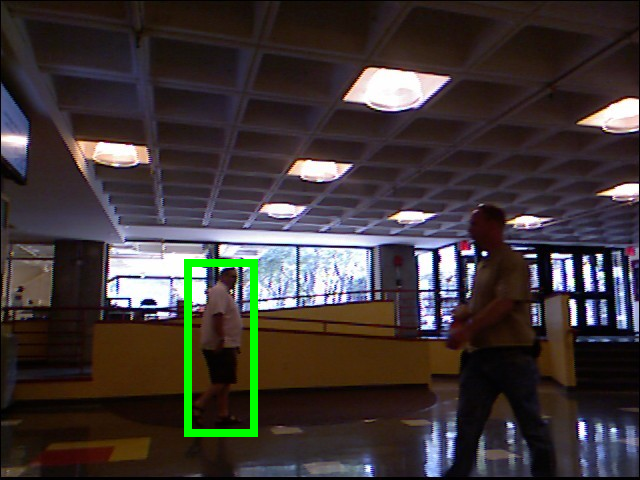}~%
\includegraphics[width=0.159\linewidth]{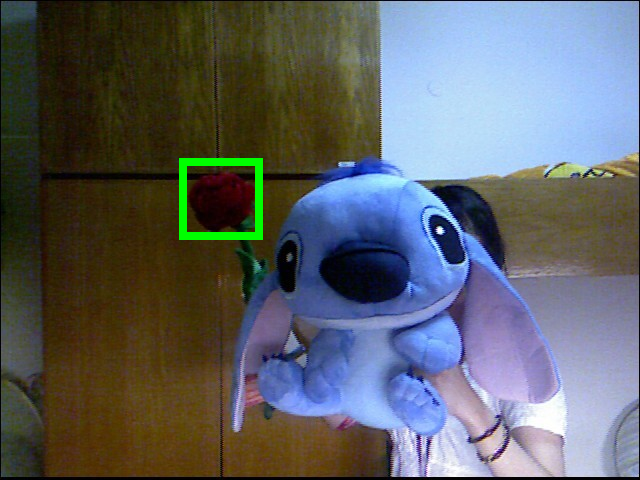}~%
\includegraphics[width=0.159\linewidth]{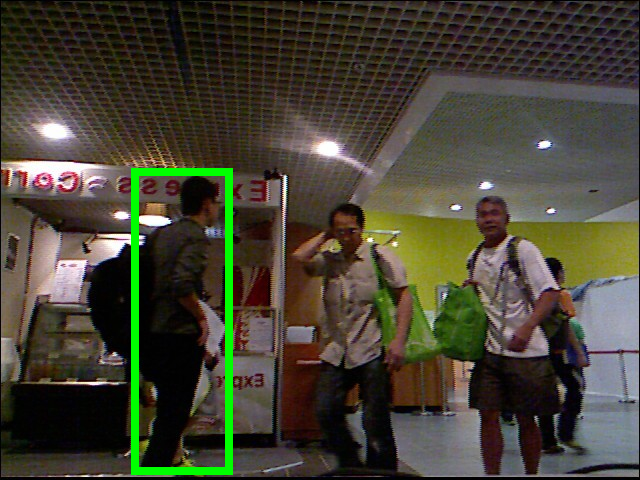}~%
\includegraphics[width=0.159\linewidth]{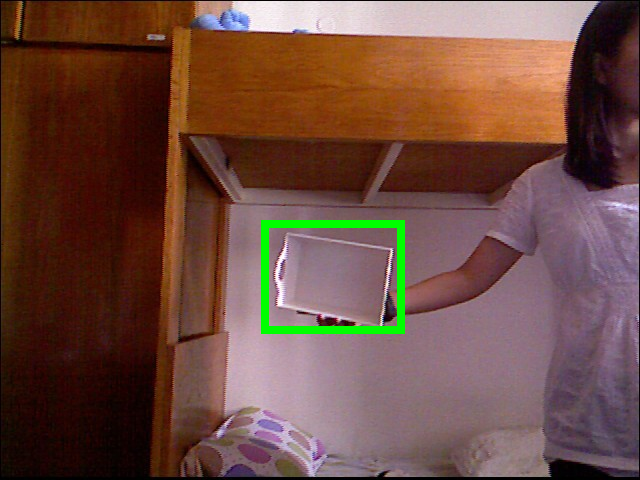}

\vspace{1mm}
\includegraphics[width=0.159\linewidth]{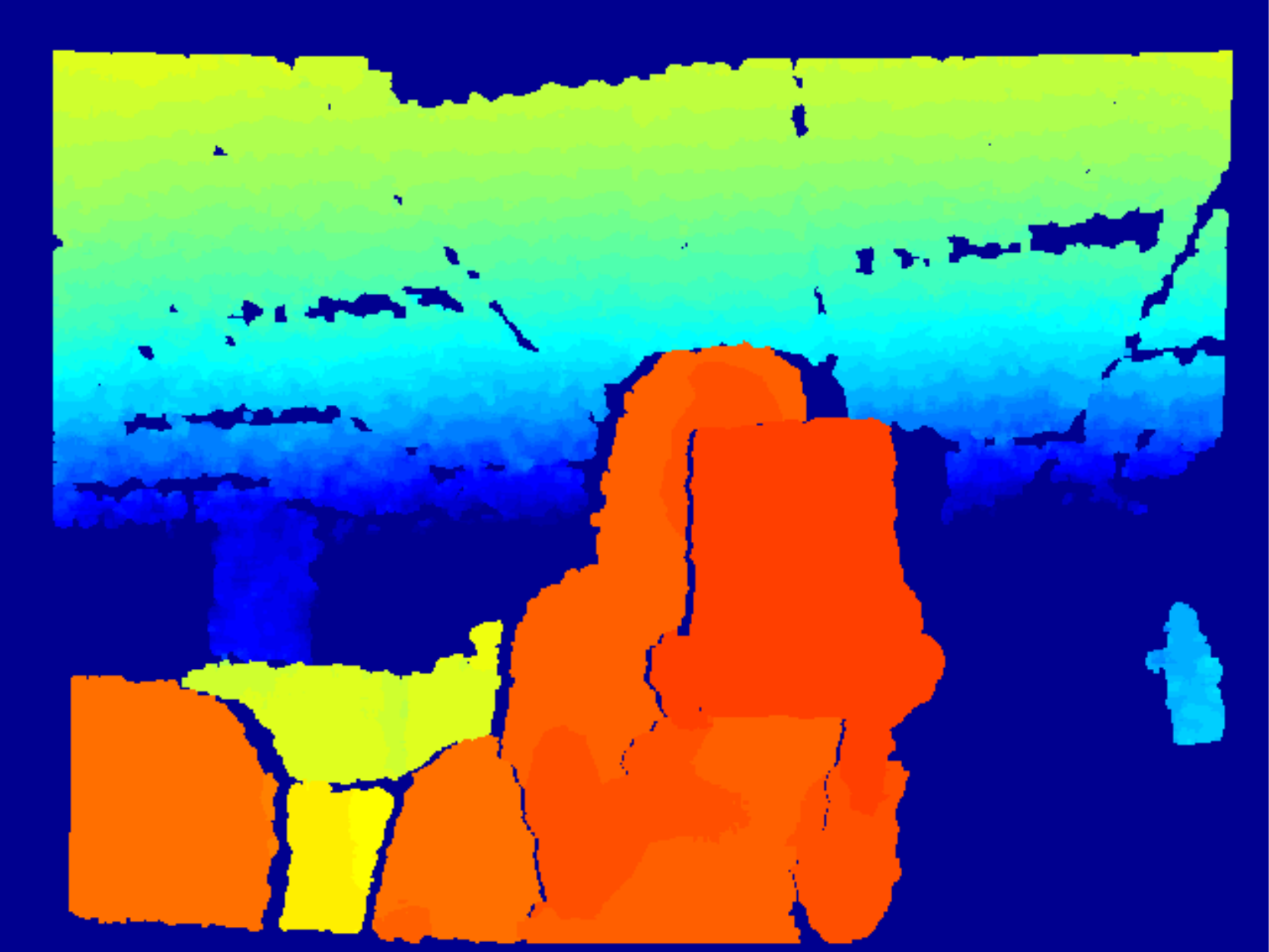}~%
\includegraphics[width=0.159\linewidth]{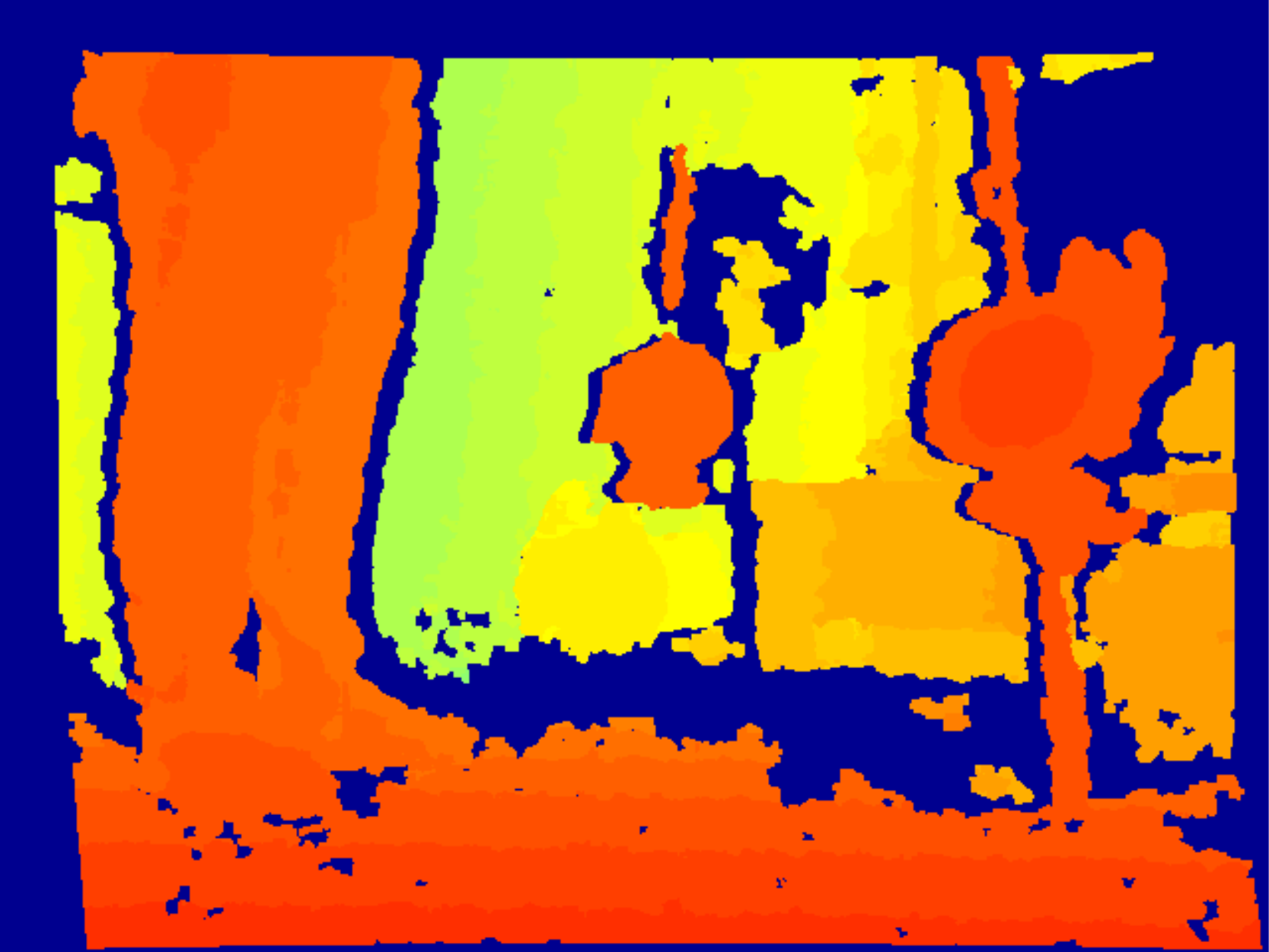}~%
\includegraphics[width=0.159\linewidth]{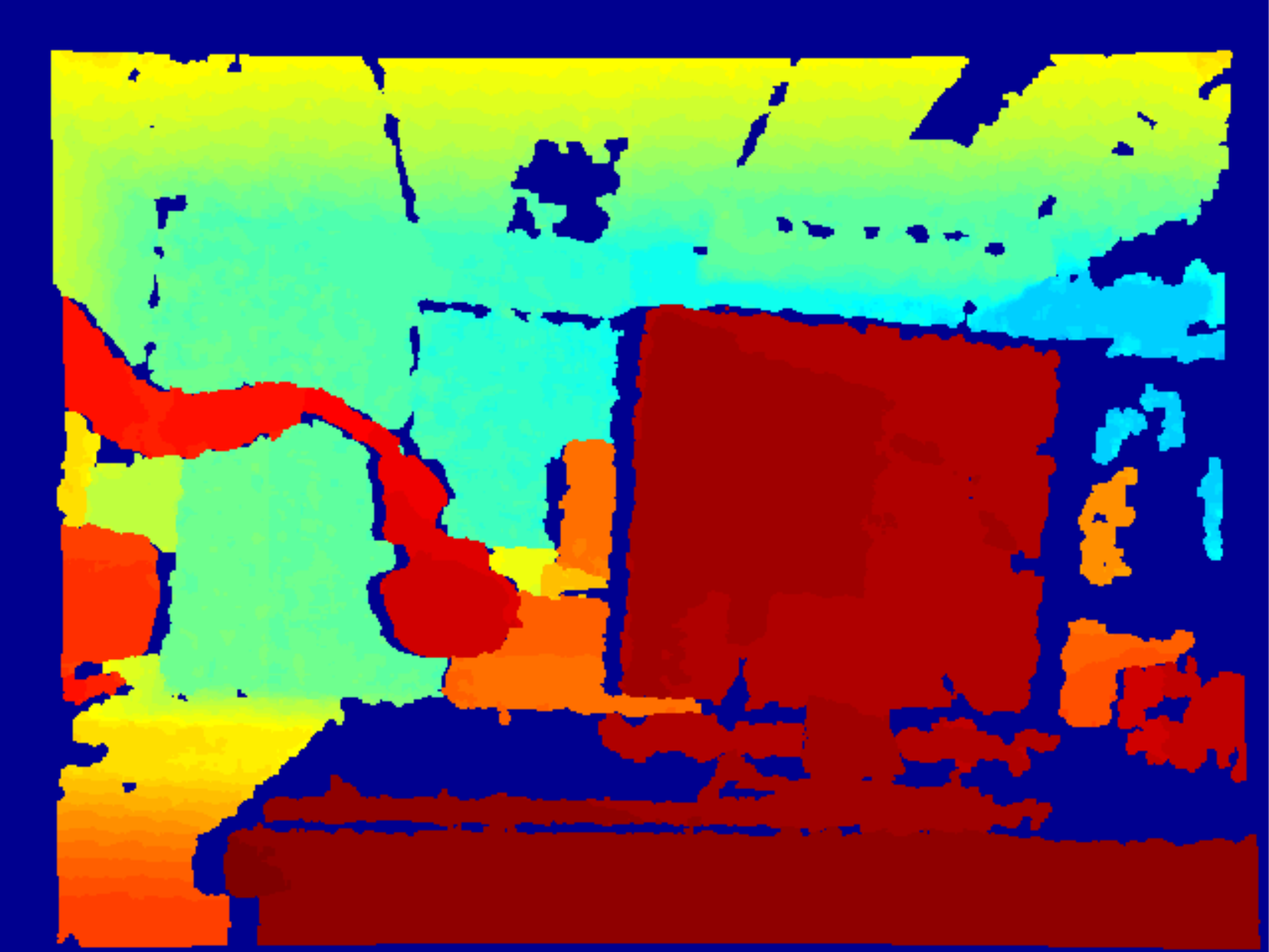}~%
\includegraphics[width=0.159\linewidth]{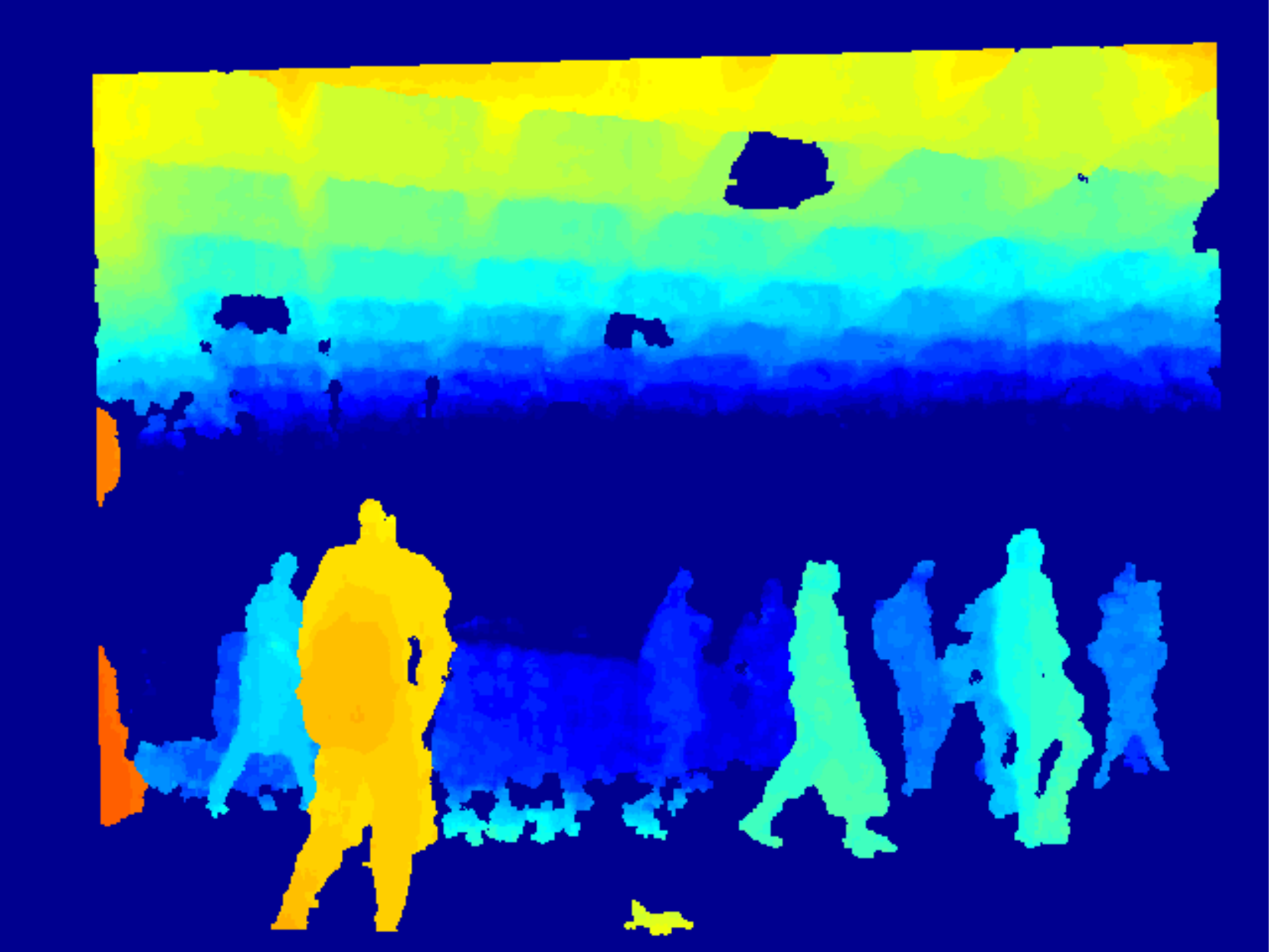}~%
\includegraphics[width=0.159\linewidth]{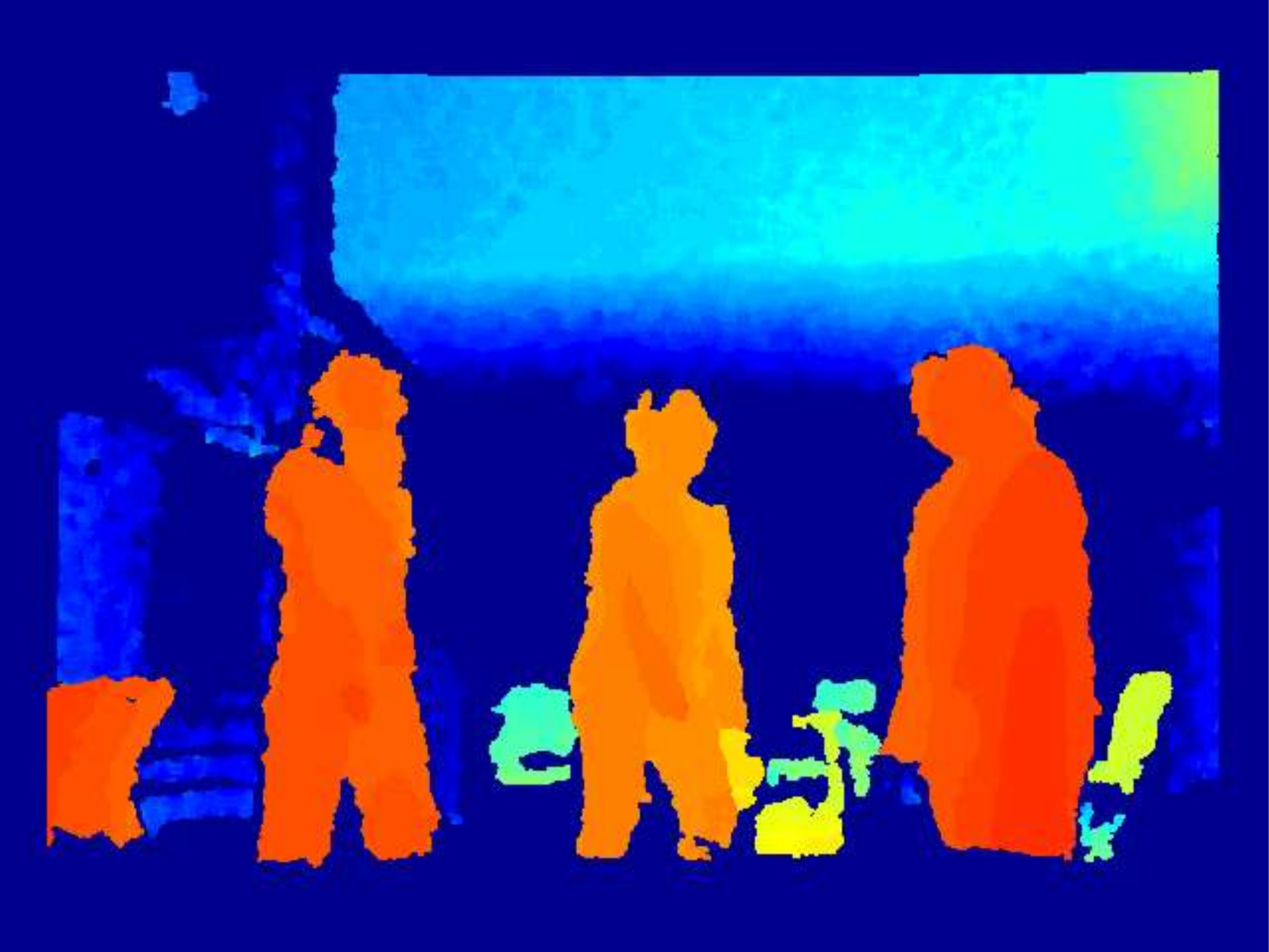}~%
\includegraphics[width=0.159\linewidth]{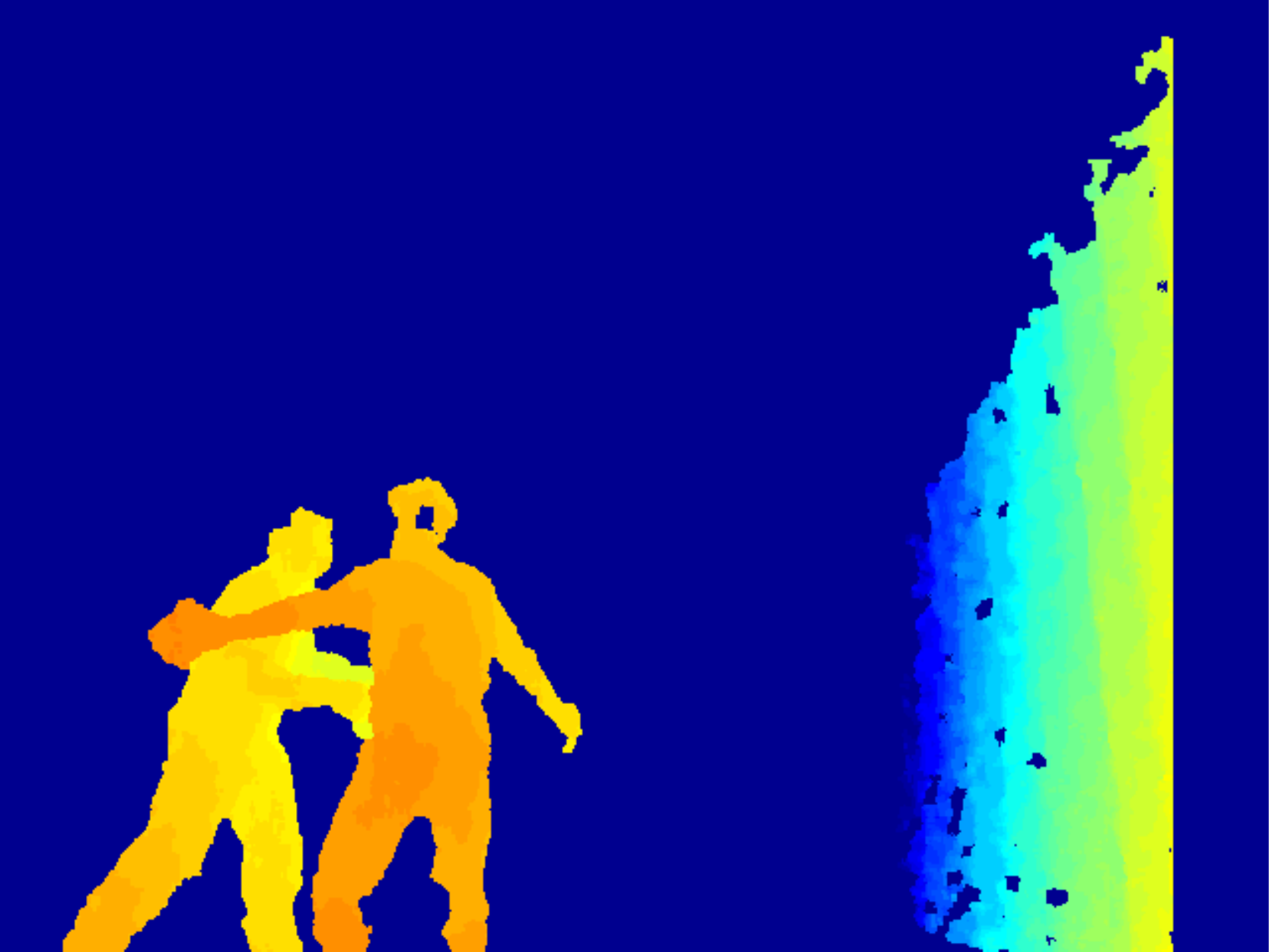}

\vspace{1mm}
\includegraphics[width=0.159\linewidth]{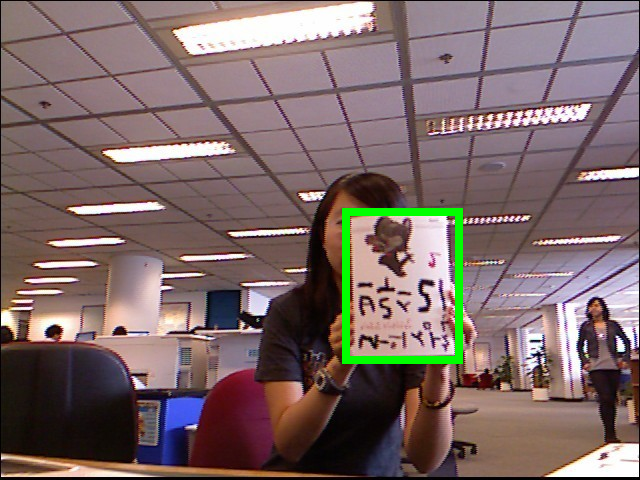}~%
\includegraphics[width=0.159\linewidth]{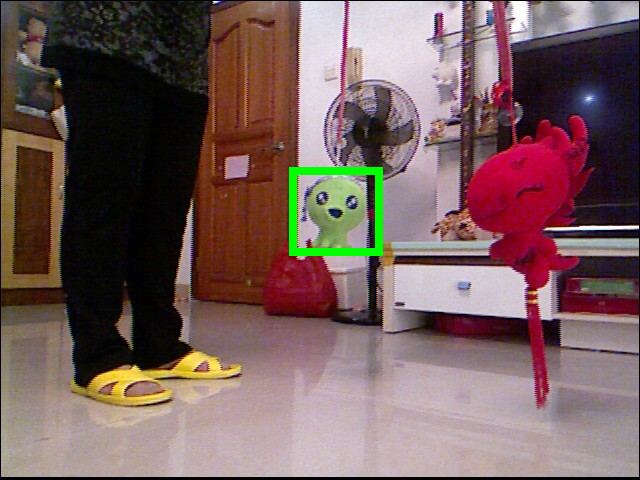}~%
\includegraphics[width=0.159\linewidth]{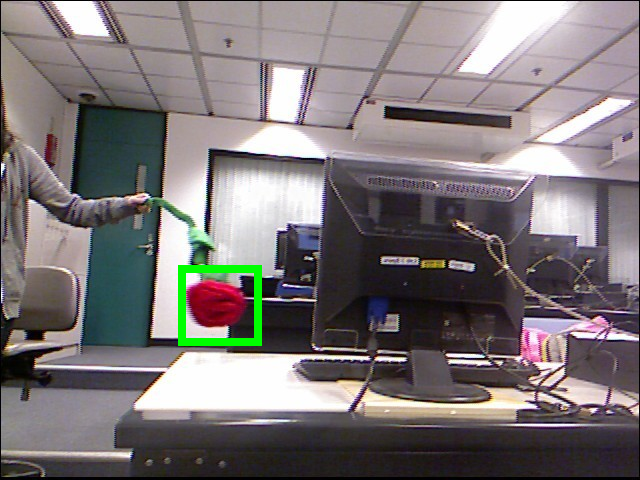}~%
\includegraphics[width=0.159\linewidth]{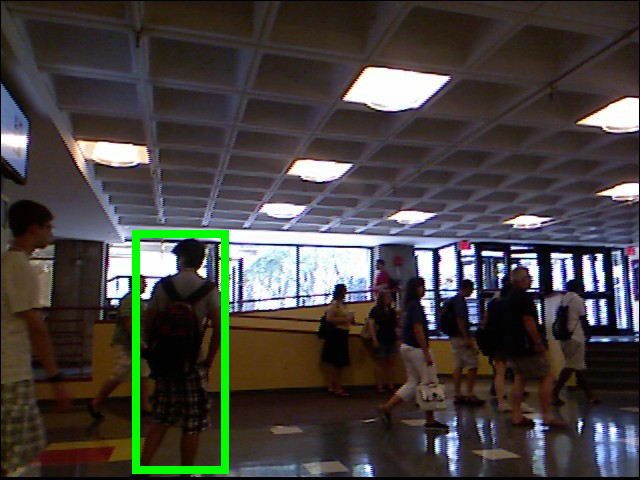}~%
\includegraphics[width=0.159\linewidth]{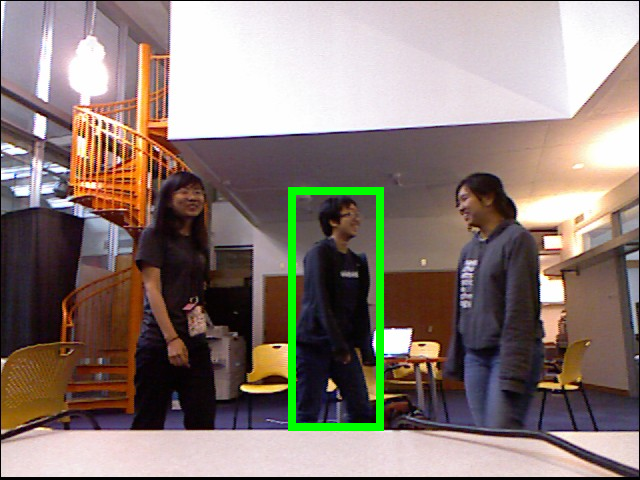}~%
\includegraphics[width=0.159\linewidth]{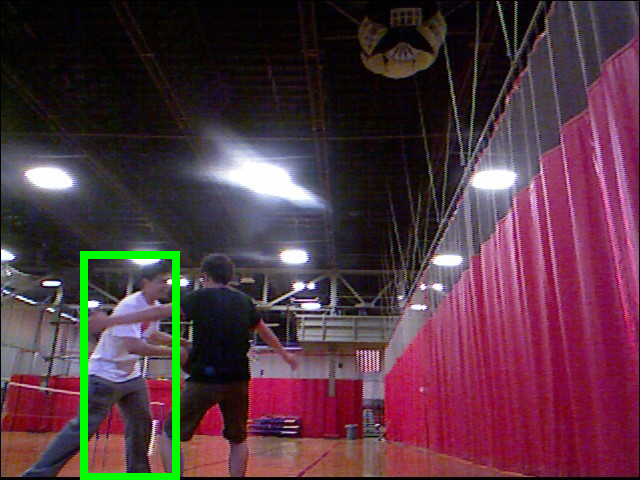}

\vspace{1mm}
\includegraphics[width=0.159\linewidth]{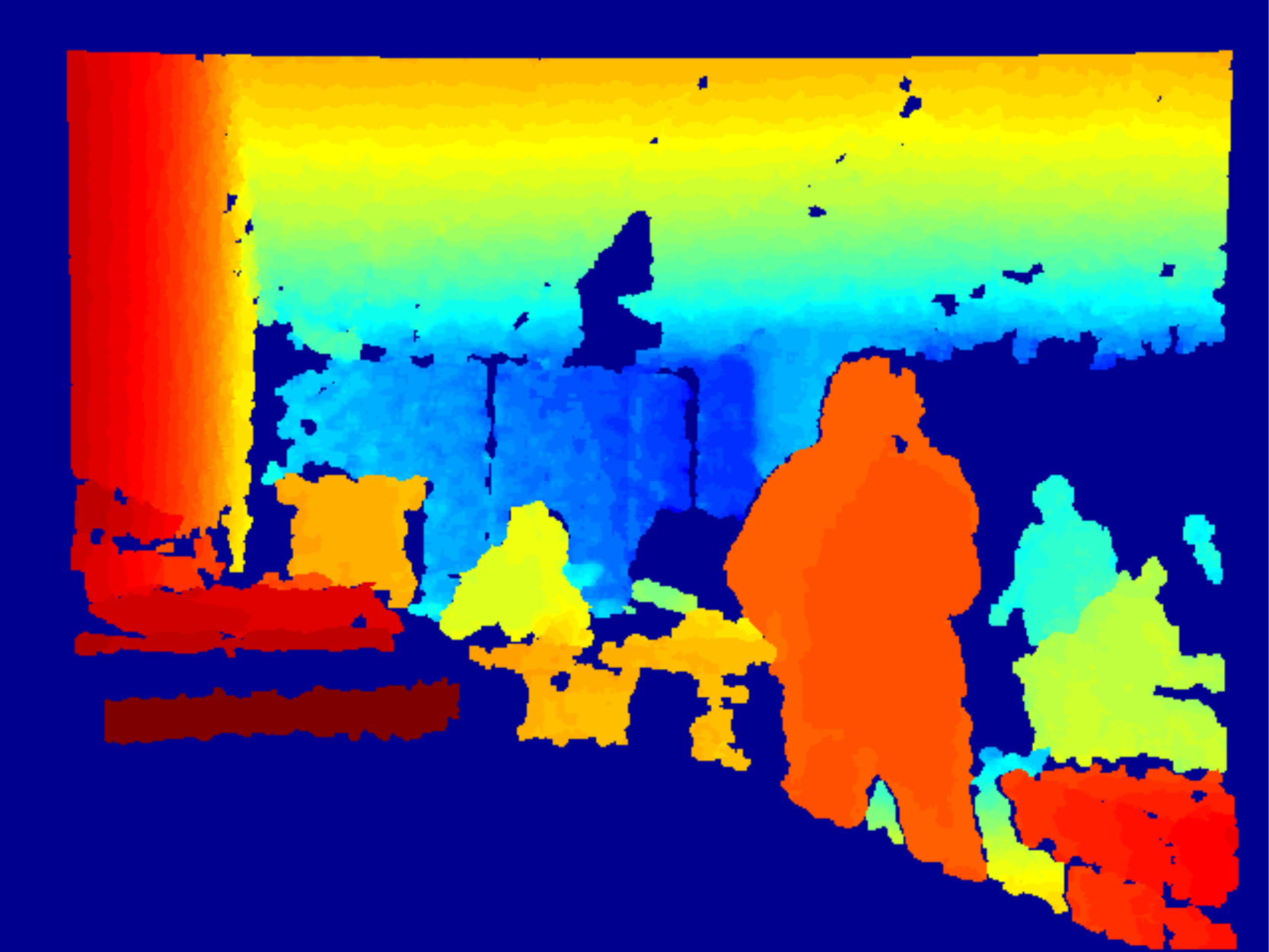}~%
\includegraphics[width=0.159\linewidth]{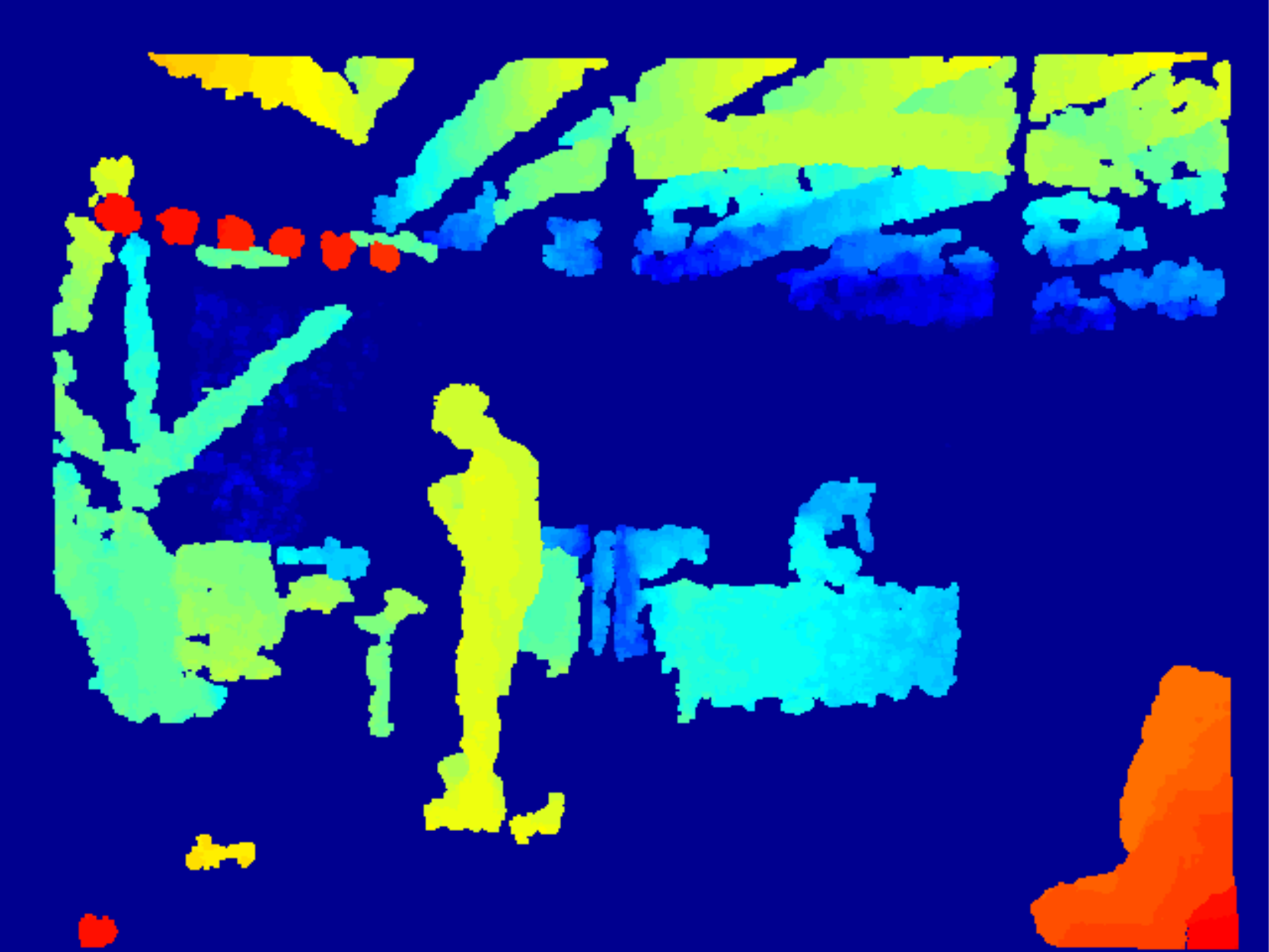}~%
\includegraphics[width=0.159\linewidth]{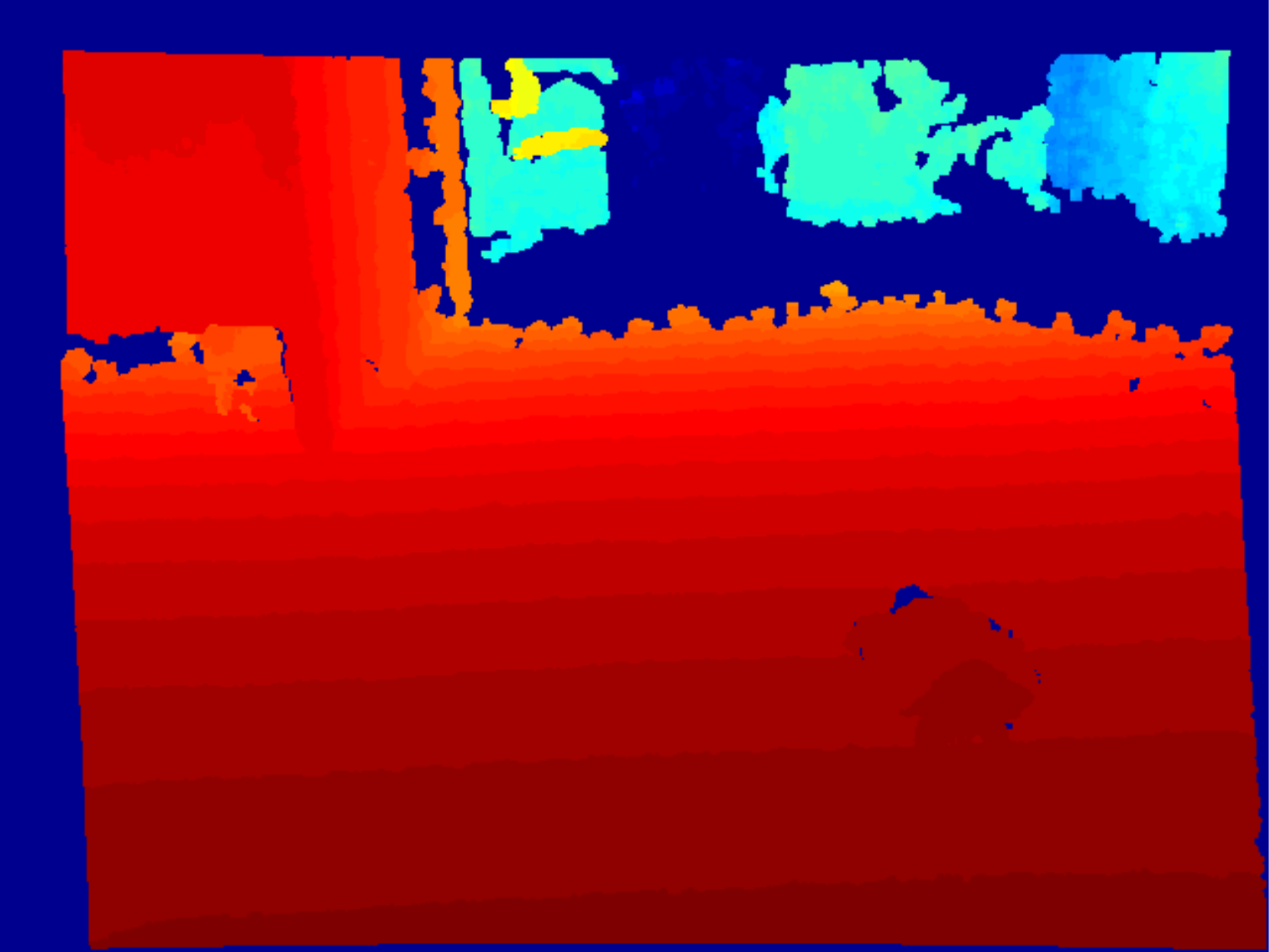}~%
\includegraphics[width=0.159\linewidth]{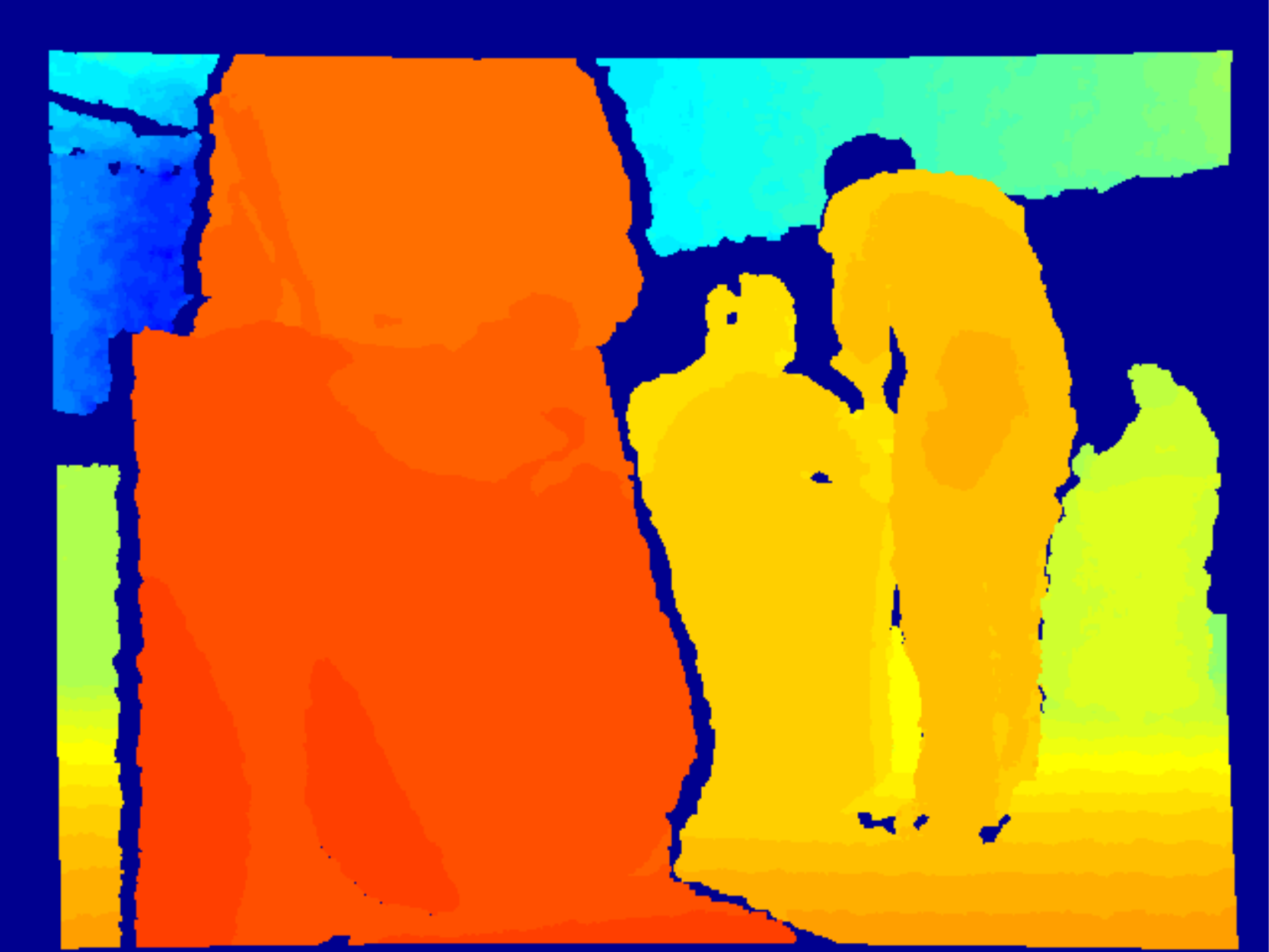}~%
\includegraphics[width=0.159\linewidth]{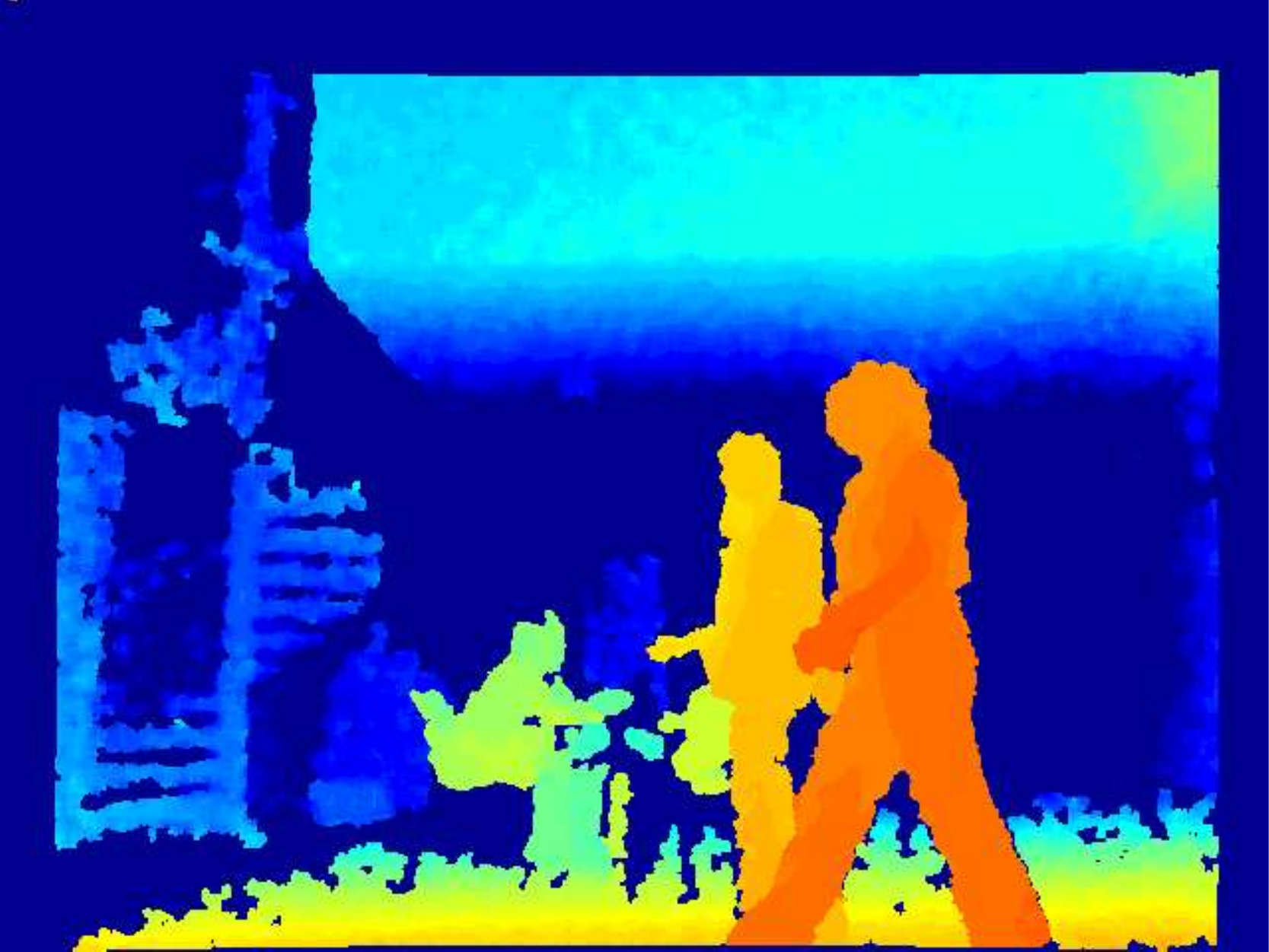}~%
\includegraphics[width=0.159\linewidth]{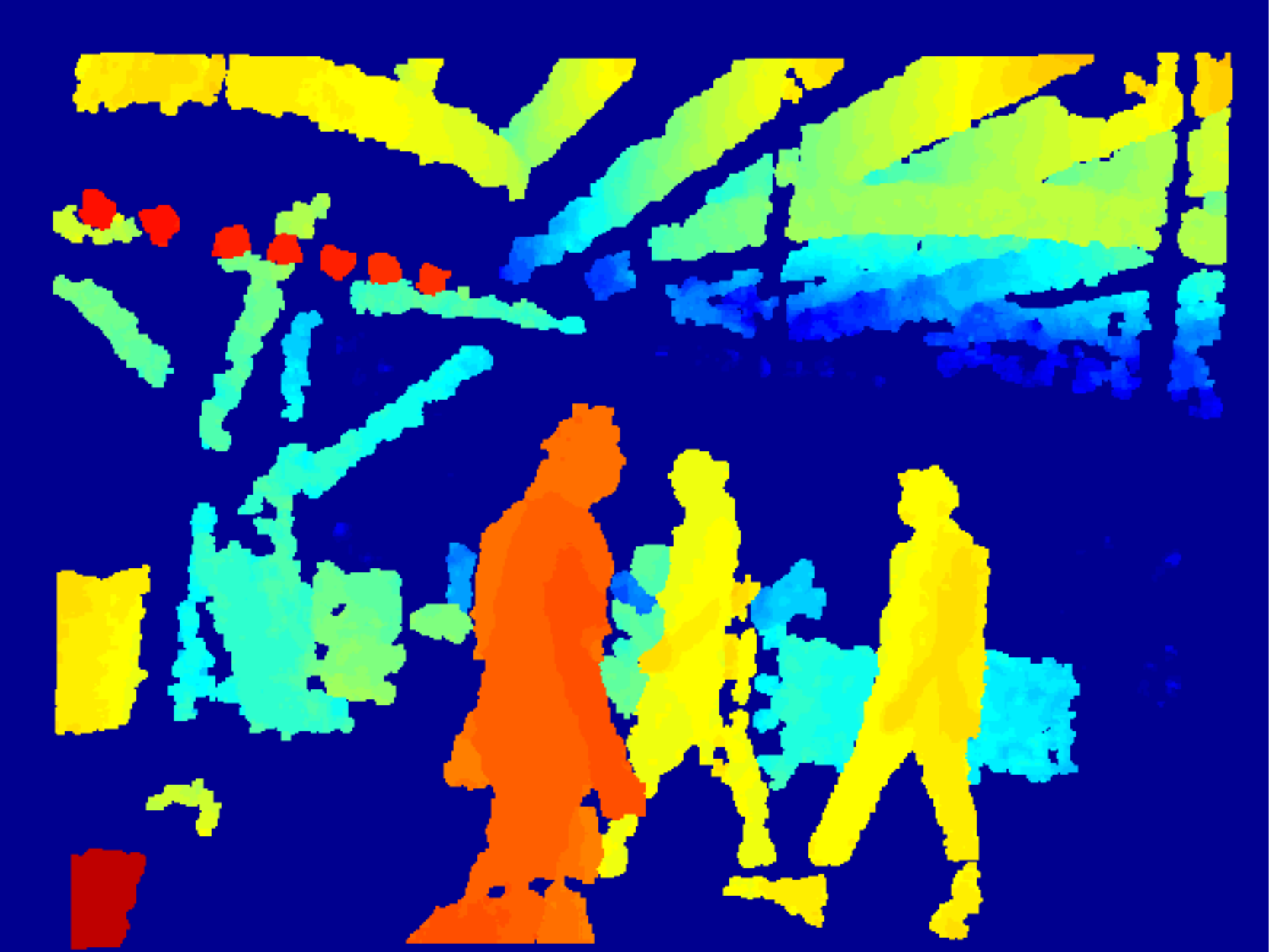}

\vspace{1mm}
\includegraphics[width=0.159\linewidth]{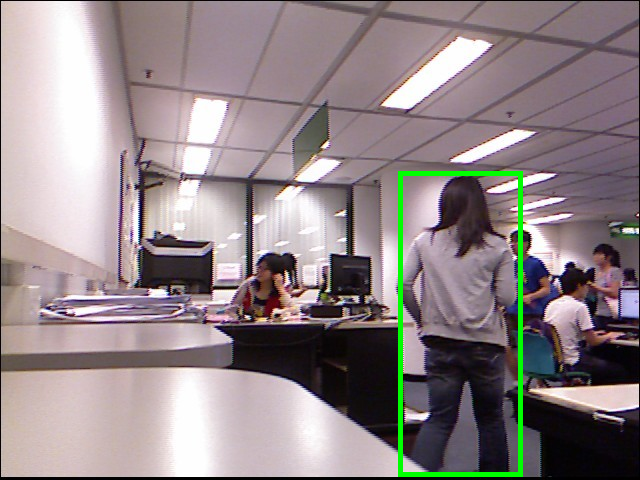}~%
\includegraphics[width=0.159\linewidth]{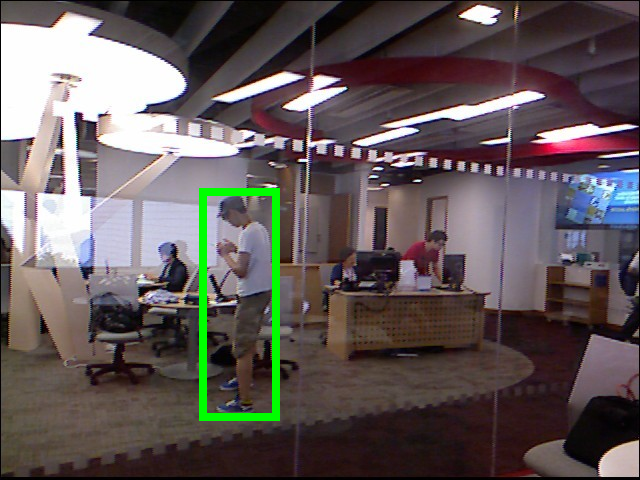}~%
\includegraphics[width=0.159\linewidth]{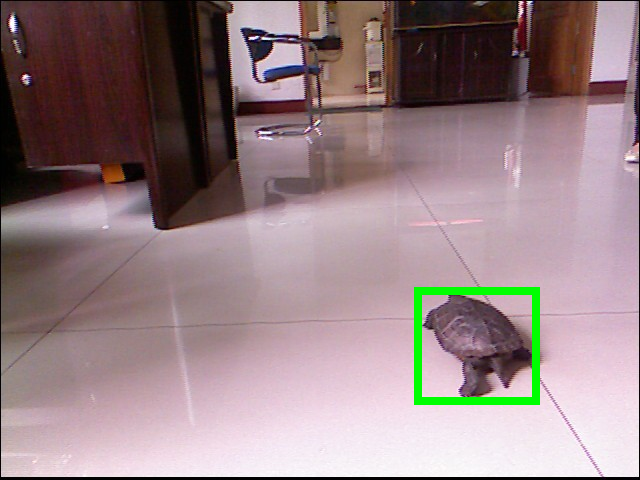}~%
\includegraphics[width=0.159\linewidth]{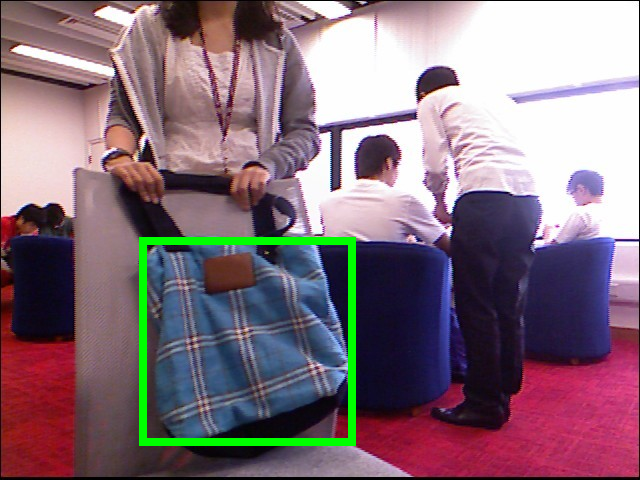}~%
\includegraphics[width=0.159\linewidth]{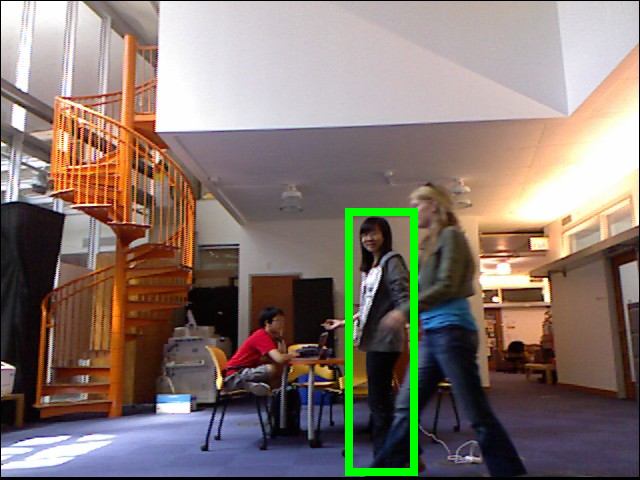}~%
\includegraphics[width=0.159\linewidth]{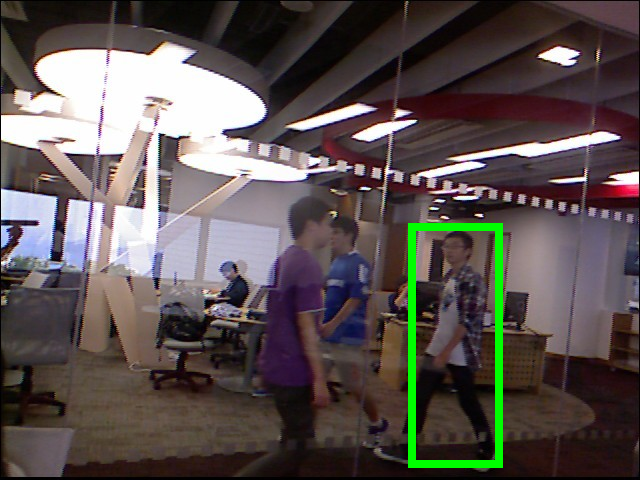}

\vspace{1mm}
\includegraphics[width=0.159\linewidth]{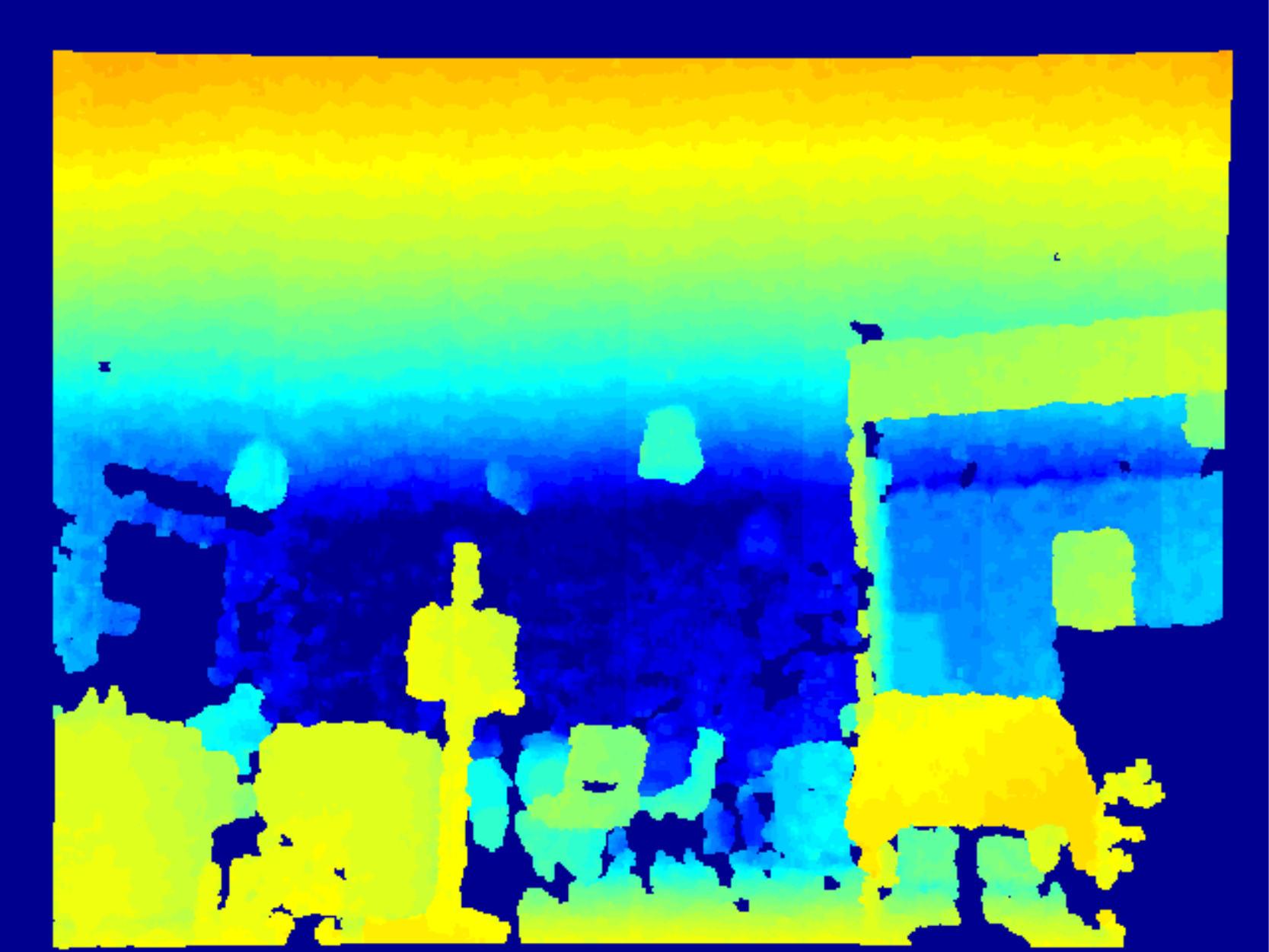}~%
\includegraphics[width=0.159\linewidth]{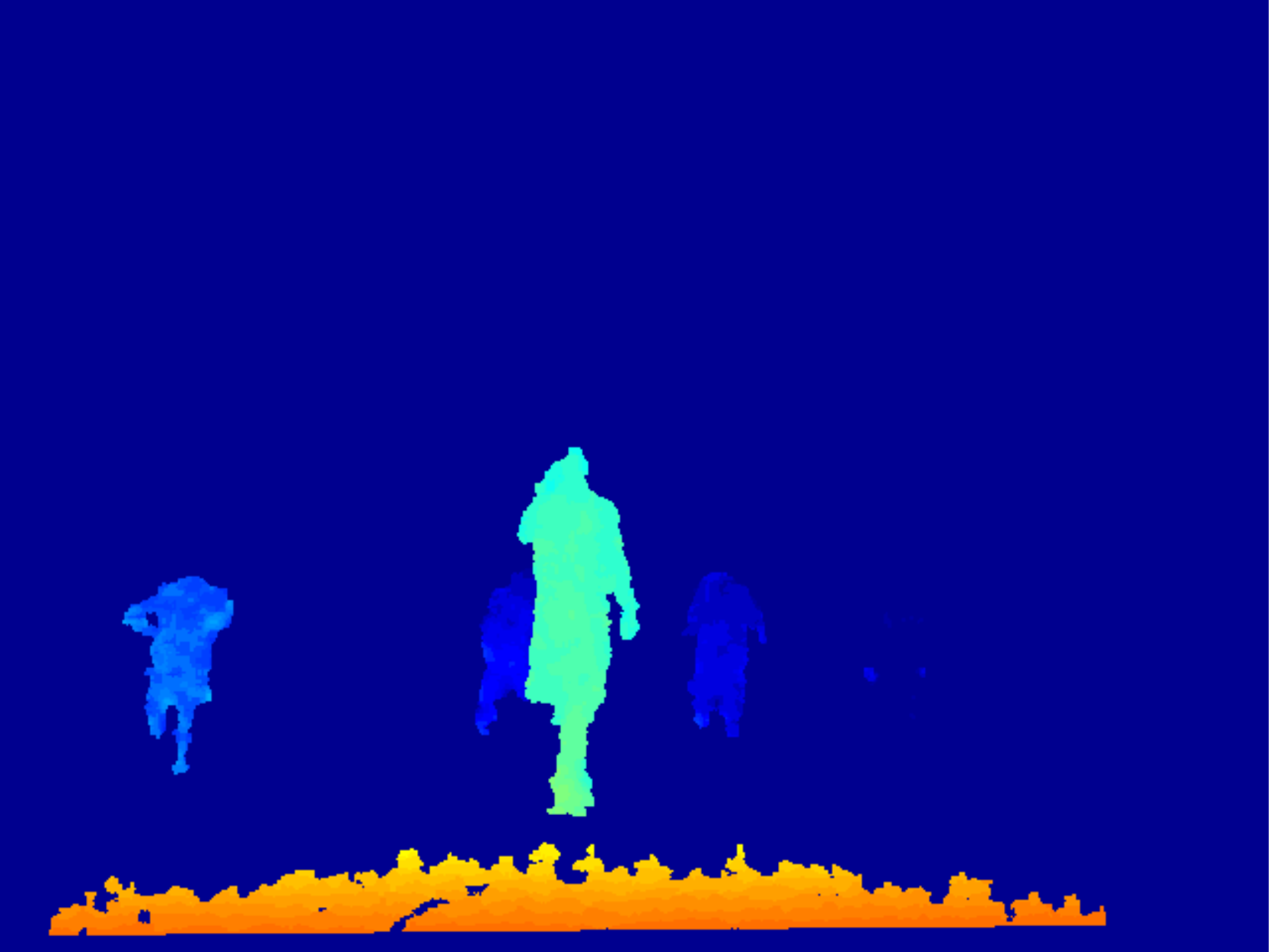}~%
\includegraphics[width=0.159\linewidth]{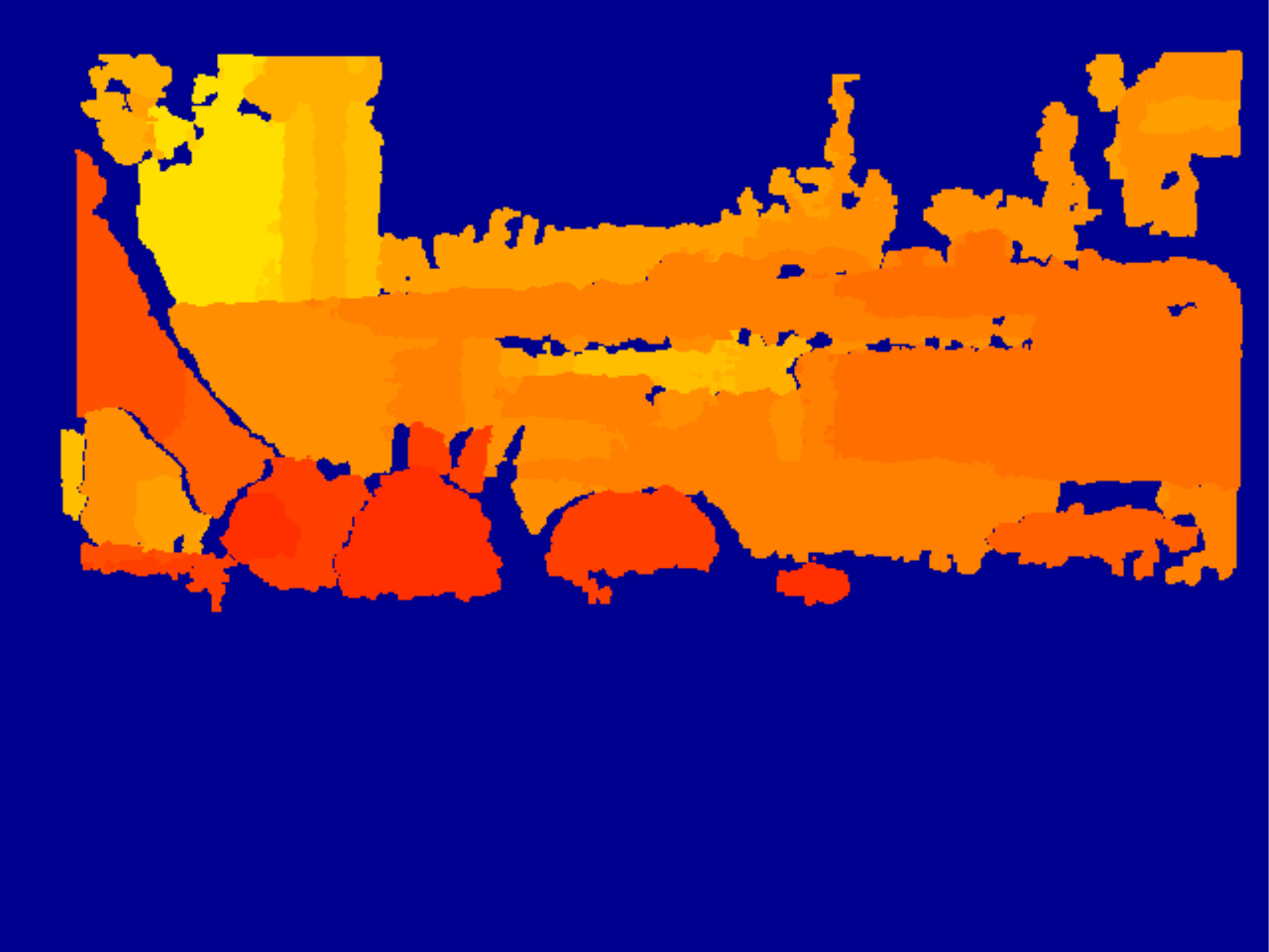}~%
\includegraphics[width=0.159\linewidth]{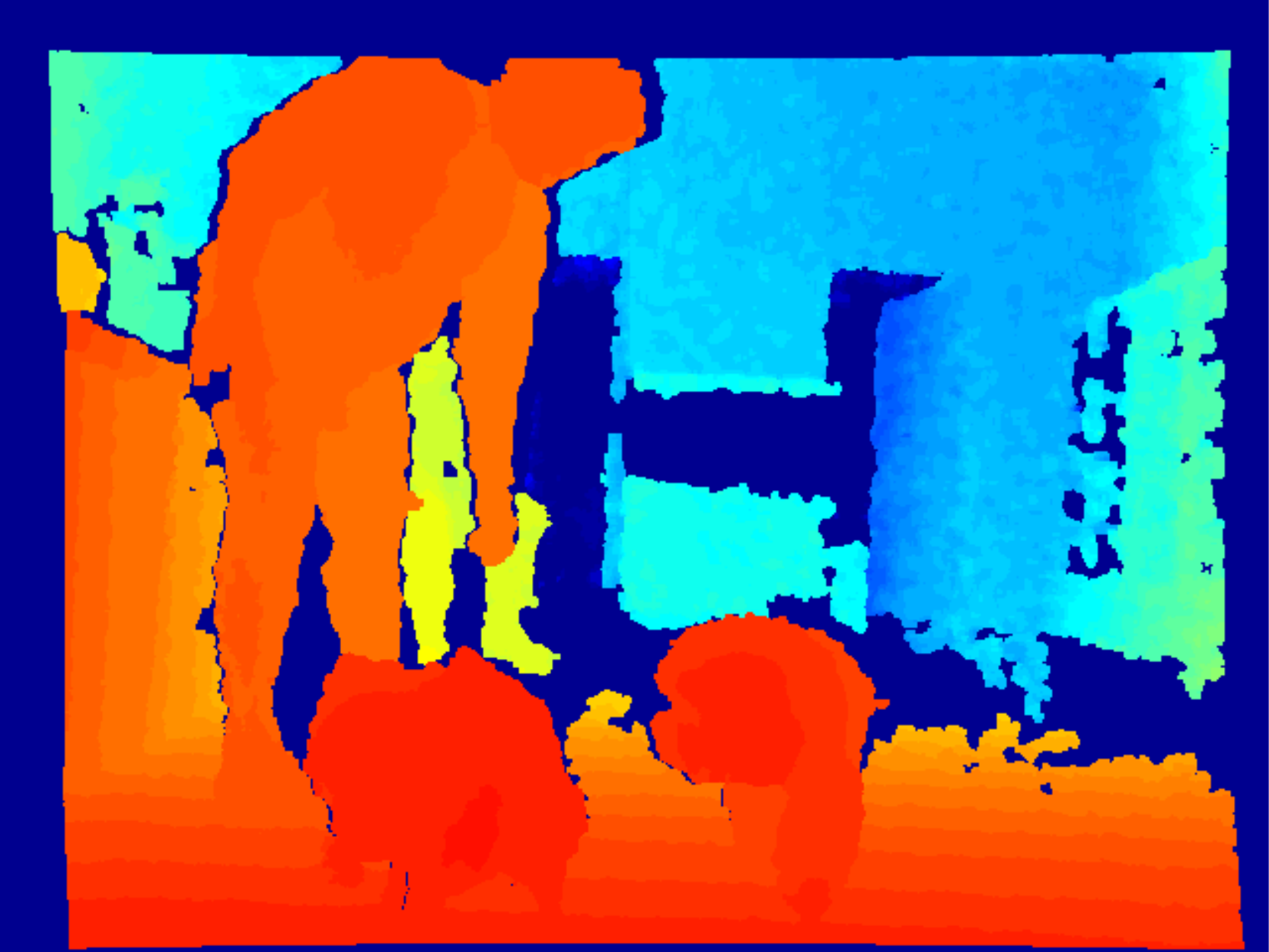}~%
\includegraphics[width=0.159\linewidth]{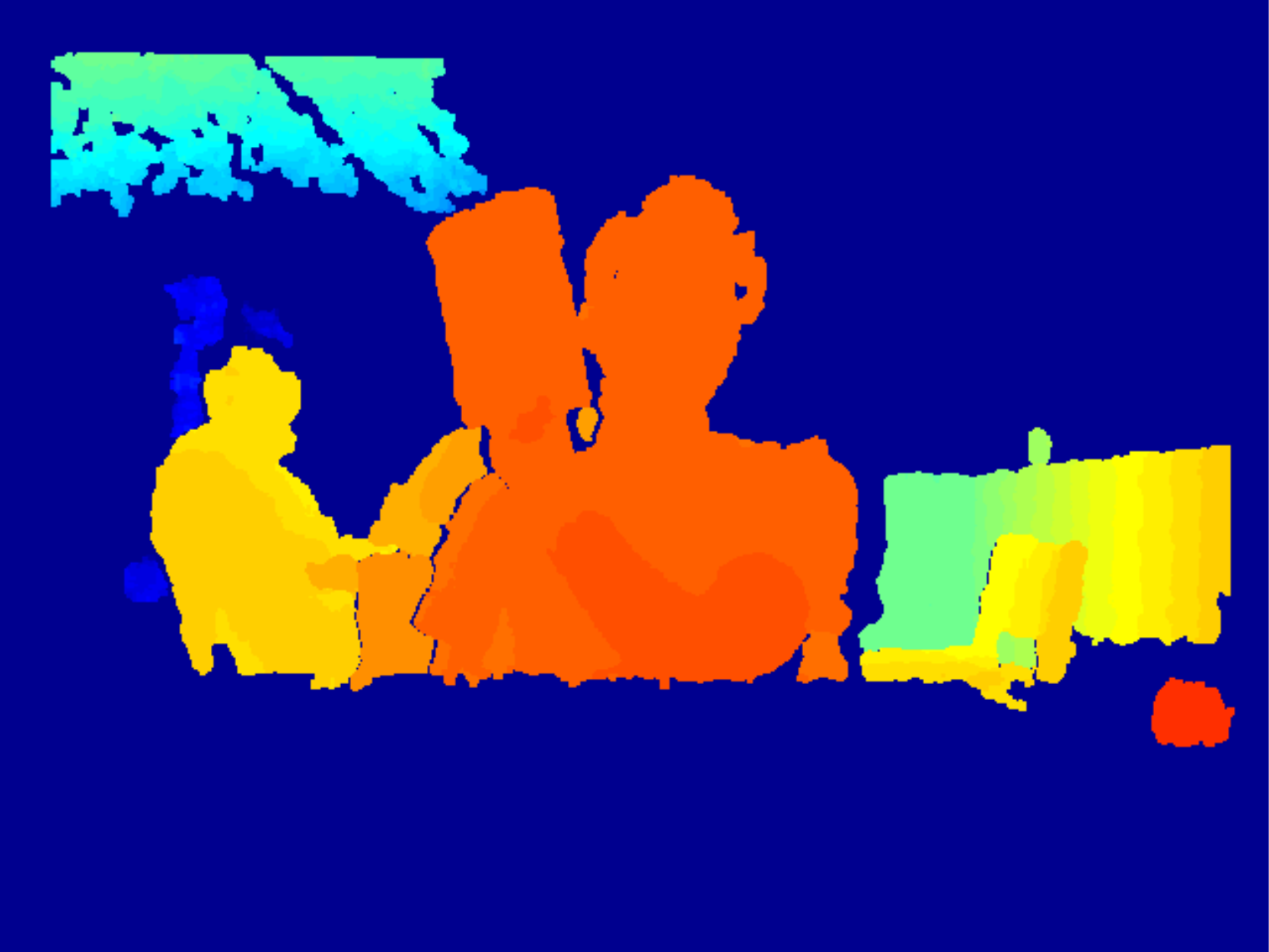}~%
\includegraphics[width=0.159\linewidth]{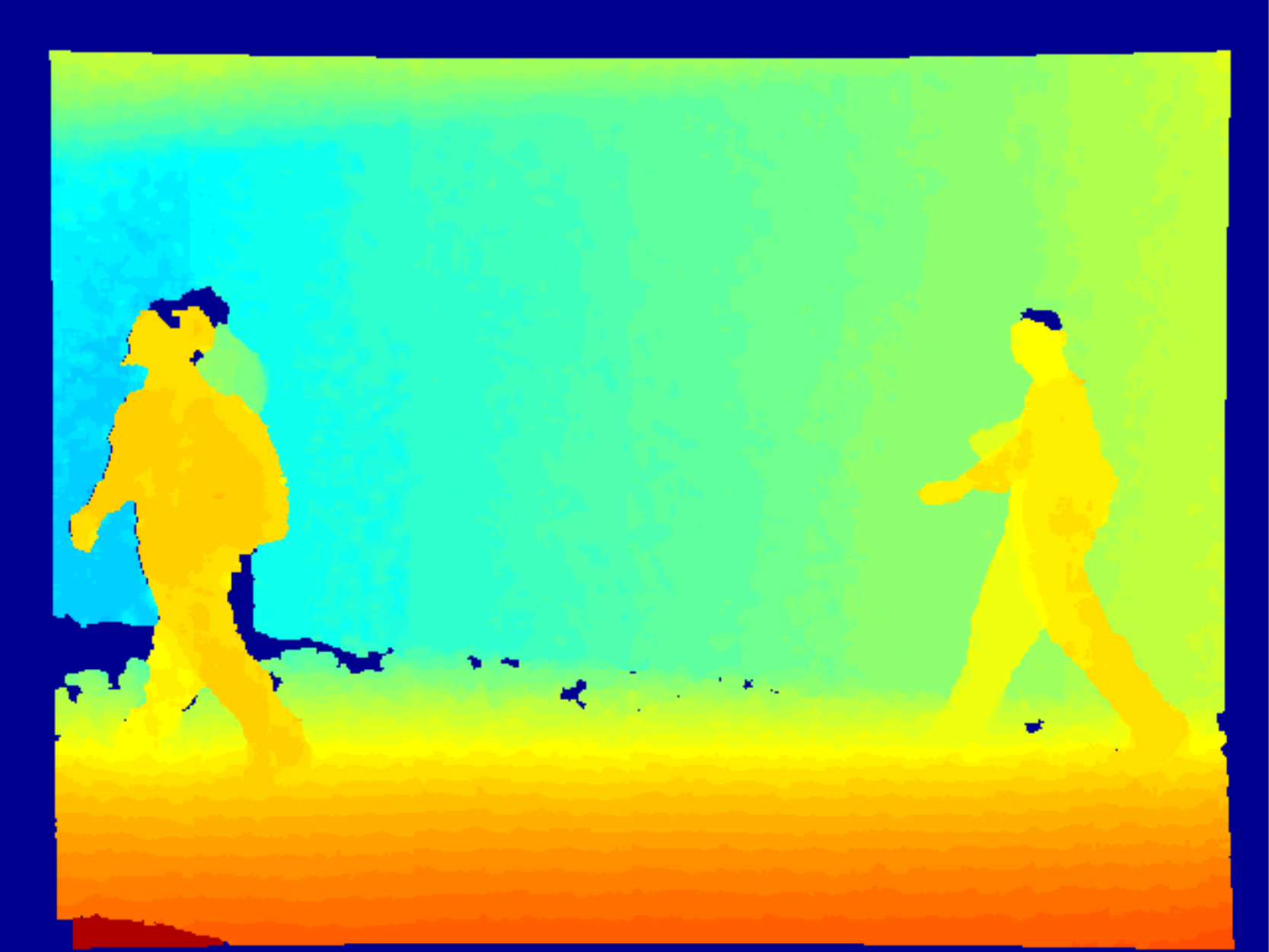}

\vspace{1mm}
\includegraphics[width=0.159\linewidth]{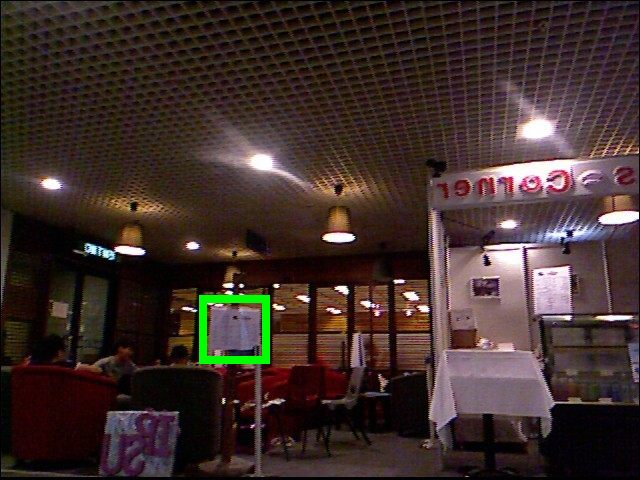}~%
\includegraphics[width=0.159\linewidth]{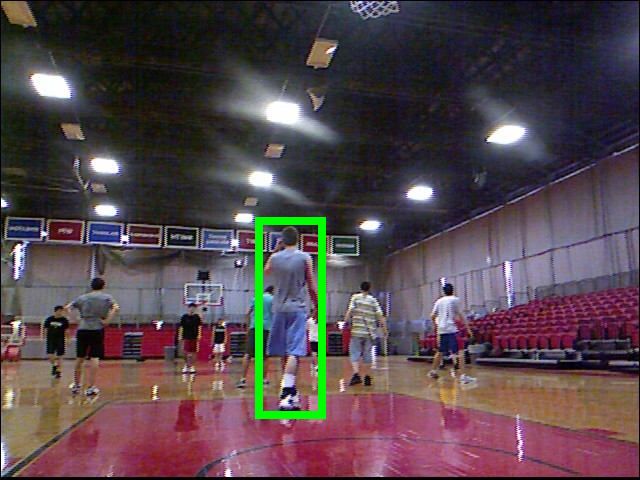}~%
\includegraphics[width=0.159\linewidth]{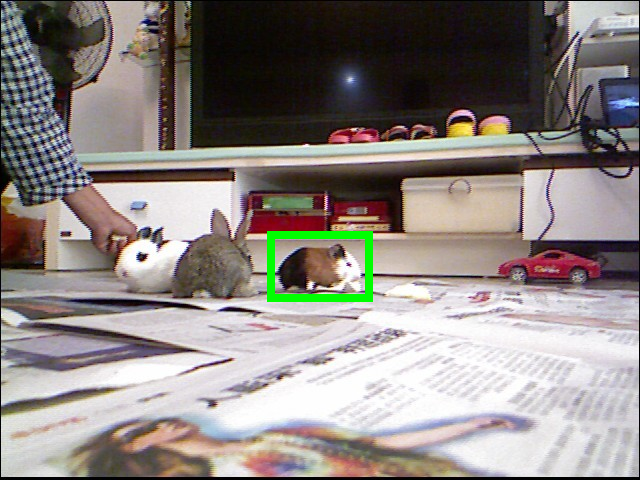}~%
\includegraphics[width=0.159\linewidth]{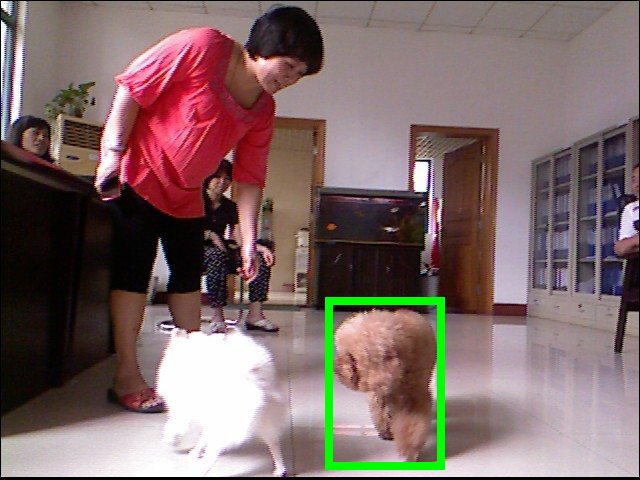}~%
\includegraphics[width=0.159\linewidth]{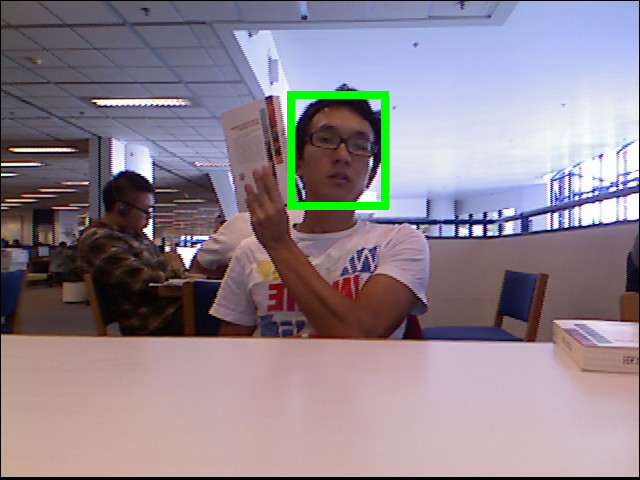}~%
\includegraphics[width=0.159\linewidth]{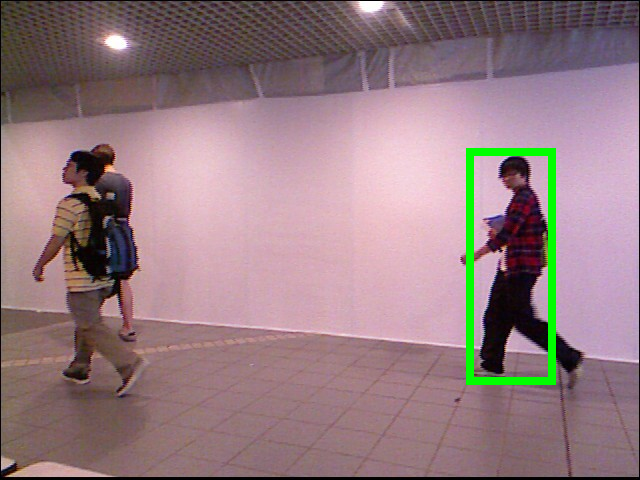}

\caption{Examples of our RGBD tracking benchmark dataset with manual annotation of all frames. }
\label{fig:dataset}
\end{figure}

\begin{figure*}[t]
\centering
\includegraphics[width=0.97\linewidth]{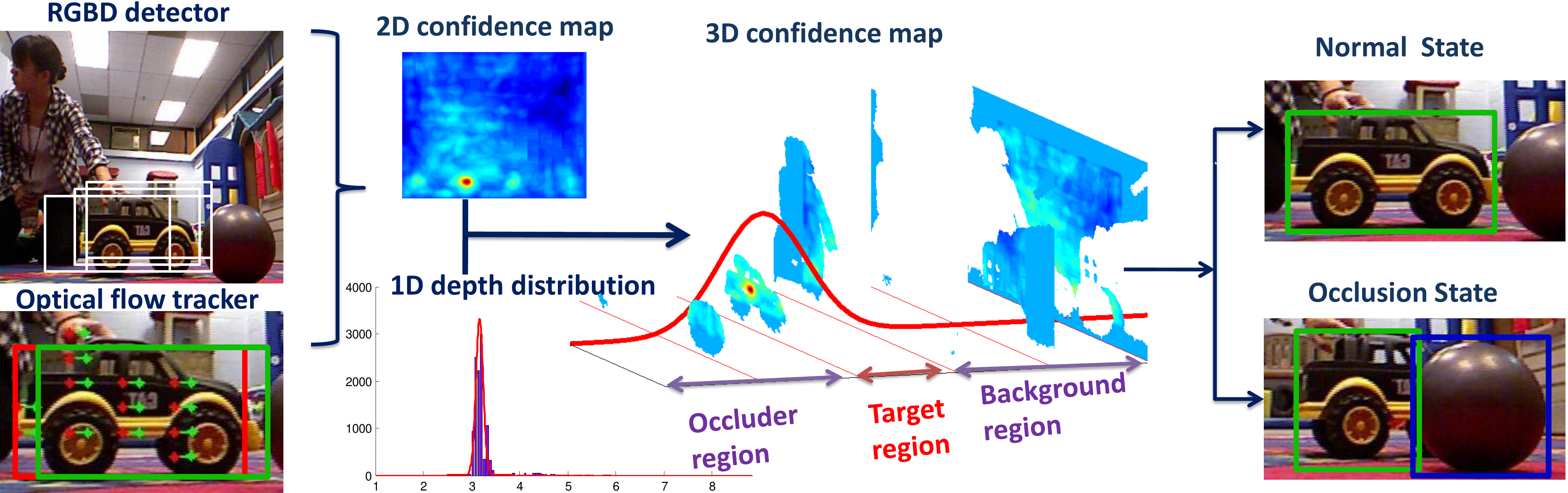}
\caption{ Illustration of our baseline RGBD tracking algorithm. The 2D confidence map is the combined confidence map from classifier and optical flow tracker. The 1D depth distribution is a Gaussian estimated from target depth histogram. A 3D confidence map is computed by applying threshold from the 1D Gaussian on the 2D confidence map. In the output, the target location (the green bounding boxes) is position of the highest confidence. Occluder is recognized from its depth value.}
\label{overview}
%\vspace{-3mm}
\end{figure*}

How much does depth information help in tracking?
What is the baseline performance for tracking given the depth information?
How far are we from claiming that we have solved the tracking problem if we have reliable depth accessible?
What is a reasonable baseline algorithm for tracking with RGBD data, and how
do the state-of-the-art RGB tracking algorithms perform compared with this RGBD baseline?

This paper seeks to answer these questions by proposing a very simple but powerful baseline algorithm and conducting a quantitative benchmark evaluation.
To build a reasonable baseline,
we use the state-of-the-art HOG features \cite{HOG,DPM} 
sliding window detection with linear SVM \cite{SVM,PrimalSVM},
which incorporate depth information to prevent model drift, %%%%%
and robust optical flow \cite{LargeFlow},
and propose a very simple model to represent the depth distribution for occlusion handling.

To evaluate the algorithms,
we construct a large RGBD video dataset of 100 videos with high diversity,
including deformable objects, various occlusion conditions, and moving cameras, under different lighting conditions and in different scenes (Figure \ref{fig:dataset}).
We aim to lay the foundation for further research in this task, for both RGB and RGBD tracking approaches, by providing a good benchmark and baseline.
We will withhold the ground truth annotation for a portion of the dataset,
provide instructions for submitting new models,
and host an online evaluation server to allow public submission of the results from new models.

\subsection{Related works}

There are many noteworthy tracking algorithms which have been proposed in the last decade.
Here we briefly summarize only a partial list of them, due to space constraints.
\cite{mil} proposes a very robust system with online multiple instance learning. 
\cite{tld} designs a framework to integrate tracking, learning, and detection using P-N loops.
\cite{semiB} uses semi-supervised online boosting to increase tracking robustness.
\cite{EigenTracking} learns a view-based representation to account for object articulations, 
while \cite{Adam06robustfragments-based} handles it using a fragments-based model.
To address target appearance changes, 
\cite{robustonline} uses a Gaussian Mixture Model built from online expectation maximization (EM), 
and \cite{Ross08incrementallearning} presents an incremental subspace learning algorithm. 
More recently, \cite{ct} proposes using compressive sensing for real-time tracking, 
and \cite{DBLP:conf/iccv/HareST11} presents structured output prediction to avoid intermediate classification.
There are also some important works on multiple target tracking and motion flow estimation, such as \cite{dahua,xiaogang}.

There has been also some seminal works on tracking using RGBD cameras \cite{luber11learning,spinelloIROS11,luberIROS11,spinelloICRA11}, but they all focus on tracking the human body.
The publicly available RGBD People Dataset \cite{spinelloIROS11,luberIROS11}
contains only one sequence with 1132 frames captured with static cameras with only people moving,
which is obviously not enough to evaluate tracking algorithms for general objects.

There has been several great benchmarks for various computer vision tasks that help to advance the field 
and shape computer vision as a rigorous experimental science,
\eg two-view stereo matching benchmark \cite{stereoBenchmark},
multiple-view stereo reconstruction benchmark \cite{mviewBenchmark},
optical flow benchmark \cite{opticalFlowBenchmark},
Markov Random Field energy optimizaiton benchmark \cite{MRFBenchmark},
object classification, detection, and segmentation benchmark \cite{voc},
scene classification benchmark \cite{SUNDB} and 
large scale image classification benchmark \cite{ImageNet}.
This paper is an addition to the list to provide a benchmark of tracking,
for both RGB and RGBD video.

\section{Baseline algorithm}

The goal is to build a simple but strong baseline algorithm leveraging state-of-the-art feature, detection and optical flow algorithms with simple but reasonable occlusion handling. An overview of the baseline tracking algorithm is shown in Figure \ref{overview}. 

\subsection{Detection and optical flow}

\begin{figure}[t]
\subfigure[HOG of RGB]{\includegraphics[width=0.33\linewidth]{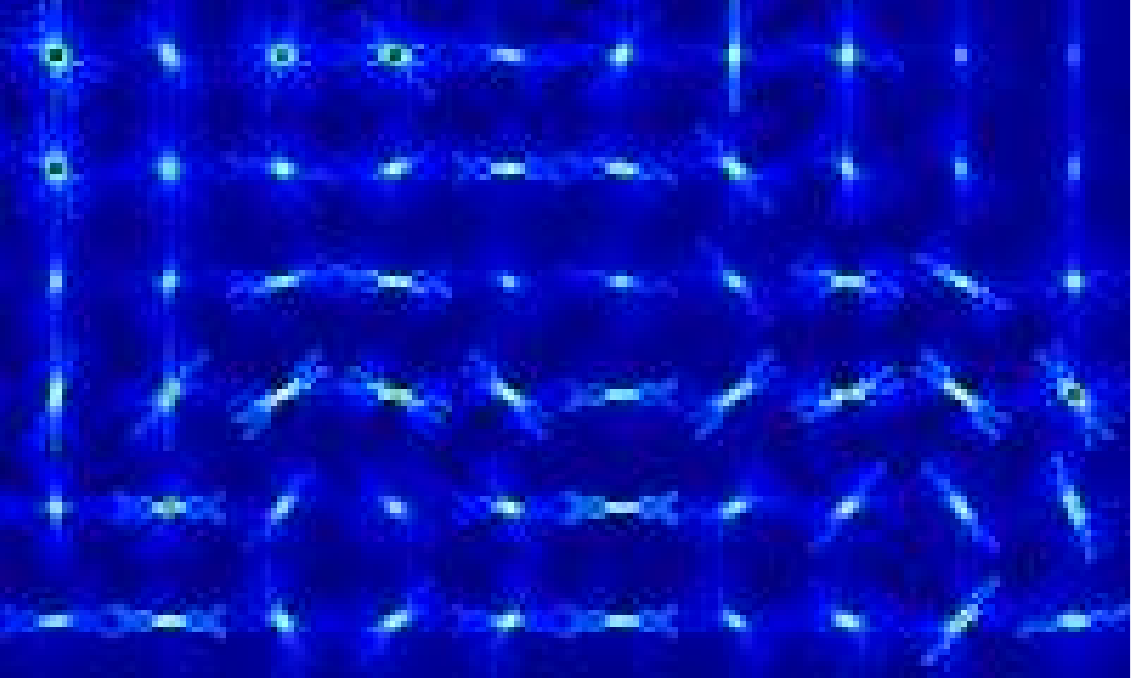}}~%
\subfigure[RGB image]{\includegraphics[width=0.33\linewidth]{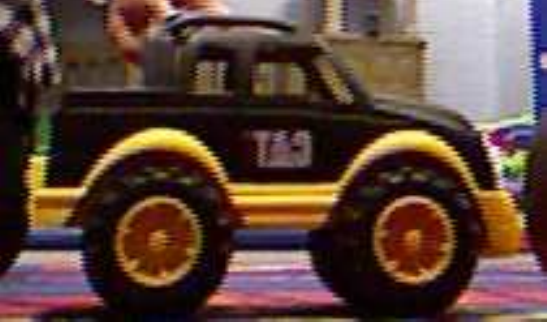}}~%
\subfigure[HOG in Depth]{\includegraphics[width=0.33\linewidth]{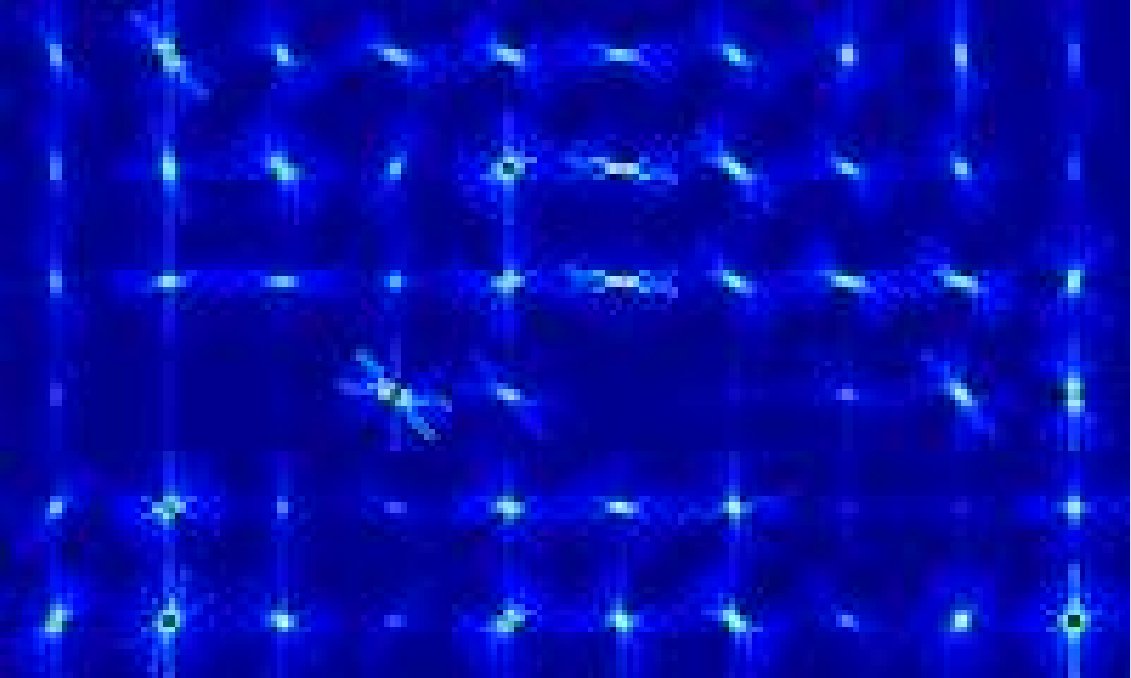}}
\caption{The features we used for our baseline algorithm.}
\label{rhogdhog2}
%\vspace{-3mm}
\end{figure}

Our baseline algorithm includes a linear support vector machine classifier (SVM \cite{SVM,PrimalSVM}) based on RGBD features, an optical flow tracker \cite{LargeFlow} and a target depth distribution model, which are initialized by the input bounding box from the first frame and updated online. The RGBD feature we used is histogram of oriented gradients (HOG\cite{HOG,DPM}) from both RGB and depth data (Figure \ref{rhogdhog2}). HOG for depth is obtained by treating depth data as a gray scale image. This RGBD HOG feature describes local textures as well as 3D shapes, in which the target is more separable from background as well as occluder, and therefore improves the robustness against model drifting, especially when there is illumination variation, lack of texture, or high similarity between target and background color. 

In the subsequent frame, a HOG pyramid is computed, and a sliding window is run using a convolution of the SVM weights, which returns several possible target locations with their confidence. Confidence of these locations are then adjusted according to the bounding box estimated from optical flow tracker \cite{LargeFlow} in the following way:
\begin{equation} 
\label{eq:finalconf}
c = c_{d} + \alpha c_{t} r_{(t,d)}
\end{equation}
in which $c_{d}$ is the confidence of detection, $c_t$ is the confidence of optical flow tracking, and $r_{(t,d)}$ is the ratio of overlap between the detection and optical flow tracker's resulting bounding boxes, \ie an indication of their consistency. 
$\alpha$ denotes the weight of the overlap ratio ($\alpha = 0.5$ in our experiment). 
After this step, the target depth distribution model, a Gaussian distribution learned from previous frames, discards bounding boxes far from estimated target depth. The most probable remaining bounding box is picked and re-centered towards the center of the nearby region whose depth agrees with target depth model. Such re-centering helps prevent drifting of output bounding boxes.
Afterwards, the target models are updated using this bounding box with hard negative mining and the tracker proceeds to the next frame.

\subsection{Occlusion handling}

In order to handle occlusions, some traditional RGB trackers like \cite{tld} use forward-backward error to indicate tracking failure caused by occlusion, and some others like \cite{Adam06robustfragments-based,mil} use a fragment-based model to reduce the models' sensitivity to partial occlusion. However, with depth information the solution for this issue becomes more straight-forward. 
Here we propose a simple but effective occlusion handler which actively detects the target occlusion and recovery during tracking process.

\begin{figure}[t]
\centering

\includegraphics[width=0.65\linewidth]{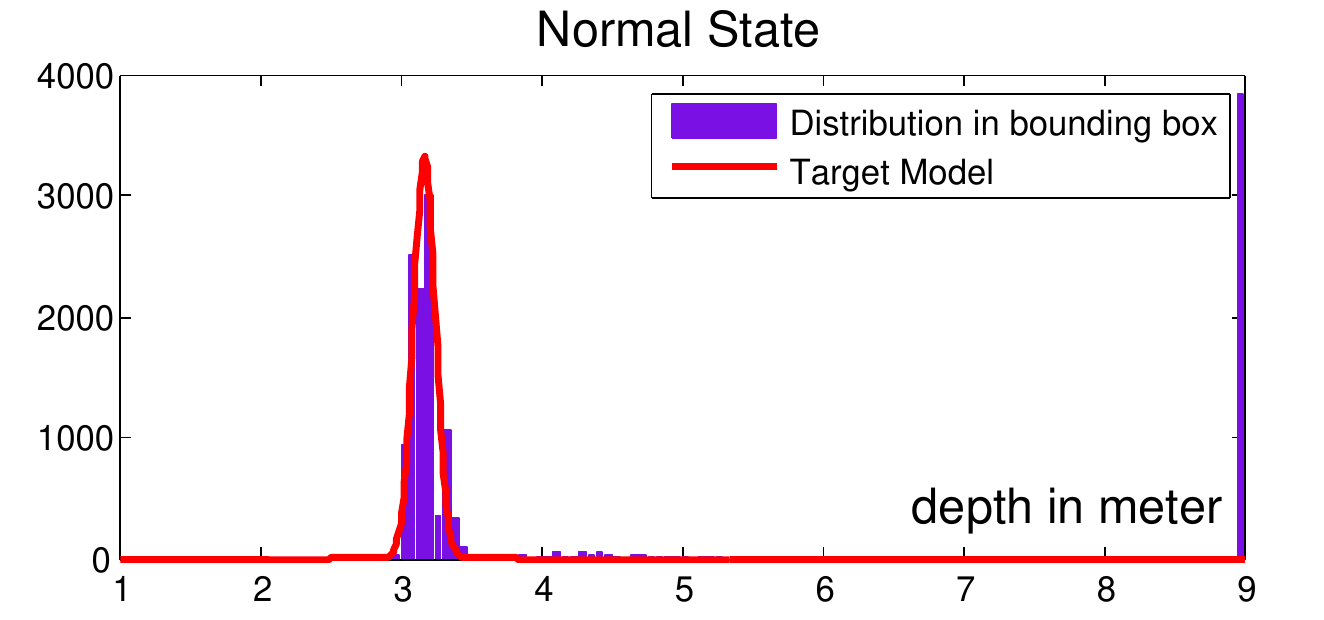}~\includegraphics[height=0.29\linewidth]{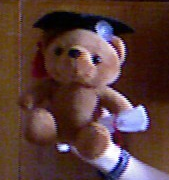}

\includegraphics[width=0.65\linewidth]{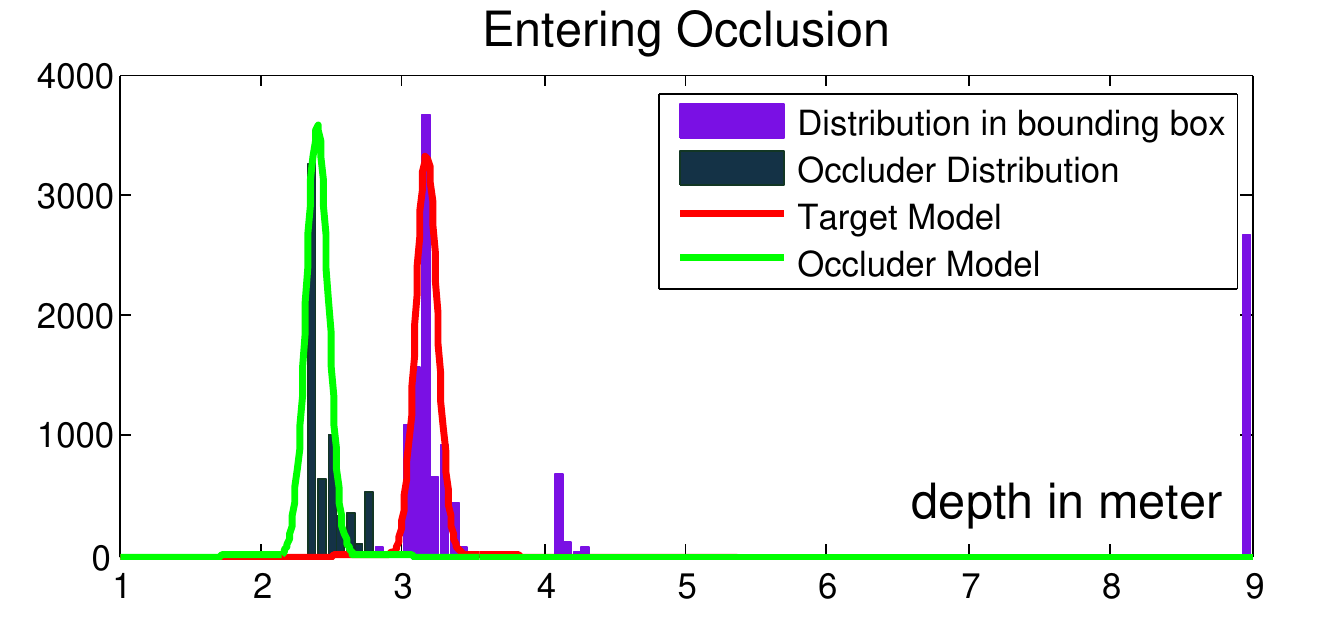}~\includegraphics[height=0.29\linewidth]{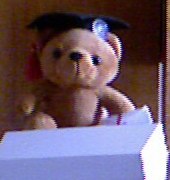}

\caption{Depth distribution inside the bounding box. The top row shows the distribution in normal state, and the bottom row shows the distribution when occlusion occurs. The red Gaussian denotes the target model, and the green denotes the occluder model.}
\label{occ_hist}
\vspace{-3mm}
\end{figure}

\paragraph{Occlusion detection}
We assume that the target is the closest object that dominates the bounding box when not occluded. 
A new occluder in front of the target inside the bounding box
indicates the beginning of occlusion state. 
Therefore, depth histogram inside bounding box is expected to have a newly rising peak with a smaller depth value than target, and/or a reduction in the size of bins around the target depth, as illustrated in Figure \ref{occ_hist}.

The depth histogram $h_i$ of all pixels inside a bounding box can
be approximated as a Gaussian distribution for the $i$-th frame:
\begin{equation}
\label{eq:targetdistribution}
h_i \sim \mathcal{N} (\mu_i,\sigma_i^2).
\end{equation}
%where $\mu_i$ and $\sigma_i$ are estimated from the histogram.
And we define the likelihood of occlusion for this frame as:
\begin{equation}
\label{eq:checkocc}
O_i = \frac{\sum\limits_{d=0}^{\mu_i-\sigma_i} h_i(d)}{\sum\limits_{d} h_i(d)},
\end{equation}
where 
$h_i(d)$ is the count in the $d$-th bin for the $i$-th frame, 
and $d=0$ is the depth of the camera.
$\mu_i-\sigma_i$ is a threshold for a point to be considered as occluder. 
The number of pixels that have smaller depth value than target depth is considered the area of the occluder that has appeared in the bounding box. 
Hence, a larger $O_i$ indicates that an occlusion is more likely. The target depth value is updated online, so a target moving towards the camera will not be treated as an occlusion.

\begin{figure}[t]
\subfigure[]{\includegraphics[width=0.24\linewidth]{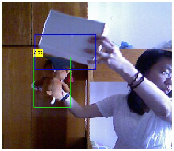}}~%
\subfigure[]{\includegraphics[width=0.24\linewidth]{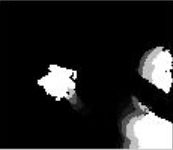}}~%
\subfigure[]{\includegraphics[width=0.24\linewidth]{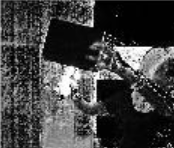}}~%
\subfigure[]{\includegraphics[width=0.24\linewidth]{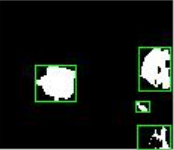}}
\vspace{-1mm}
\caption{Local search for target candidates by segmentation. (a) RGB image (the target location is indicated by the green bounding box, the occluder indicated by the blue bounding box) (b) depth segmentation (c) RGB segmentation (d) final segmentation result along with possible target candidates.}
\label{recover}
%\vspace{-3mm}
\end{figure}

\paragraph{Under occlusion}
Our occlusion model, \ie the occluder's depth distribution, is initialized when entering the occlusion state. 
In the following frames, the occluder's position is updated by the optical flow tracker. %%
A list of possible target candidates are identified either by the RGBD detection or a local search around the occluder. With color and depth distributions of target and occluder, the local search is done by performing segmentation on RGB and depth data respectively and combining their results. The combined segmentation produces a list of target candidates (Figure \ref{recover}), whose validity is then judged by the SVM classifier. If there is no candidate in the searching range or all of them are invalid, the tracker just tracks the occluder, preparing for the next frame.

\paragraph{Recovery from occlusion}
By examining the list of possible target candidates the tracker interprets target recovery when at least one candidate from the list satisfies the following condition: (1) the candidate's visible area is large enough compared to target area before entering occlusion, (2) the overlap between the occluder and the candidate is small and (3) the SVM classifier reports a high confidence. The occlusion subroutine ends if the target is recovered from occlusion.

\section{RGBD Tracking Benchmark}

\subsection{Dataset construction}
Several testing sets of RGB videos have been developed to measure the performance of different trackers. However, these datasets do not contain depth information and thus are not suitable for our purpose. In order to evaluate the performance improvement from depth information, we recorded a benchmark dataset consisting of 100 video clips 
with both RGB and depth data, manually annotated to contain the ground truth.

\paragraph{Hardware setup}
Our testing data set is captured using a standard Microsoft Kinect. It uses a paired infrared projector and camera to calculate depth value, thus its performance is severely impaired in an outdoor environment under direct sunlight. 
Also, Kinect requires a minimum and a maximum distance from the object to the cameras in order to obtain accurate depth value. 
Due to the above constraints, our videos are captured indoor, with object depth value ranging mainly from 0.8 to 6 meters.

\begin{figure}[t]
\includegraphics[width=1\linewidth]{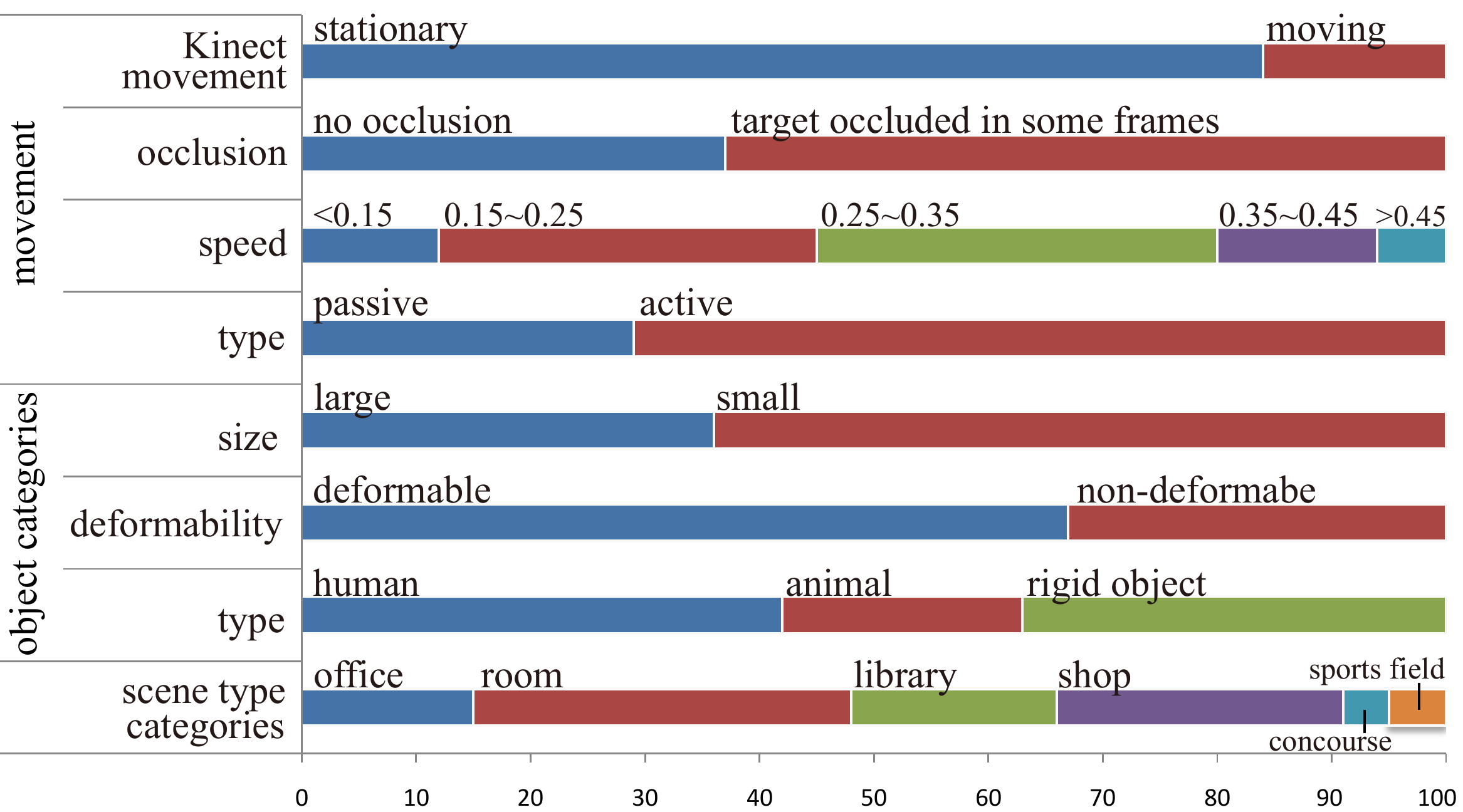}
\caption{Statistics of our RGBD tracking benchmark dataset.}
\label{tab:statistics}
\vspace{-3mm}
\end{figure}

\begin{figure*}[t]
\subfigure[Test cases without occlusion]{\includegraphics[width=0.33\linewidth]{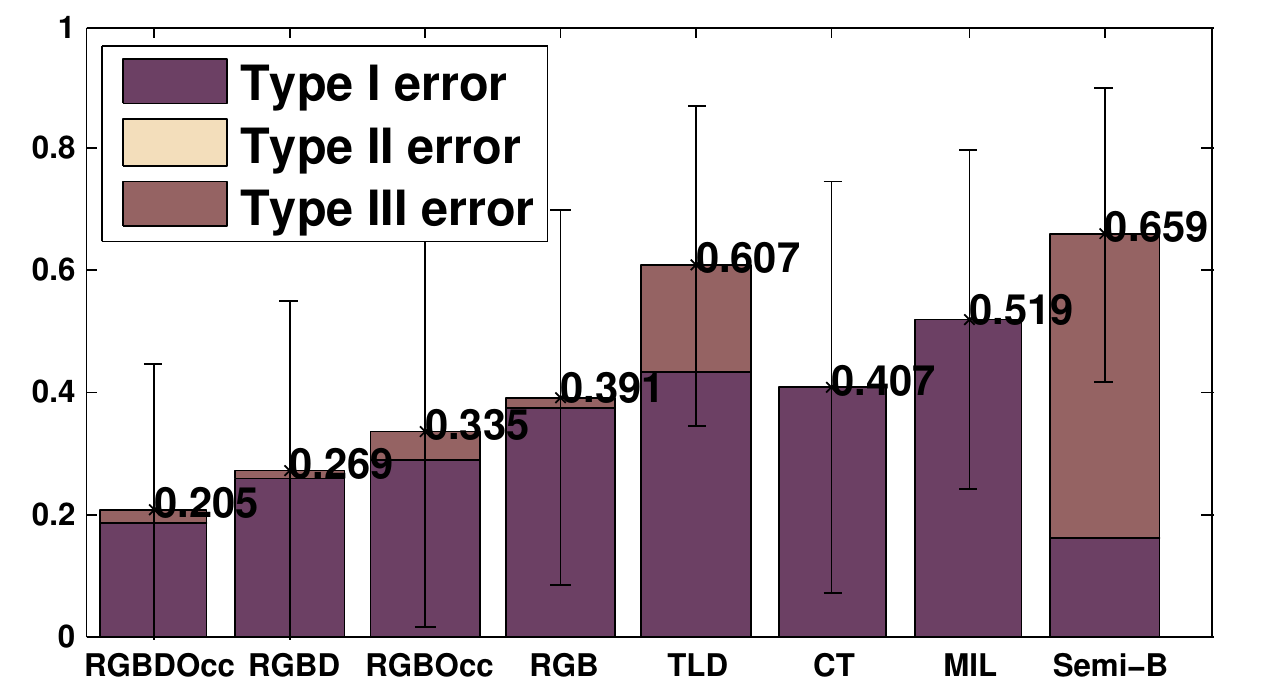}}~%
\subfigure[Test cases with occlusion]{\includegraphics[width=0.33\linewidth]{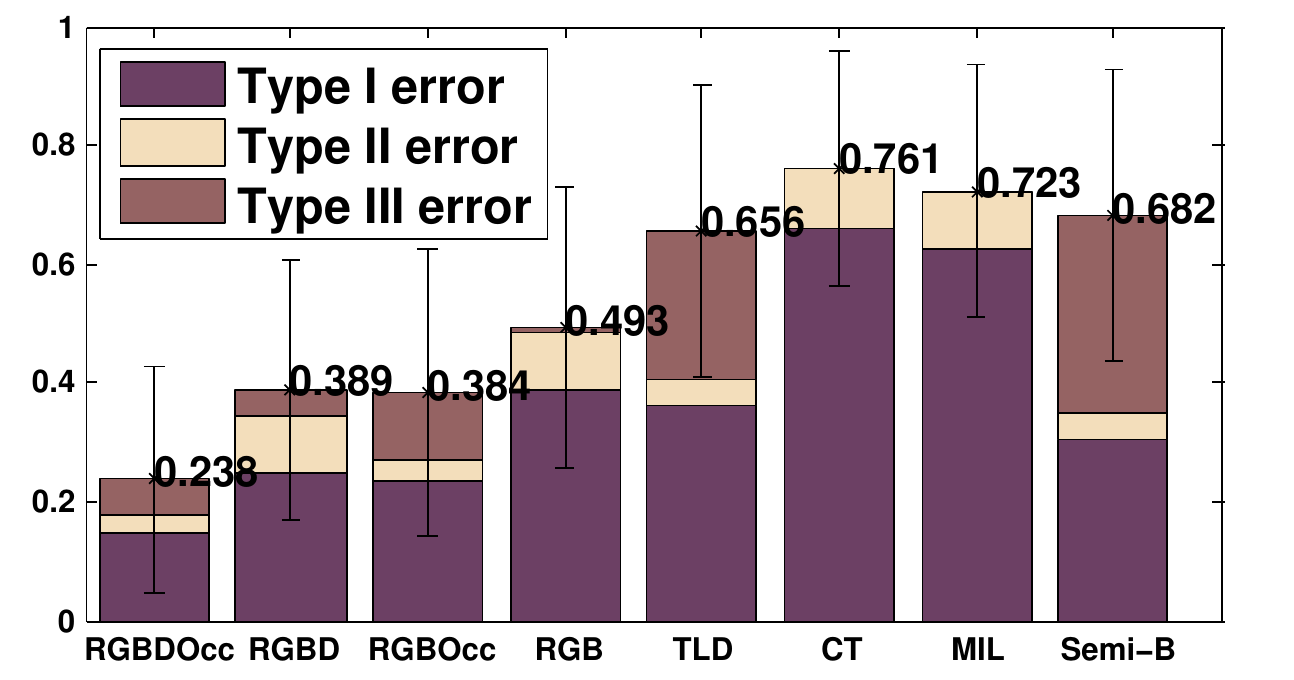}}~%
\subfigure[All test cases]{\includegraphics[width=0.33\linewidth]{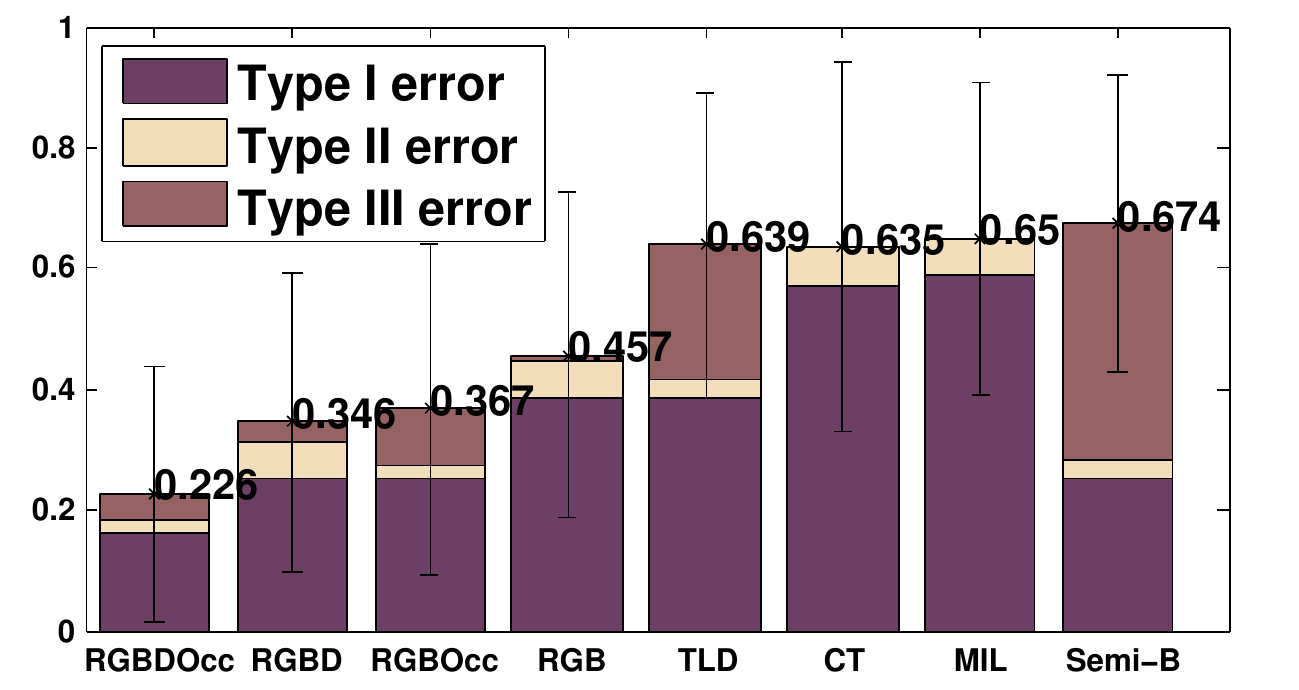}}
\caption{Average error rate composed of three types evaluated on different categories of test cases.}
\label{errRate}
%\vspace{-3mm}
\end{figure*}

\paragraph{Annotation}
We manually annotate the ground truth (target location) of the dataset by drawing bounding box on each frame as follows: 
A minimum bounding box covering the target is initialized on the first frame. On the next frame, if the target moves or its shape changes, the bounding box will be adjusted accordingly; otherwise, it remains the same. 
One author manually annotated all frames, to ensure high consistency.
Because we manually annotate each frame, there is no interpolation or choosing of key frames.
When occlusion occurs, the ground truth is defined as the minimum bounding box covering only the visible portion of the target. 
For example, if a person is occluded and so that only his/her left arm can be seen, then we provide the bounding box of the left arm instead of a predicted position of the whole human body. 
When the target is completely occluded there will be no bounding box for this frame. The same labeling criteria is also used in PASCAL VOC challenge. We annotate all following frames in this way.

\subsection{Dataset statistics}
Since the aim of the dataset is to cover as many scenarios as possible in real world tracking applications, the diversity of the video clips is important. Figure \ref{tab:statistics} summaries the statistics of our RGBD tracking dataset, which presents varieties in the following aspects:

\paragraph{Target type}
We divide targets into three types: human, animal and relatively rigid object. Rigid objects, such as toys and human faces, can only translate or rotate. Animals include dogs, rabbits and turtles, whose movement usually consists of out-of-plane rotation and some deformation.
The degrees of freedom for human body motion is very high, and body parts, such as arms and legs, are often slim, resulting in a variety of deformation which may increase the difficulty for tracking.

\paragraph{Target speed}
Tracking difficulty is often related to target speed. We denote target speed using $1- r_{(i,i+1)}$, where $r_{(i,i+1)}$ is the ratio of overlap between target bounding boxes in two consecutive frames when no occlusion occurs. Target speed of a video sequence is defined by its maximum during the sequence. 
Compared to the real speed of the target,
this definition of speed has a more direct influence on tracking performance, as it takes into account differences in frame rate. Average target speed in our video data set ranges from 0.057 to 0.599.

\paragraph{Scene type}
Background clutter is also an important factor affecting tracking performance. In our data set, we provide several types of scenes to with different levels of background clutter. The scenes include cafe, concourse, library, living room, office, playground and sports field. The living room, for example, has a simple and mostly static background, while the background of a cafe is complex, with many people passing by.

\paragraph{Presence of occlusion}
Out of 100 videos in our dataset, occlusion occurs in 63 videos, in which the targets are totally occluded in 16.3 frames on average.  The videos cover several aspects that may affect tracking performance under occlusion, \eg how long the target is occluded, whether the target moves or changes in appearance during occlusion, and the similarity between the occluder and the target.

\begin{figure*}[t]
\centering
\includegraphics[width=0.8\linewidth]{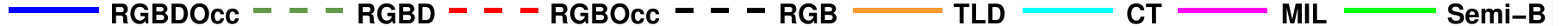}
\vspace{-2mm}

\subfigure[Test cases without occlusion]{\includegraphics[width=0.33\linewidth]{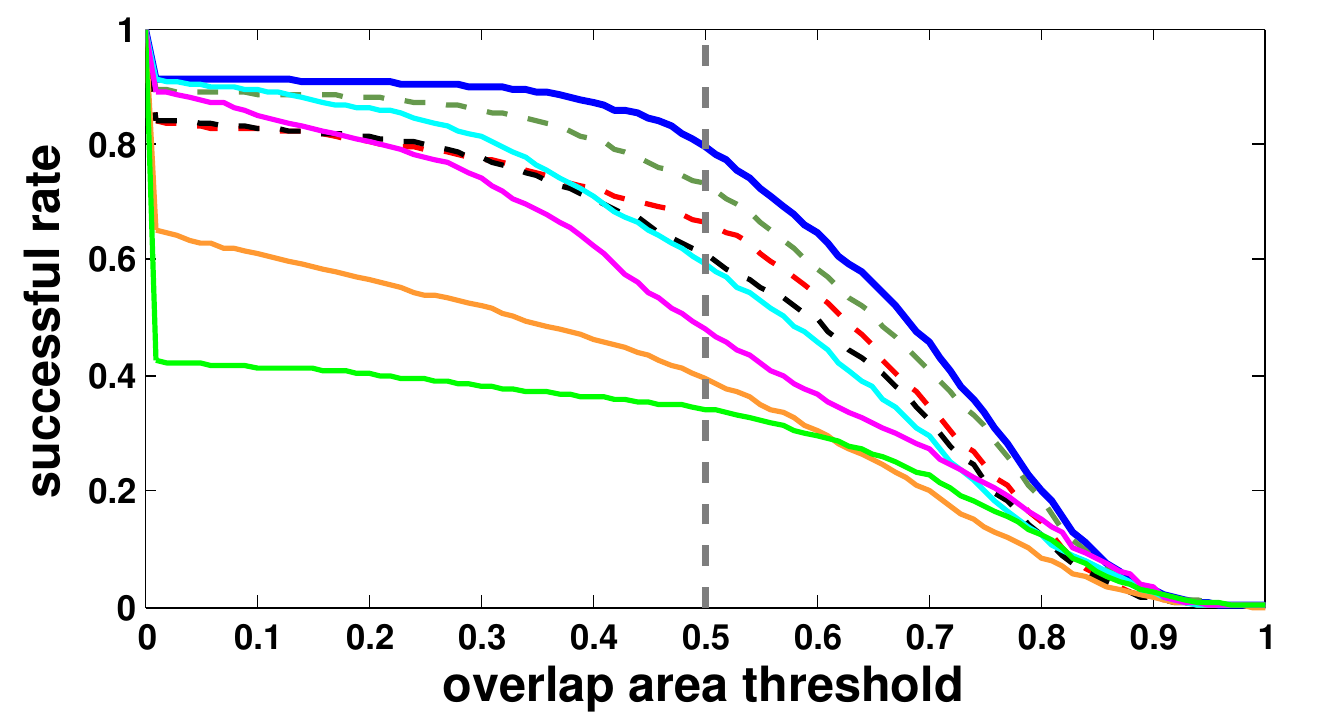}}~%
\subfigure[Test cases with occlusion]{\includegraphics[width=0.33\linewidth]{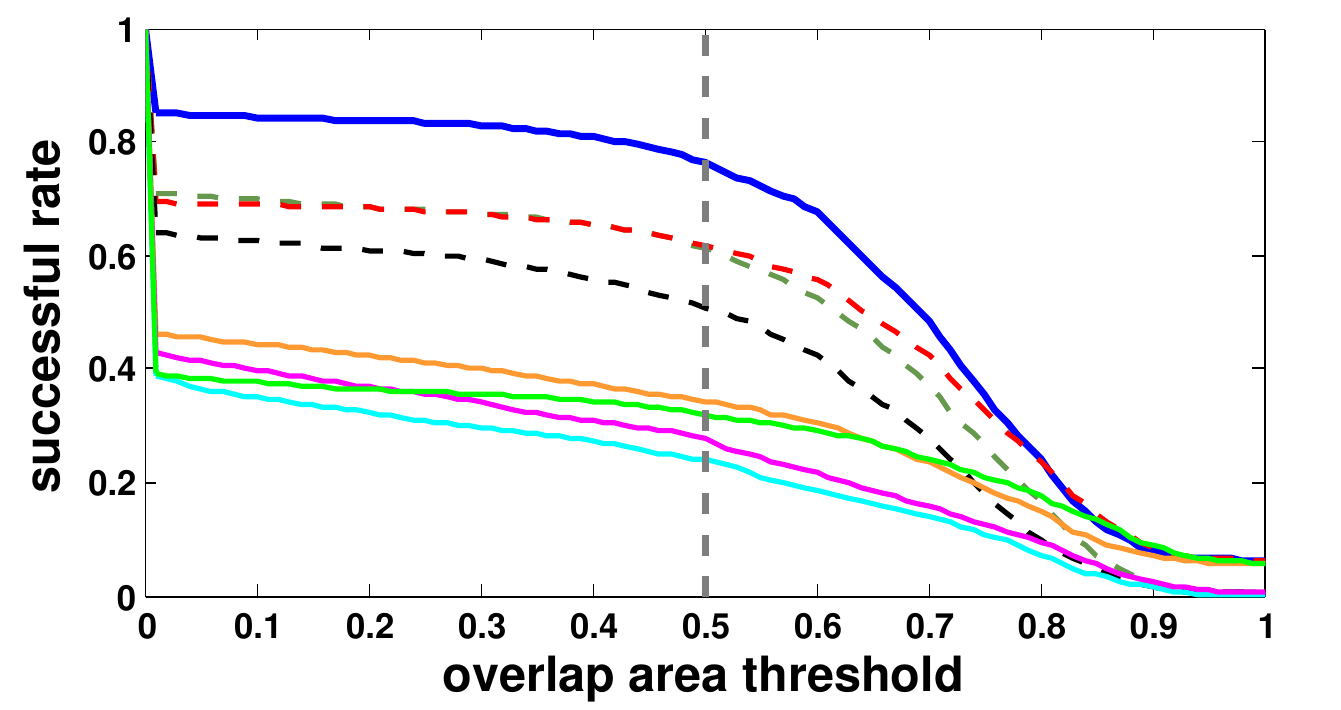}}~%
\subfigure[All test cases]{\includegraphics[width=0.33\linewidth]{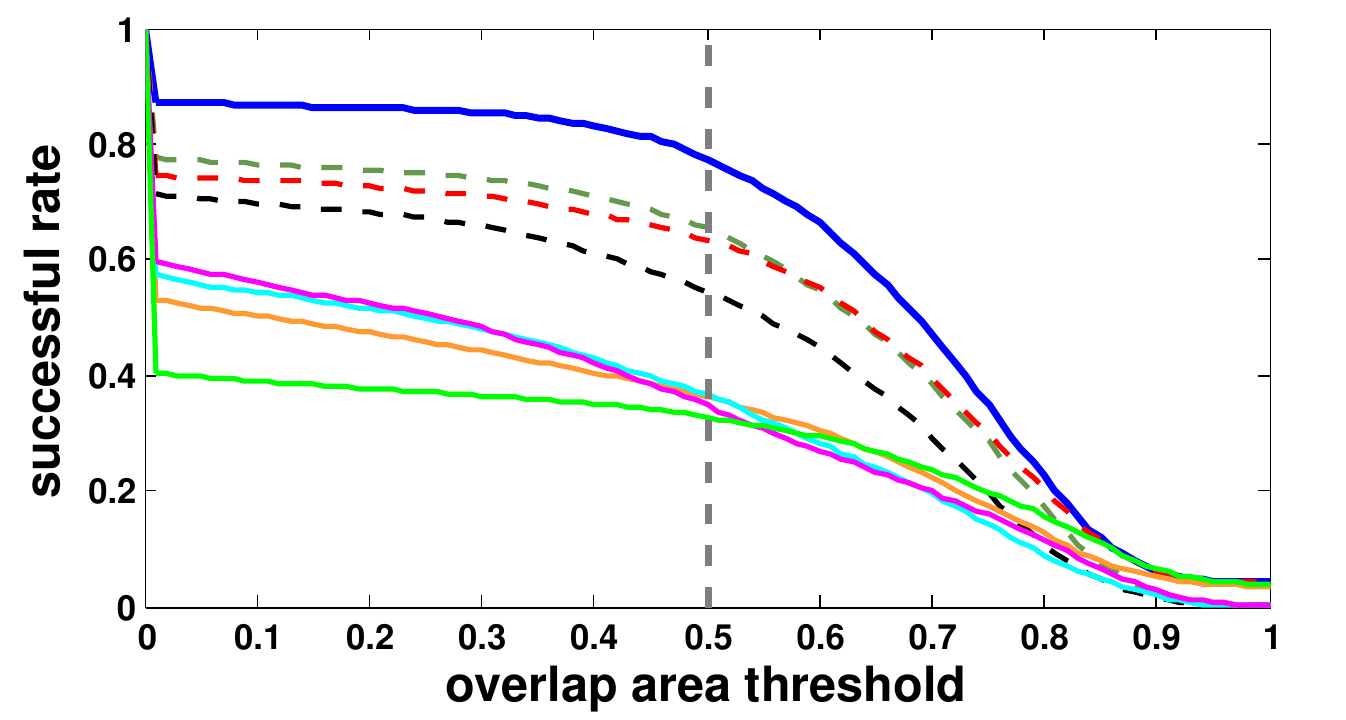}}
\caption{Average success rate vs. threshold of overlap ratio ($r_t$) evaluated on different categories of test cases.}
\label{SR}
%\vspace{-3mm}
\end{figure*}

\subsection{Evaluation metric}

We used two metrics to evaluate the proposed baseline algorithm with other state-of-the-art trackers. One metric is center position error (CPE) which is the Euclidean distance between the centers of output target bounding boxes and the ground truth. The above metric shows how close the tracking results are to the ground truth in each frame. However, the overall performance of the trackers cannot be measured by averaging this distance. When the trackers are misled by background clutter, such distances can be huge, thus the average distance may be dominated by only a few frames. Also, this distance is undefined when trackers fail to output a bounding box or there is no ground truth bounding box (target is totally occluded).
		
To evaluate the overall performance, we employ the criterion used in the PASCAL VOC challenge \cite{voc}, the ratio of overlap $r_i$ between the output and true bounding boxes:
\begin{equation}
\label{eq:overlap}
r_{i}=\begin{cases}
\frac{\text{area}(\text{ROI}_{T_{i}}\cap\text{ROI}_{G_{i}})}{\text{area}(\text{ROI}_{T_{i}}\cup\text{ROI}_{G_{i}})} & \text{if both \ensuremath{\text{ROI}_{T_{i}}} and \ensuremath{\text{ROI}_{G_{i}}} exist}\\
1 & \text{if both \ensuremath{\text{ROI}_{T_{i}}} and \ensuremath{\text{ROI}_{G_{i}}} not exist}\\
-1 & \text{otherwise}
\end{cases}
\end{equation}
where $\text{ROI}_{T_i}$ is the target bounding box in the $i$-th frame and $\text{ROI}_{G_i}$ is the ground truth bounding box. By setting a minimum overlapping area $r_t$, we can calculate the average success rate $R$ of each tracker as follows: 
		\begin{align}
			R &= \dfrac{1}{N} \sum\limits_{i=1}^{N}u_i,\\
			u_i &= \begin{cases}
				1 & \text{if } r_i>r_t\\
				0 & \text{otherwise}
			\end{cases},
		\end{align}
where $u_i$ is an indicator denoting whether the output bounding box of the $i$-th frame is acceptable, and $N$ is the number of frames. According to the above calculation, the minimum overlap ratio $r_t$ makes a hard decision on whether an output is valid or not. Since some trackers may produce outputs that have small overlap ratio over all frames while others give large overlap on some frames and fail completely on the rest, $r_t$ must be treated as a variable to conduct a fair comparison.

In Figure \ref{errRate}, we further divide tracking failures into three types:
\begin{align*}
\text{Type I  }:& \text{ROI}_{T_i}\ne null\text{ and } \text{ROI}_{G_i}\ne null\text{ and }r_i<r_t\\
\text{Type II }:& \text{ROI}_{T_i}\ne null\text{ and } \text{ROI}_{G_i}=null \\
\text{Type III}:& \text{ROI}_{T_i}=null\text{ and } \text{ROI}_{G_i}\ne null
\end{align*}
Type I error is the case where target is visible, but tracker's output is far away from the target. Type II error is where target is invisible but tracker outputs a bounding box. Type III error is where target is visible but tracker fails to give any output.

Running time of each algorithm is not included in our metrics, 
because our main focus is on the performance of 
the tracking algorithm on RGBD data. 
For our current implementation of the baseline algorithm, 
we tried to keep the system as simple as possible, 
so the code is written in Matlab and is not optimized for speed at all. 
There are many potential ways to speed up the algorithm, 
if we wish to use it in real time applications.
For example, instead of a naive convolution for the sliding window detector, 
we can use \cite{FastConv} for acceleration. 
Instead of an optical flow \cite{LargeFlow} running in CPU,
we can use an optical flow running on GPU \cite{GPUFlow}.
Furthermore, 
there are also many ways to maximize the efficiency using special hardware,
such as FPGA or other customized hardware ASIC circuits.

\subsection{Evaluation results}

To understand how much of the performance improvement is due to the use of depth data and how much is due to occlusion handling, we tested four versions of our proposed baseline tracker, which are: 
\begin{description}
\vspace{-2mm}
\item[RGB] uses RGB features without occlusion handling. 
\vspace{-2mm}
\item[RGBD] uses RGBD features without occlusion handling. 
\vspace{-2mm}
\item[RGBOcc] uses RGB features with occlusion handling. 
\vspace{-2mm}
\item[RGBDOcc] uses RGBD features with occlusion handling enabled, which is our complete baseline algorithm.
\end{description}

We also compare the baseline algorithms to 
four state-of-the-art RGB trackers: 
TLD\cite{tld}, CT\cite{ct}, MIL\cite{mil}, semi-B\cite{semiB}, 
The performance measured by CPE and the corresponding snapshots are shown in Figure \ref{fig:result}, and the success rates measured by overlap ratio are shown in Figure \ref{SR}. Error decomposition of each tracker is shown in Figure \ref{errRate}.
Furthermore, 
we define an average ranking of different algorithms,
based on a combination of several indicators, 
as shown in Table \ref{table:ranking}.

We can clearly see that
the proposed baseline RGBD tracker significantly outperforms all others,
which indicates that the extra depth map with some occlusion reasoning 
provides valuable information which helps to achieve a better tracking result.
The proposed methods use very powerful but more computationally expensive classifiers (with hard negative mining) as well as a state-of-the-art optical flow algorithm, 
while other trackers mainly focus on real-time performance. 
Thus our RGB tracker is expected to have higher accuracy at the cost of longer running time. 
However, the effect of using depth data can still be seen by comparing the results of the tracker with depth input (RGBD) and without (RGB). With depth data, error is reduced by 10.9\%. After enabling the occlusion handler of the RGBD tracker, its error rate further decreased by 12.3\%.
When compared with other state-of-art trackers, the proposed algorithm achieves an average 42.3\% reduction on error rate. In particular, when occlusion is present the occlusion detection and handling is critical to reduce error, as shown in Figure \ref{errRate} and \ref{SR} (b).

Distinguishing three types of error helps analyze different sources of error. 
For example, TLD and SemiB have a relative high Type III error, suggesting that their models are sensitive to target appearance change or partial occlusion,
while MIL, CT, and RGBD have high Type II error, 
resulting from the lack of an active occlusion detection mechanism. 
However, each error type cannot be considered separately as a direct indicator of performance. For example, MIL and CT use target models which are less sensitive to occlusion and thus have a very low Type III error at the cost of high Type II error when target is occluded, 
and possible high Type I error in the following frames after occlusion if trackers are misled by the occluder. 
Our proposed tracker robustly handles different scenarios and achieves the lowest overall error rate.

\begin{table*}[t]
\caption{Evaluation results: successful rate \% and corresponding ranking (in parentheses) under different categorizations.}
\vspace{1mm}
\centering
\setlength{\tabcolsep}{5.1pt}
{
\small
\begin{tabular}{c|c|c|c|c|c|c|c|c|c|c|c|c}
\hline 
\multirow{2}{*}{algorithm} & \multirow{2}{*}{\specialcell{avg.\\ rank}} &
\multicolumn{3}{c|}{target type} & \multicolumn{2}{c|}{target size} & \multicolumn{2}{c|}{speed} & \multicolumn{2}{c|}{oclussion }  & \multicolumn{2}{c}{motion type}\tabularnewline
\cline{3-13} 
 &  & human & animal & rigid & large& small&slow & fast & occ & no occ & passive & active\tabularnewline
\hline 
RGBDOcc & 1 & 82.0(1) & 60.2(1) & 81.9(1) &82.9(1)&74.3(1)& 77.5(1) & 77.3(1) & 76.1(1) & 79.5(1) & 81.5(1) & 75.7(1)\tabularnewline
\hline 
RGBD & 2.36 & 66.0(2)&58.3(2)&68.2(3)&69.5(2)&62.2(2)&64.9(2)&65.8(2)&61.1(3)&73.1(2)&67.7(3)&63.7(2)\tabularnewline
\hline 
RGBOcc&2.63& 65.3 (3)&49.2(3)&73.4(2)&64.9(3)&62.0(3)&64.8(3)&63.1(3)&61.5(2)&66.5(3)&75.4(2)&58.1(3)\tabularnewline
\hline
RGB   &4   & 54.7 (4)&48.4(4)&56.8(4)&55.4(4)&53.5(4)&55.5(4)&53.2(4)&50.7(4)&60.9(4)&62.0(4)&51.0(4)\tabularnewline
\hline 
TLD\cite{tld} &6.09 &28.2(7) & 30.7(8) &43.7(5)&32.4(7) &38.5(5) & 44.8(6) & 29.5(5) & 34.3(5) & 39.3 (7)& 47.5(5) & 31.8(7)\tabularnewline
\hline 
CT\cite{ct}& 6.27& 33.0 (6) & 43.6(5) & 33.4(8) &41.3(5) &33.5(7) &47.3(5) & 27.5(8) & 23.8(8) & 59.2(5) & 38.9(7) & 35.3(5)\tabularnewline
\hline 
MIL\cite{mil}& 6.54 &34.3(5)& 34.8(6)& 34.2 (7)& 39.6(6) & 32.9(8) & 40.4(8) & 29.5 (5)& 27.7 (7)& 48.1(6) & 38.8(8) &34.0(6)\tabularnewline
\hline 
SemiB\cite{semiB}& 6.90 & 26.1(8) & 31.7(7)  & 38.8(6) &27.6(8) &35.0(6) & 44.8(6) & 29.0(7) & 31.8(6) & 34.1(8) & 46.3(6)&26.7(8)\tabularnewline
\hline 
\end{tabular}
}
\label{table:ranking}
\vspace{-3mm}
\end{table*}

\subsection{Discussion}
From the evaluation results obtained in the previous section, we observed that traditional RGB trackers produce relative high error in the following scenario:

\paragraph{Target rotation and deformation}
Target rotation, especially out-of-plane rotation, and deformation are the main causes of model drifting for traditional RGB trackers. Target appearance can change significantly after rotation or deformation, making recognition difficult. In the video ``stuffed bear'' (Figure \ref{fig:result} Row 1), TLD, CT, MIL and Semi-B lose tracking when the stuffed bear starts to rotate out of plane. In ``basketball player'' (Figure \ref{fig:result} Row 2), those trackers gradually fail to follow the player as he moves his arms and legs. 
However, the RGBD tracker was robust in these situations as we used depth information to identify the target. The depth feature is still distinguishable when the similarity in RGB vanishes.

\paragraph{Different types of occlusion}
There are several factors which may affect the difficulty of tracking under occlusion:
size of target's occluded portion, target movement or appearance variations during occlusion, similarity between occluder and target, and background clutter.

When partially occluded, the target appearance is less similar to the pre-trained models and often cannot pass the threshold. In the video ``human face'' (Figure \ref{fig:result} Row 4), if only RGB data is available, fragment based trackers can locate the target but sometimes mistake background clutter for the target, because with only part of target visible, the detection confidence drops. Conservative approaches, which do not produce output with very low confidence, often lose tracking. When the target is completely occluded (video ``sign'', Figure \ref{fig:result} Row 3), optical flow tracking becomes uninformative. However, from depth data, our method is able to identify the occluder and raise the confidence in its neighboring 3D region, compensating for the confidence loss due to partial occlusion, and thus identifies the target more accurately.

If the occlusion happens gradually, the occluder, if not excluded, slowly grows inside the target bounding box and finally dominates the bounding box (video ``student with bag'', Figure \ref{fig:result} Row 5). On this occasion, optical flow trackers and classifiers are often misled to track or detect the occluder. It is difficult for the trackers to make corrections afterwards because their models are updated incorrectly. Our method detects occlusion more reliably using depth data to recognize the occluder. And by only examining objects around the occluder, we prevent outputting the occluder as the result and update models accordingly.

\section{Conclusions}

Thanks to the great popularity of low cost depth sensors in the consumer market,
tracking can be made easier by using reliable depth data as input.
Object depth data provides information for discriminating between different objects, which cannot be obtained from RGB data alone. 
But there are many questions about how valuable such reliable depth information is for handling occlusion and preventing model drift.
In this paper, we construct a benchmark dataset of 100 RGBD videos with high diversity, including deformable objects, various occlusion conditions and moving cameras.
We propose a very simple but strong baseline model for RGBD tracking, and present a quantitative comparison of several state-of-the-art tracking algorithms.
We have demonstrated that by incorporating depth data, trackers can achieve better performance and handle occlusion more easily as well as more accurately.
With depth data, the baseline RGBD tracker outperforms current state-of-the-art RGB trackers significantly.

We believe that this benchmark dataset and baseline algorithm can provide a better comparison of different tracking algorithms,
and start a new wave of research advances in the field 
by making experimental evaluation more standardized and easily accessible.
The datasets, evaluation details, source code of the baseline algorithm, and instructions for submitting new models will be made available online after acceptance.
In future work, 
we would like to investigate stronger models for both RGB and RGBD tracking, 
such as modeling deformable objects by parts \cite{DPM,treeDPM}.

	\def \n{0.135}
\begin{figure*}[t]
	\centering
	\includegraphics[width=0.8\linewidth]{./image/label.pdf}

	\includegraphics[width=0.135\linewidth]{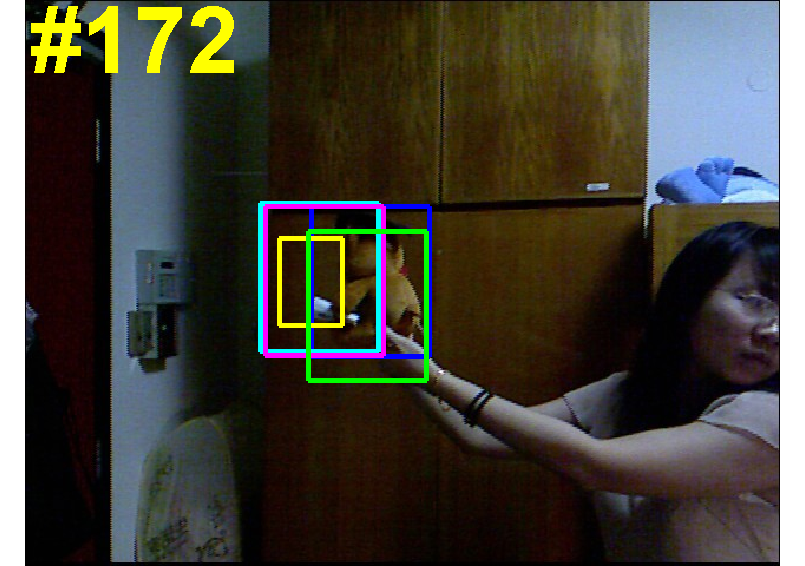}\hspace*{-3pt}
	\includegraphics[width=0.135\linewidth]{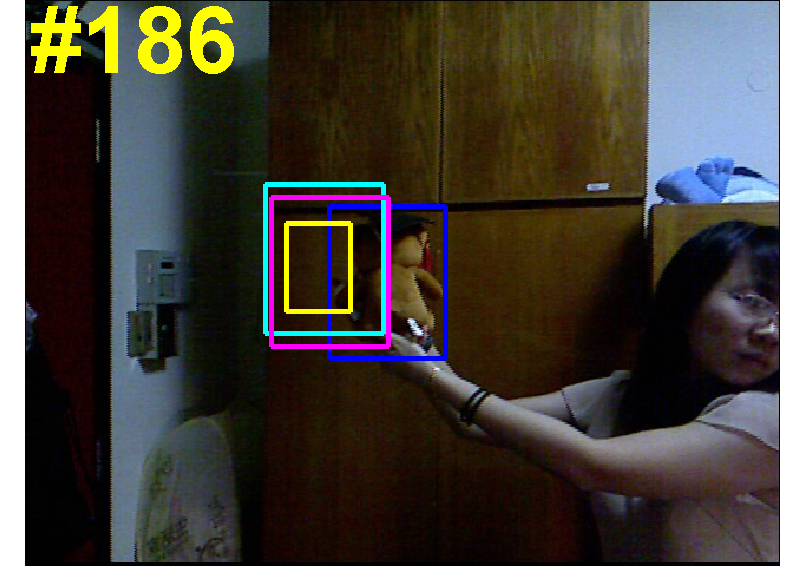}\hspace*{-3pt}
	\includegraphics[width=0.135\linewidth]{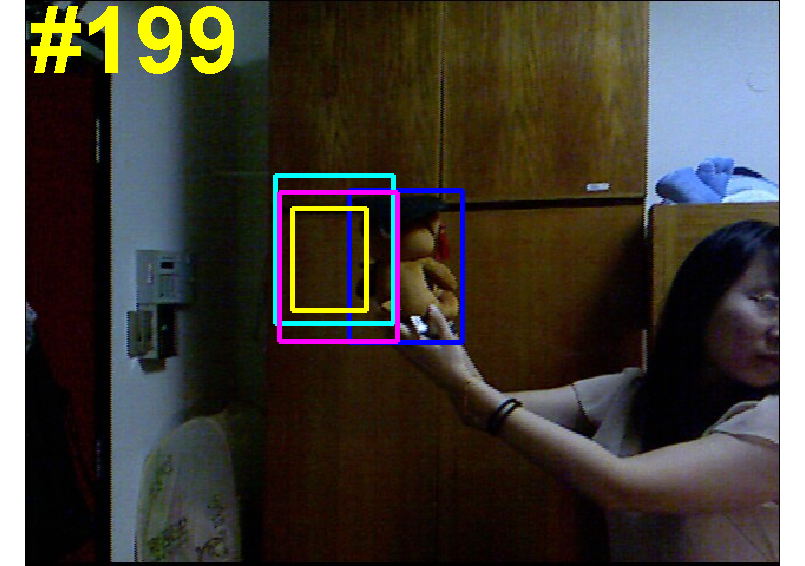}\hspace*{-3pt}
	\includegraphics[width=0.135\linewidth]{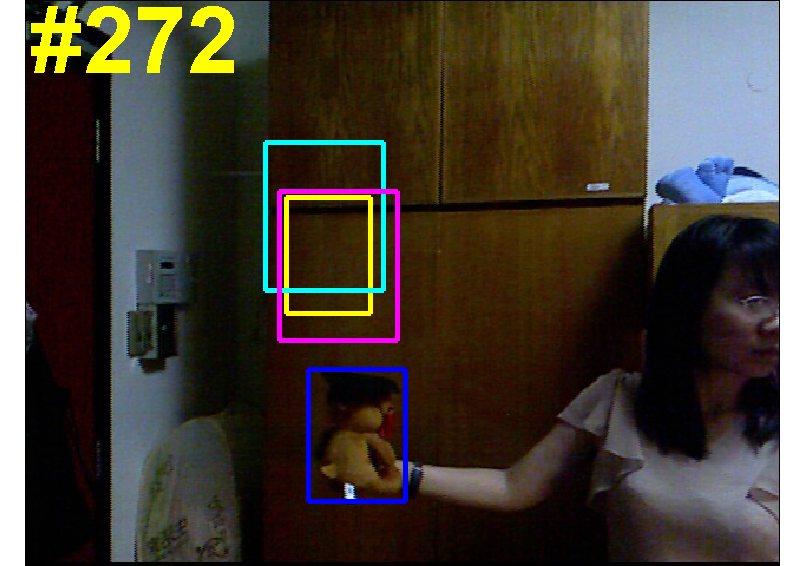}\hspace*{-3pt}
	\includegraphics[width=0.135\linewidth]{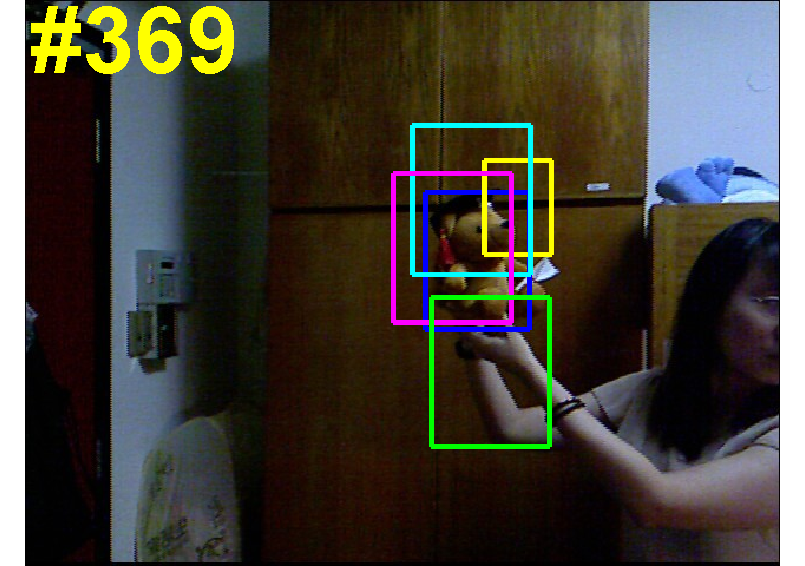}\hspace*{-3pt}
	\includegraphics[width=0.135\linewidth]{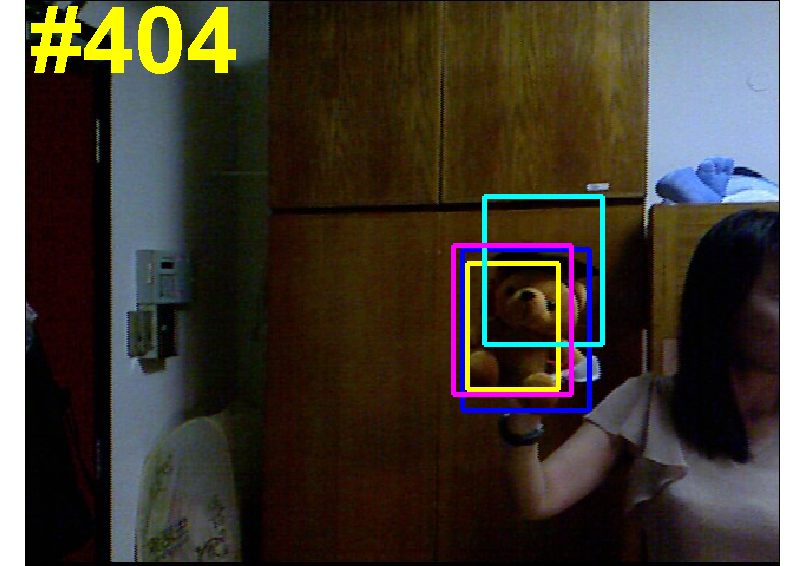}\hspace*{-3pt}
	\includegraphics[width=0.205\linewidth]{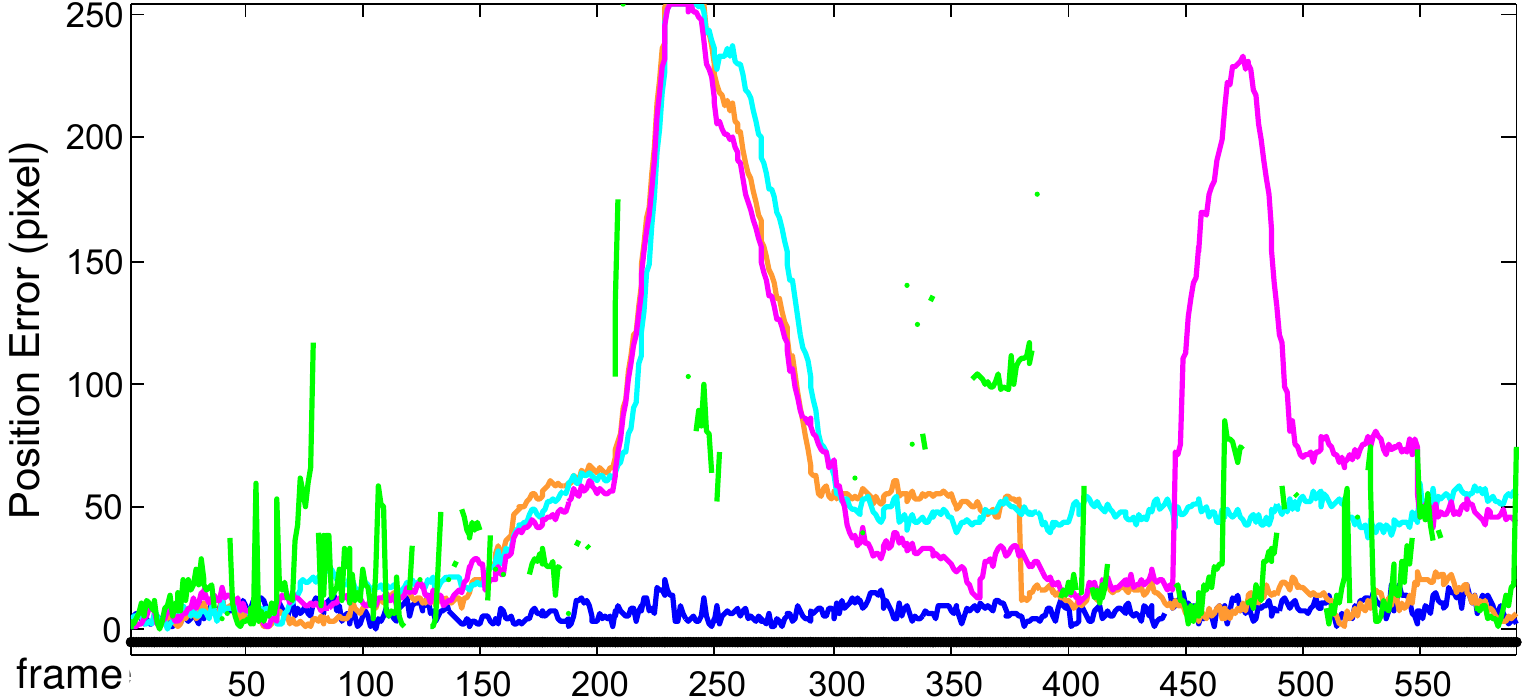}	

	\vspace{1.5mm}

	%\caption{stuffed bear}\label{bear}
	\includegraphics[width=\n\linewidth]{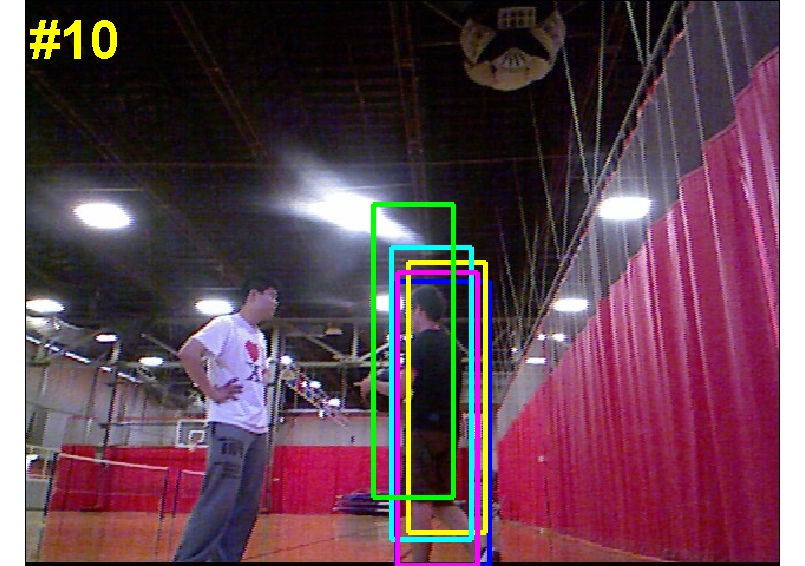}\hspace*{-3pt}
	\includegraphics[width=\n\linewidth]{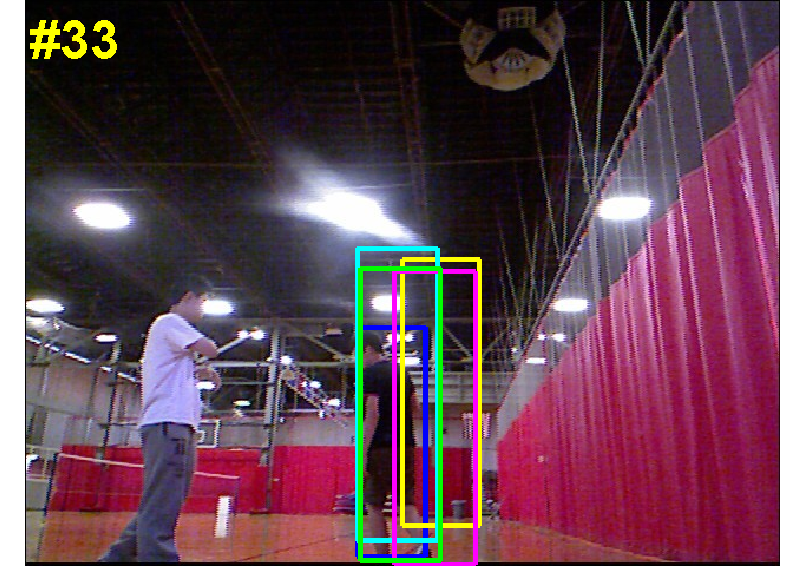}\hspace*{-3pt}
	\includegraphics[width=\n\linewidth]{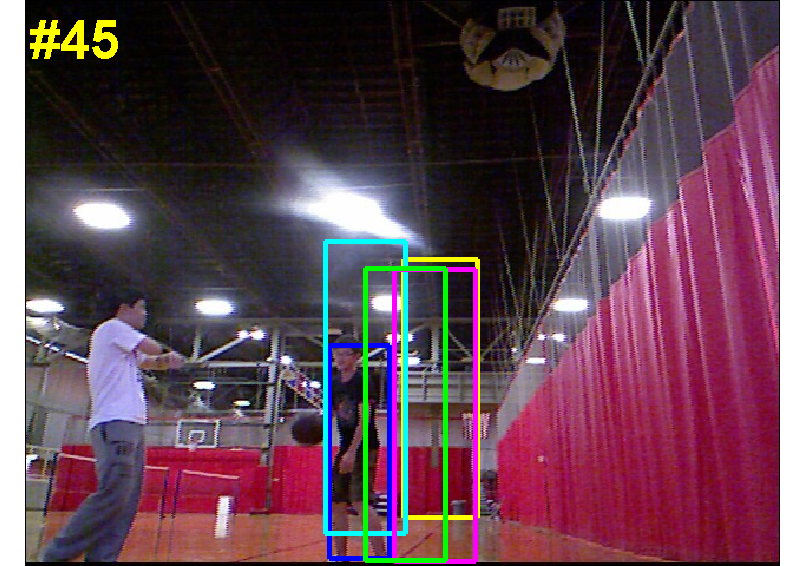}\hspace*{-3pt}
	\includegraphics[width=\n\linewidth]{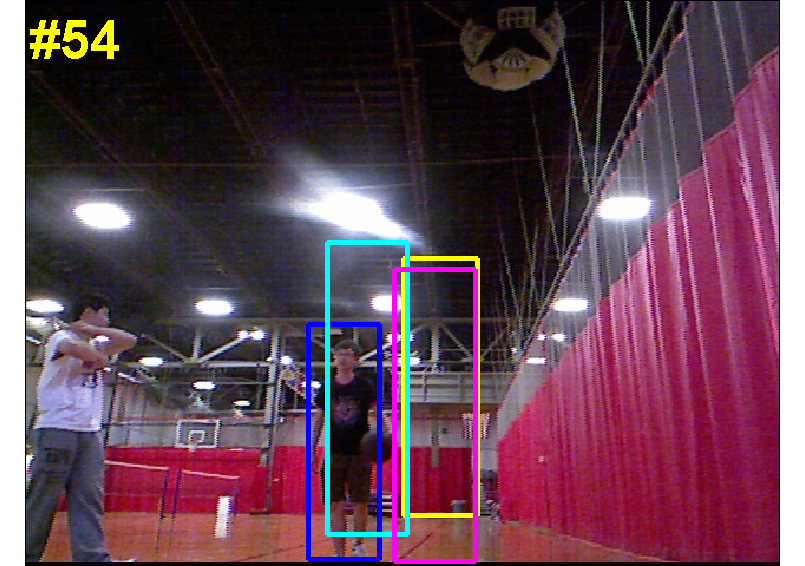}\hspace*{-3pt}
	\includegraphics[width=\n\linewidth]{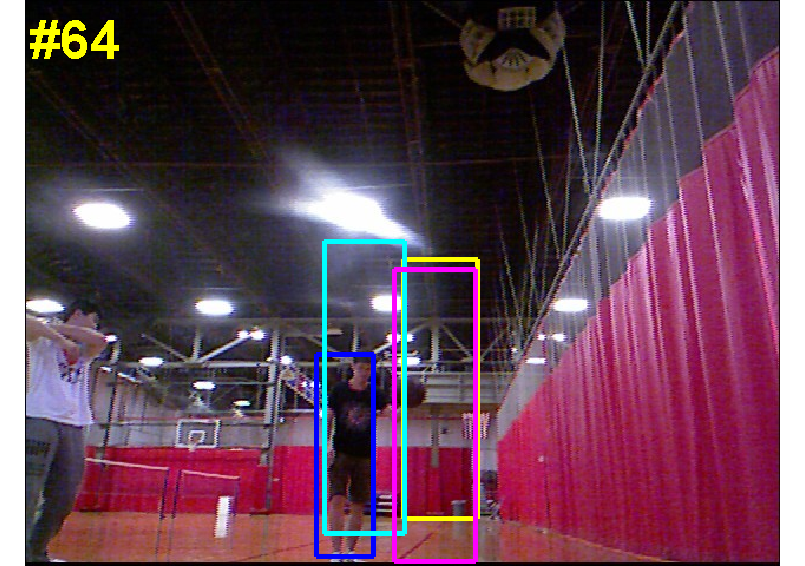}\hspace*{-3pt}
	\includegraphics[width=\n\linewidth]{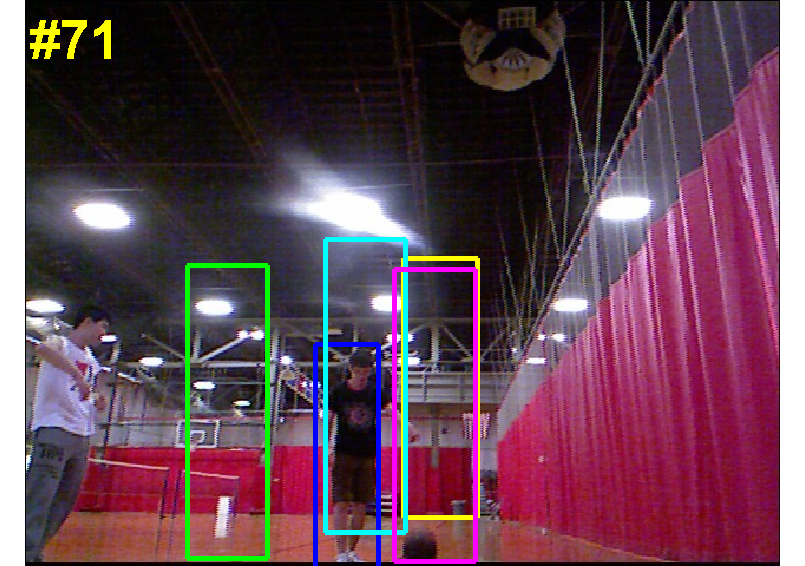}\hspace*{-3pt}
	\includegraphics[width=0.205\linewidth]{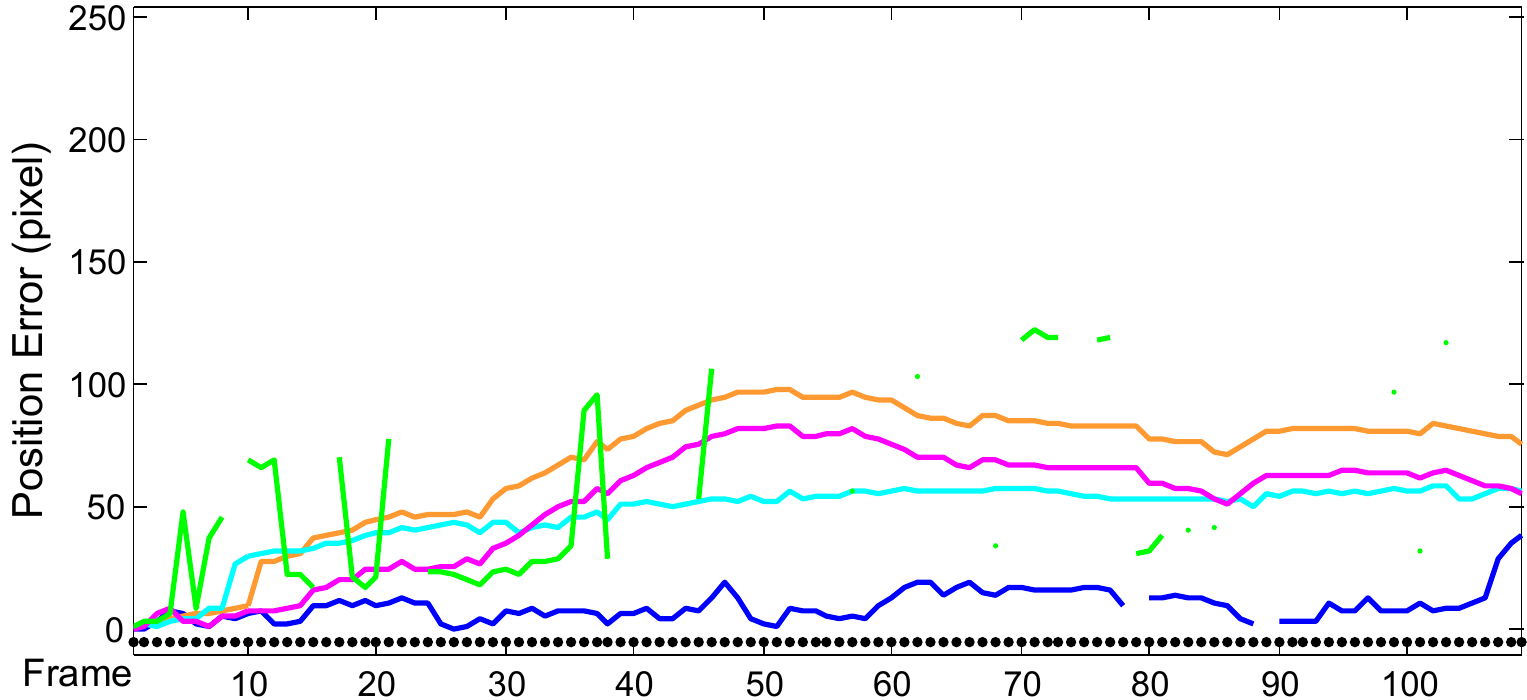}\hspace*{-3pt}
	%\caption{baketball palyer}\label{basketball}

	\vspace{1.5mm}
	
	\includegraphics[width=\n\linewidth]{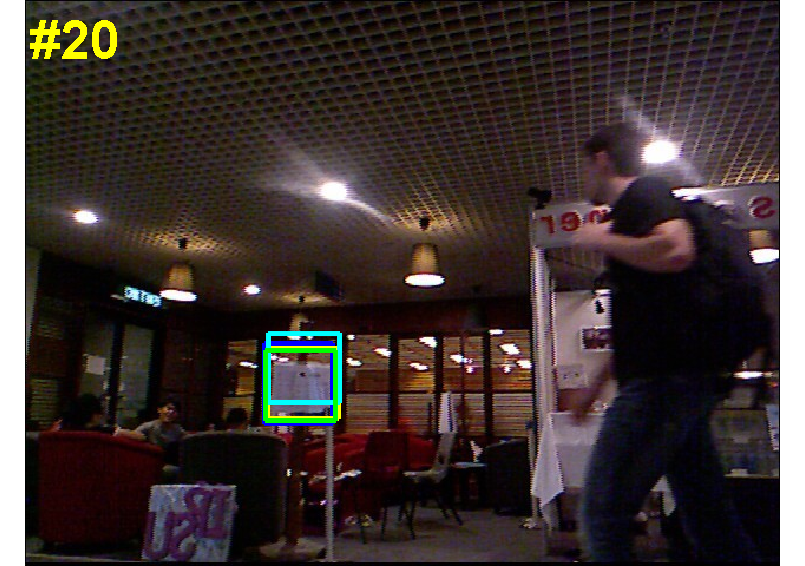}\hspace*{-3pt}
	\includegraphics[width=\n\linewidth]{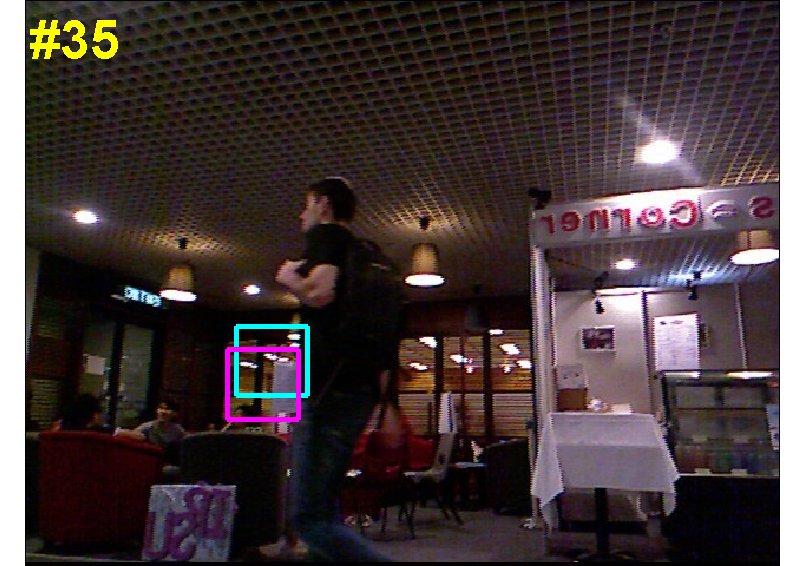}\hspace*{-3pt}
	\includegraphics[width=\n\linewidth]{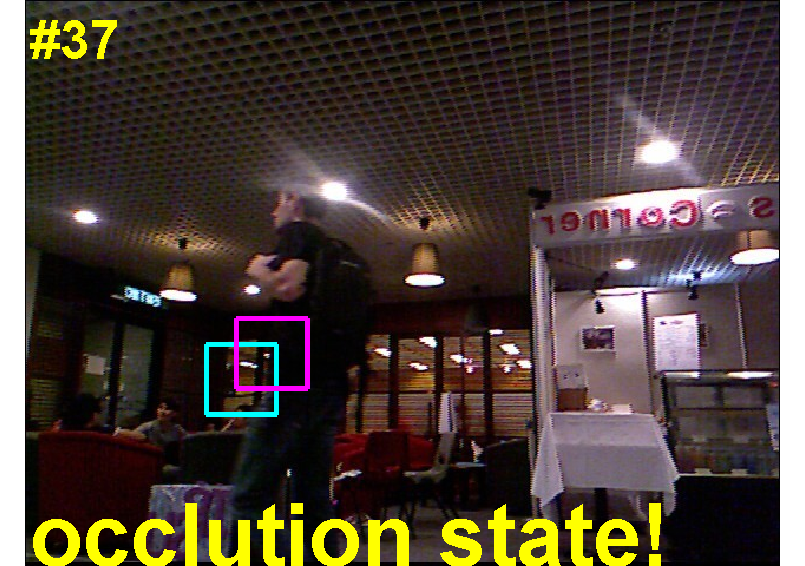}\hspace*{-3pt}
	\includegraphics[width=\n\linewidth]{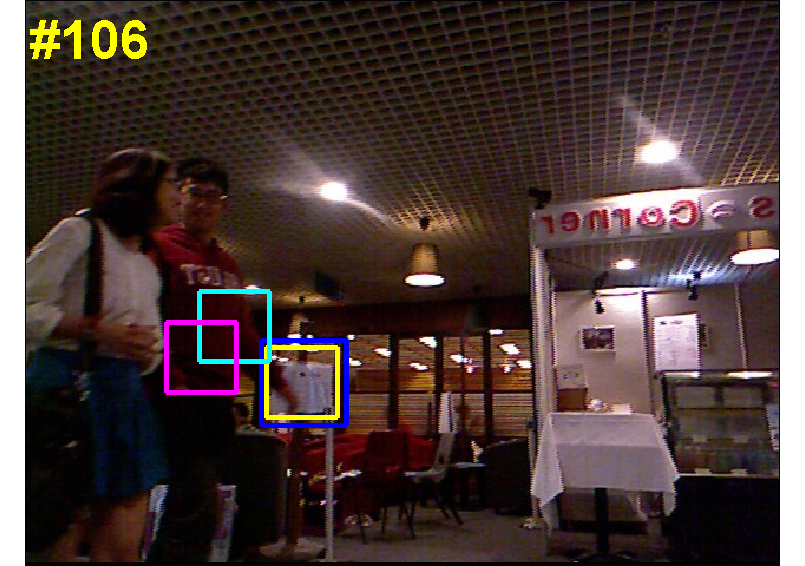}\hspace*{-3pt}
	\includegraphics[width=\n\linewidth]{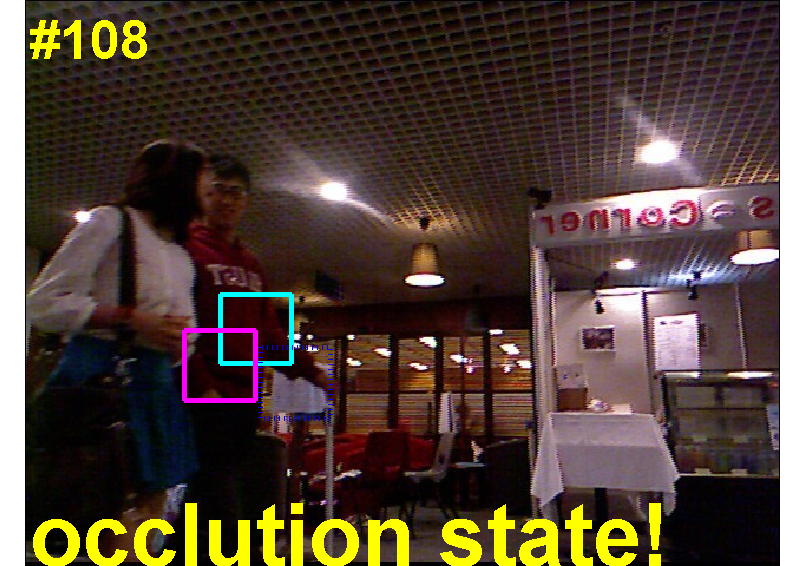}\hspace*{-3pt}
	\includegraphics[width=\n\linewidth]{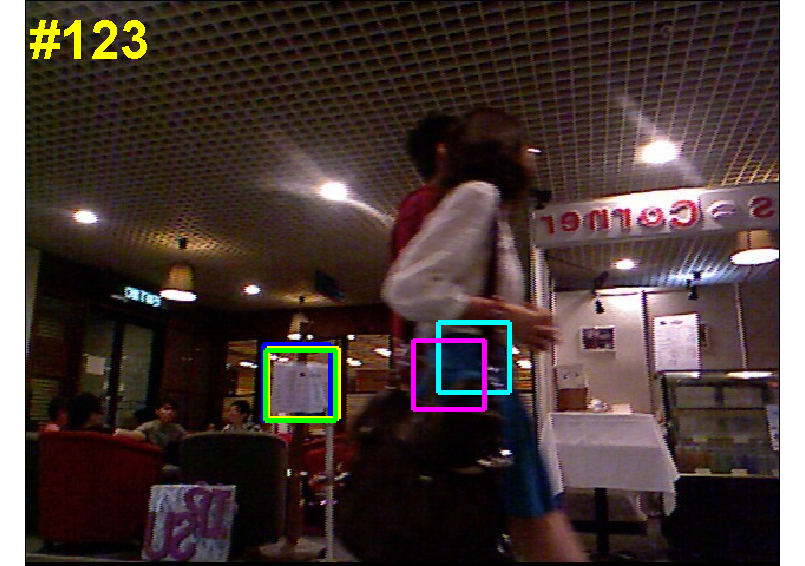}\hspace*{-3pt}
	\includegraphics[width=0.205\linewidth]{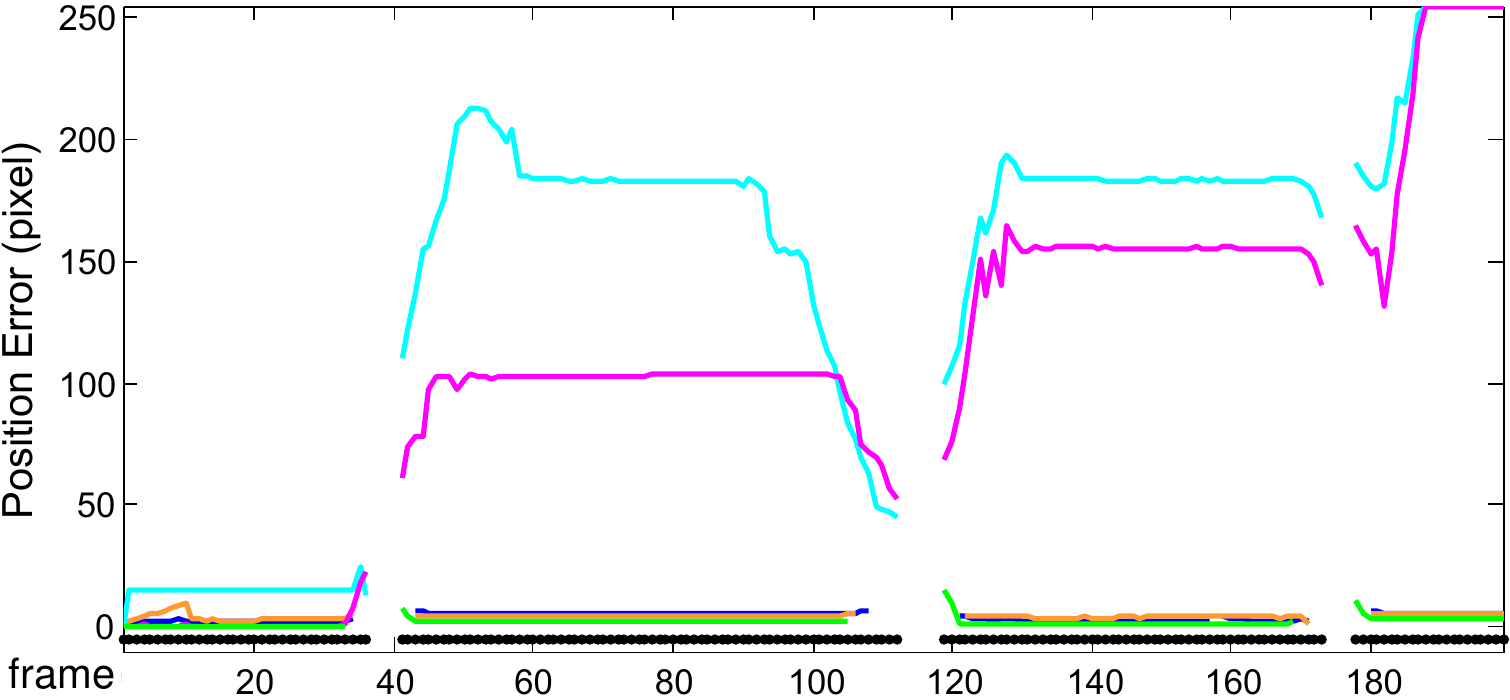}\hspace*{-3pt}
	%\caption{sign}\label{sign}

	\vspace{1.5mm}
	
	\includegraphics[width=\n\linewidth]{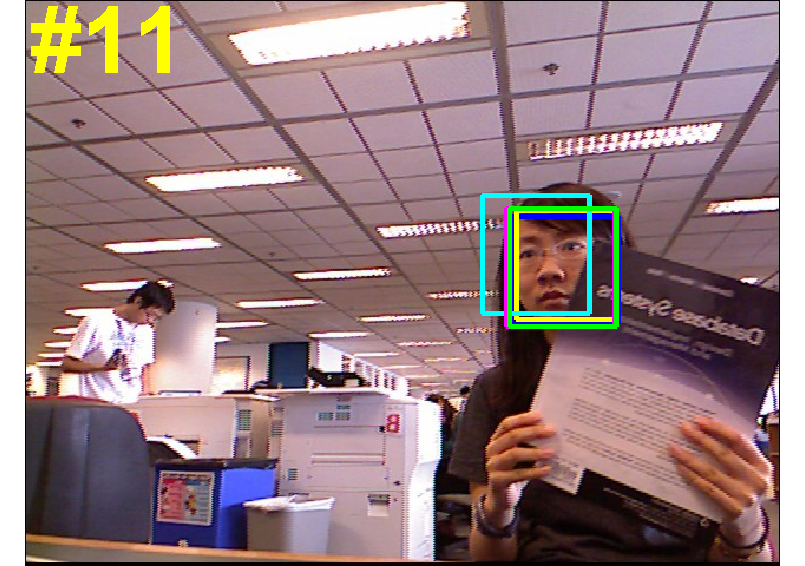}\hspace*{-3pt}
	\includegraphics[width=\n\linewidth]{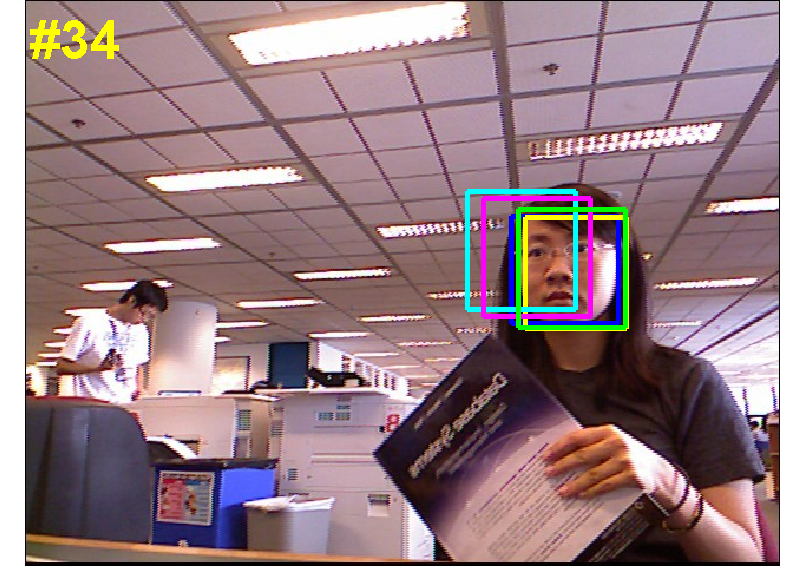}\hspace*{-3pt}
	\includegraphics[width=\n\linewidth]{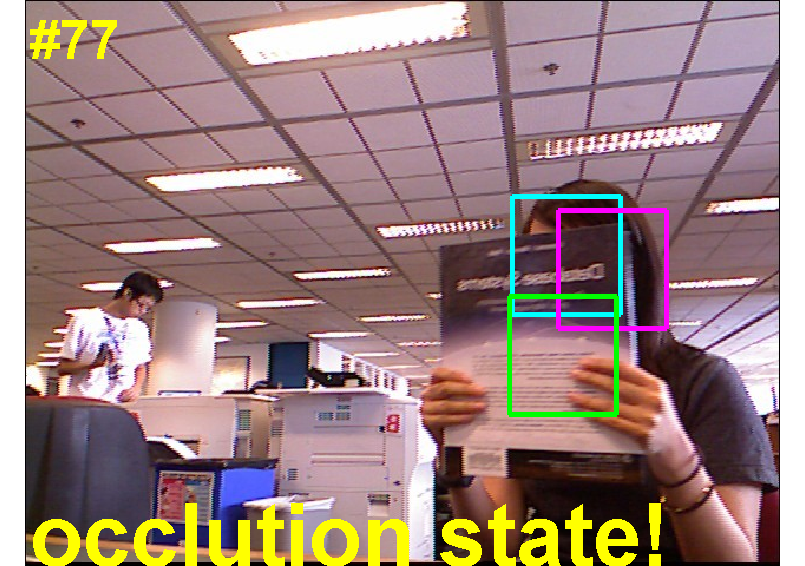}\hspace*{-3pt}
	\includegraphics[width=\n\linewidth]{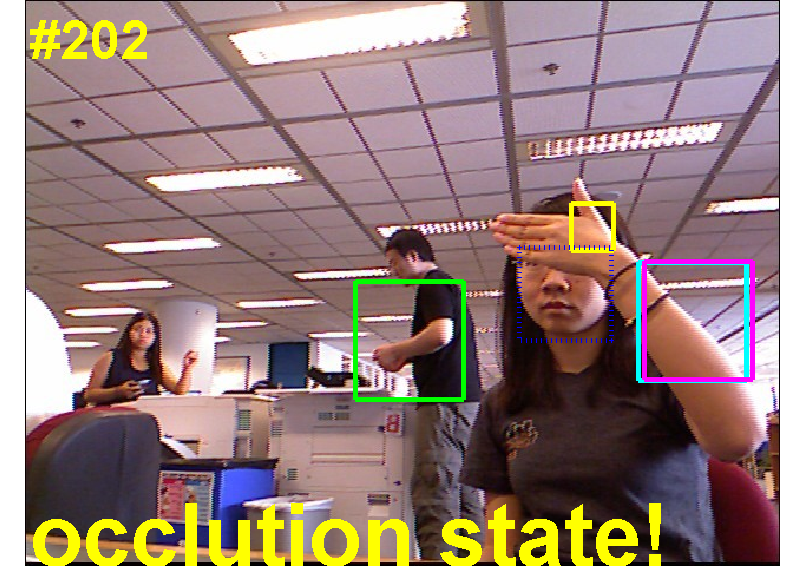}\hspace*{-3pt}
	\includegraphics[width=\n\linewidth]{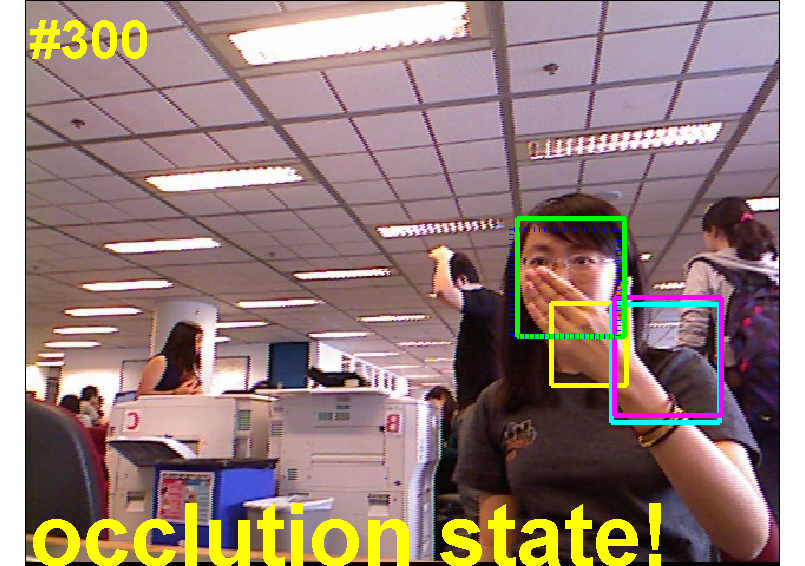}\hspace*{-3pt}
	\includegraphics[width=\n\linewidth]{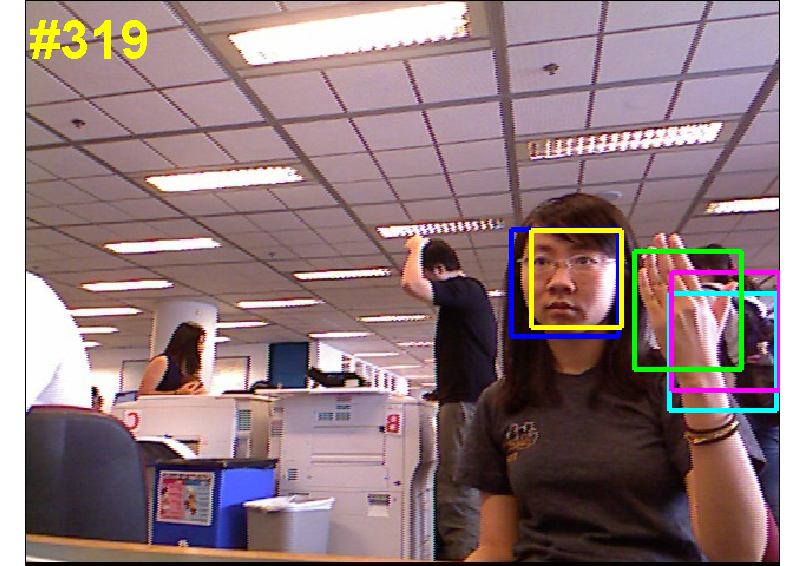}\hspace*{-3pt}
	\includegraphics[width=0.205\linewidth]{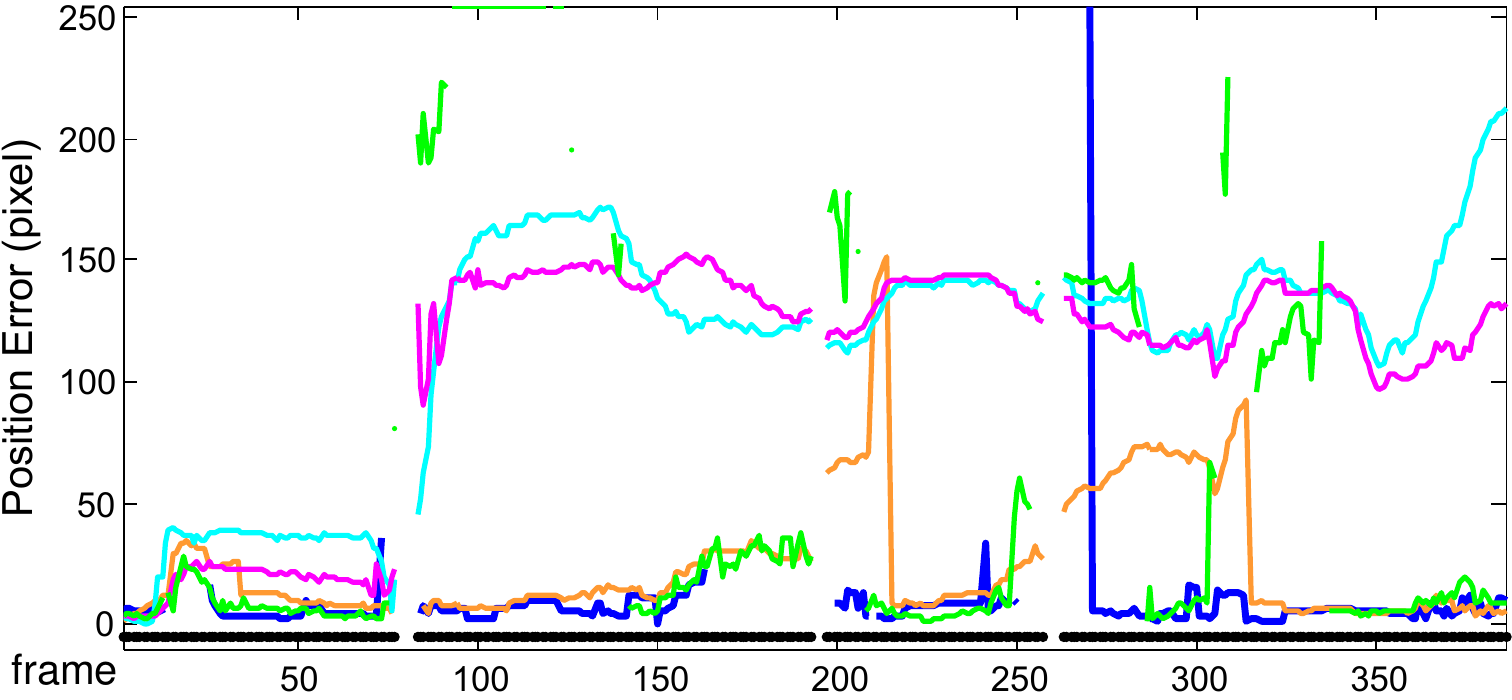}\hspace*{-3pt}
	%\caption{human face}\label{face}

	\vspace{1.5mm}
	
	\includegraphics[width=\n\linewidth]{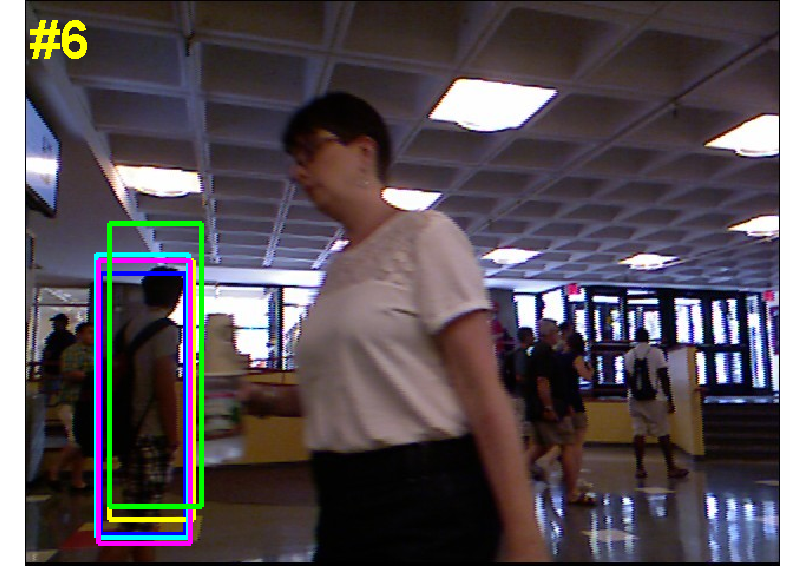}\hspace*{-3pt}
	\includegraphics[width=\n\linewidth]{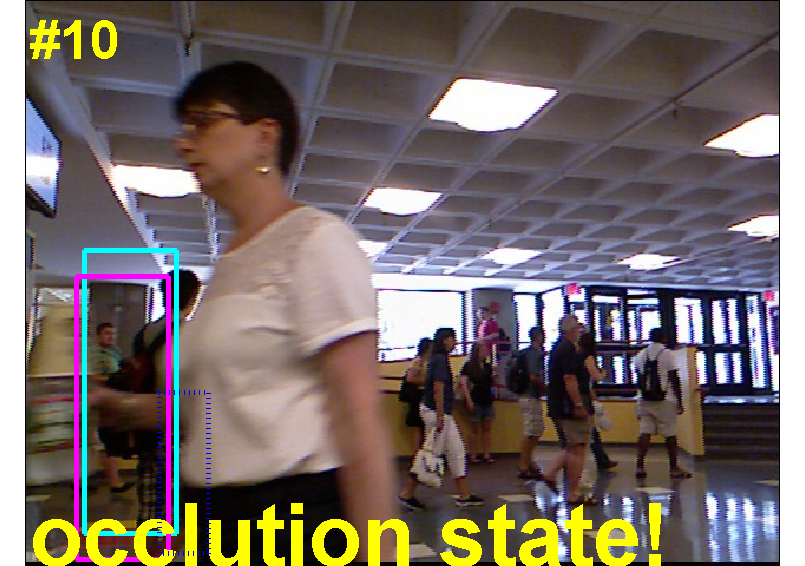}\hspace*{-3pt}
	\includegraphics[width=\n\linewidth]{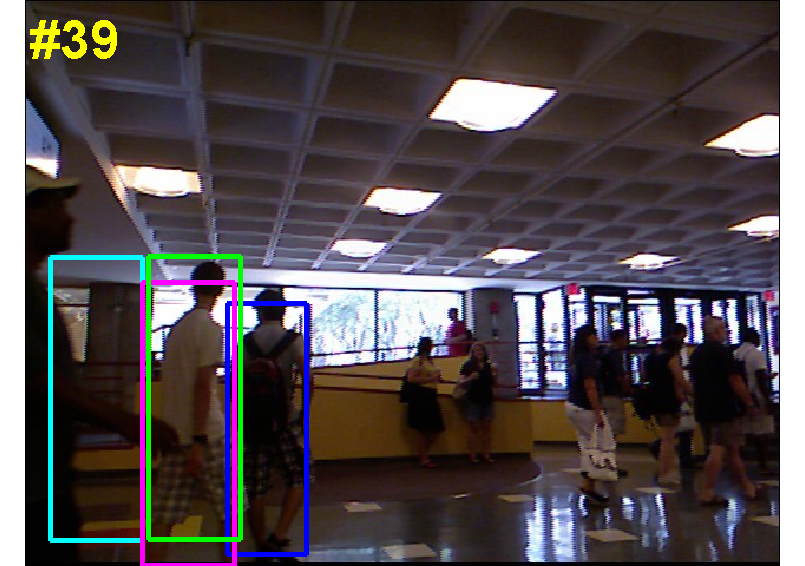}\hspace*{-3pt}
	\includegraphics[width=\n\linewidth]{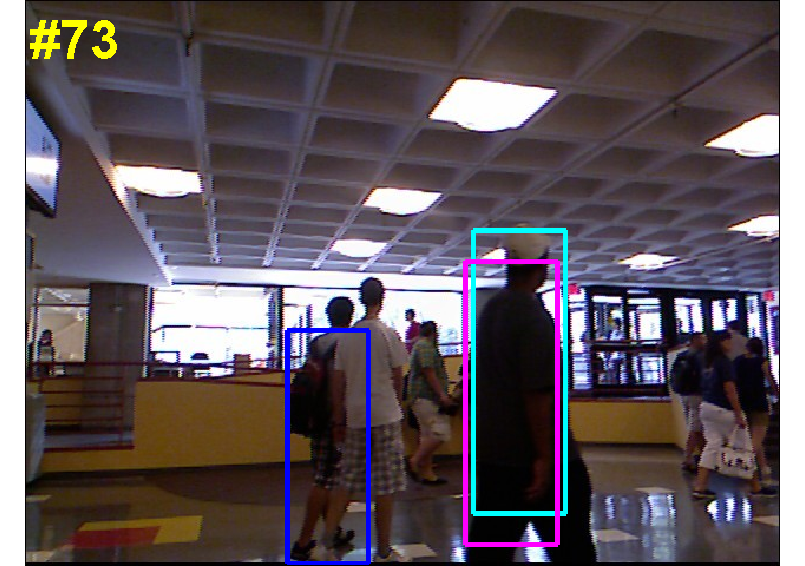}\hspace*{-3pt}
	\includegraphics[width=\n\linewidth]{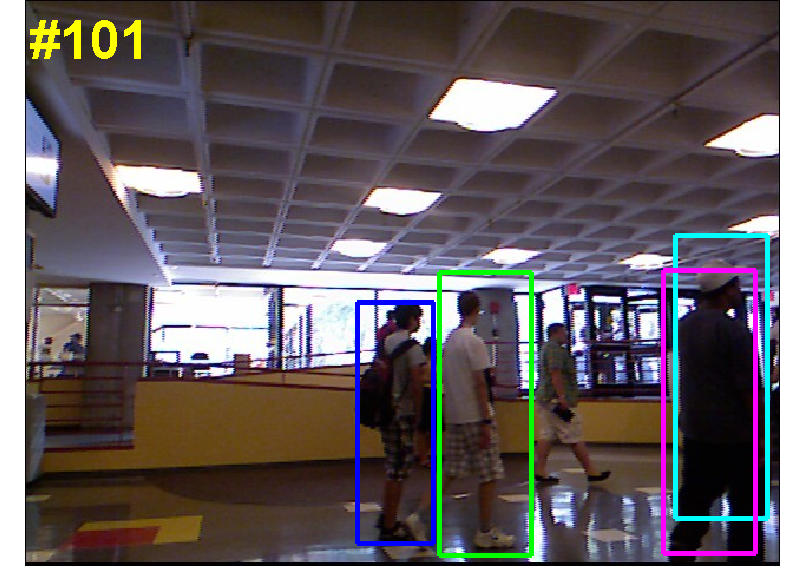}\hspace*{-3pt}
	\includegraphics[width=\n\linewidth]{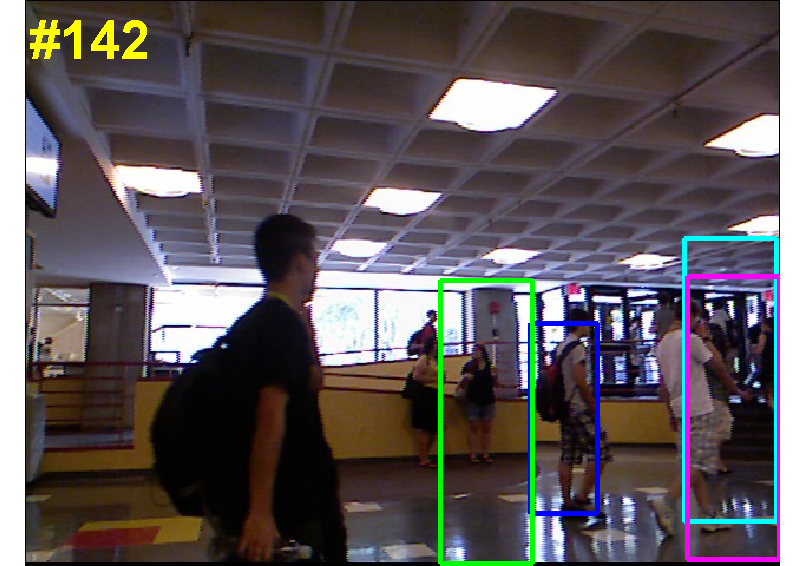}\hspace*{-3pt}
	\includegraphics[width=0.205\linewidth]{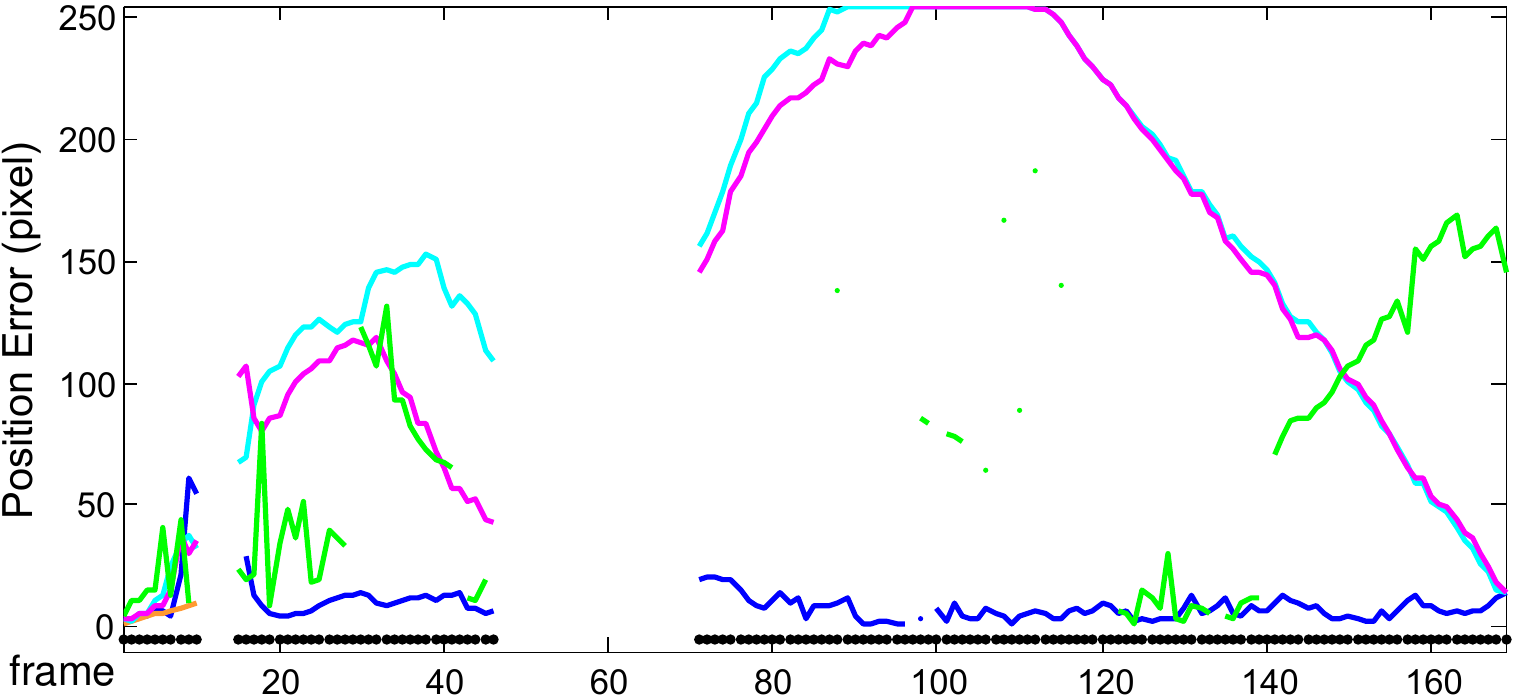}\hspace*{-3pt}
	%\caption{student with bag}\label{sc}
	
\caption{Example results comparing several approaches. More results are available in the supplementary videos. For the sake of clarity we only plot the center position error (CPE) of RGBDOcc, TLD, CT, MIL, Semi-B. The CPE is undefined when trackers fail to output a bounding box or there is no ground truth bounding box (target is totally occluded).}
\label{fig:result}
\vspace{-3mm}
\end{figure*}

{\small
\bibliographystyle{ieee}
\bibliography{RGBDtracker}
}
\end{document}